\documentclass[sageh,times]{sagej}

\usepackage{amsmath,amssymb,amsfonts}
\usepackage{mathtools}
\usepackage[font=small,labelfont=bf]{caption}
\usepackage[font=small,labelfont=bf]{subcaption}
\usepackage{graphicx}
\usepackage{textcomp}
\usepackage[nopatch=footnote]{microtype}
\usepackage{dsfont}
\usepackage{sidecap}
\usepackage{bm}
\usepackage{enumitem}

\usepackage[english]{babel}
\usepackage[autostyle]{csquotes}
\usepackage{hyperref}
\usepackage[table,dvipsnames]{xcolor}
\usepackage{multirow}

\usepackage{multicol}
\usepackage[bottom]{footmisc}
\usepackage{nicefrac}
\usepackage{algorithm2e}
\usepackage[capitalize]{cleveref}
\usepackage{booktabs}
\usepackage{adjustbox}

\usepackage{siunitx}
\RestyleAlgo{boxruled}
\crefname{algocf}{alg.}{algs.}
\Crefname{algocf}{Algorithm}{Algorithms}

\DeclareMathOperator\erf{erf}
\newtheorem{remark}{Remark}
\graphicspath{ {./images/} }

\usepackage{url}

\usepackage{etoolbox}
\makeatletter
\patchcmd{\@maketitle}{10.1177/ToBeAssigned}{10.1177/02783649261508865}{}{\PackageWarning{main}{DOI patch failed}}
\def\volumeyear{2026}
\makeatother

\begin{document}
\runninghead{Paiola et al.}
\renewcommand{\journalname}{The International Journal of Robotics Research}
\title{Risk Density RRT$^*$:\\Planning in the Space of Probability}
\author{Lorenzo Paiola\affilnum{1,2}, Giorgio Grioli\affilnum{1}, and Antonio Bicchi\affilnum{1,2}}
\affiliation{\affilnum{1} Soft Robotics for Human Cooperation and Rehabilitation, Istituto Italiano di Tecnologia, Genova 16163, Italy\\
\affilnum{2} Department of Information Engineering and Centro di Ricerca ``Enrico Piaggio'', University of Pisa, Pisa 56122, Italy}
\corrauth{Lorenzo Paiola, Istituto Italiano di Tecnologia, Genova 16163, Italy.}
\email{lorenzo.paiola@iit.it}
\begin{abstract}
Planning under uncertainty is a fundamental problem in robotics. While much of the literature seeks minimum-cost paths subject to probabilistic constraints, this paper instead considers collision probability, or risk classically defined as probability times severity, directly as the planning objective. We propose Risk Density RRT$^*$ (RD-RRT$^*$), an extension of RRT$^*$ for high-level planning in environments with deterministic and uncertain obstacles. RD-RRT$^*$ builds on Risk Density, a functional that evaluates a continuous path as a whole and provides a tractable approximation of its collision probability. The framework also supports severity-weighted risk and a hybrid variant that adds a path-length penalty, balancing risk and geometric efficiency while regularizing exploration in low-density regions. We validate the method across a wide range of scenarios. The results show that RD-RRT$^*$ consistently finds paths with lower estimated collision probability by optimizing a global risk-based objective. We further demonstrate the fidelity of the underlying approximation through Monte Carlo simulations. Finally, we discuss how the topological structure induced by Risk Density may provide a useful representation of the environment and offer insights into risk-aware planning.
\end{abstract}

\maketitle

\footnotetext[0]{\sagesf\noindent\textbf{Accepted manuscript} of an article accepted for publication in \emph{The International Journal of Robotics Research}, \textcopyright{} 2026 The Author(s). DOI: \href{https://doi.org/10.1177/02783649261508865}{10.1177/02783649261508865}.}

\section{Introduction}
\label{sec:intro}
In the field of robotics, planning under uncertainty is a critical challenge, particularly when human presence introduces dynamic elements into the environment. Even in scenarios where robots possess accurate environmental maps and minimal internal noise, the unpredictable nature of real-world interactions necessitates robust planning strategies.

One significant aspect of this challenge is managing the risk of collisions. Consider, for instance, an autonomous delivery robot navigating a city or a warehouse. While the robot may easily have access to a sufficiently precise map of its environment, it often lacks complete information about the locations and movements of potential obstacles such as humans or other vehicles. It may, however, possess some partial knowledge that can be expressed in probabilistic terms. That creates the need for planning algorithms that can effectively manage risk and uncertainty.

Traditionally, much of the literature in this domain has focused on optimizing paths under probabilistic constraints. Chance constraints (CCs) are a widely used family of methods that rely mostly on Boole's lemma to sum probabilities along a path, but tend to overestimate risk as discretization resolution increases \citep{blackmoreChanceConstrainedOptimalPath2011}. Iterative risk allocation \citep{masahiroonoIterativeRiskAllocation2008} attempts to mitigate this issue. CCs have also been applied to RRT by growing nodes with distributions known to satisfy constraints, but simplifying path probability to that of the riskiest node \citep{ludersChanceConstrainedRRT2010}.
More recent work has advanced these concepts. Risk-aware nonlinear RRT$^*$ with distributionally robust programming achieves accuracy at the cost of computational intensity \citep{safaouiRiskAverseRRTPlanning2021a}. SCORA uses $\epsilon$-shadows to efficiently bound collision probability of paths modeled as Gaussian processes, reducing conservatism while still relying on discretization \citep{dawsonChanceConstrainedTrajectoryOptimization2023}.
Monte Carlo methods offer accuracy but often at the expense of speed. Some approaches significantly accelerate evaluation, though they lack differentiability for optimization purposes \citep{lambertFastMonteCarlo2008,jansonMonteCarloMotion2015}.
Recent research has explored continuous-time risk estimation. Sum of Squares programming and the Lasserre Hierarchy have been used with RRT for accurate, continuous-time estimation, albeit with high computational demands \citep{jasourConvexRiskBounded2023,lasserreMomentSOSHierarchy2018}. Gaussian Processes have also been used to model the distance-to-collision continuously along the path, and then impose a CC valid for all timesteps \citep{johnsonChanceConstrainedMotionPlanning2021}.
Additionally, meta-algorithms such as AORRT can be used to plan with arbitrary costs by using default path planning algorithms and iteratively shrinking a feasible \enquote{cost space} \citep{hauserAsymptoticallyOptimalPlanning2015}.

Modern informed sampling-based planners instead improve search ordering over an implicit random geometric graph. BIT$^*$ combines batches of samples with heuristic graph search, while AIT$^*$ maintains an adaptive cost-to-go estimate through a reverse search \citep{gammellBatchInformedTrees2015a,strubAdaptivelyInformedTrees2020a}. These ideas are also relevant for risk-aware objectives, and we include them among the baselines studied in \cref{sec:setup} and \cref{sec:res}.

Our approach, while related, directly addresses the search for paths with low collision probability or severity-weighted risk. We define risk in its classical form as the product of probability and severity, acknowledging that the consequences of a collision can vary significantly depending on the context. For example, a collision that impedes operations in front of a hospital would be far more severe than one occurring in front of a pizza restaurant.

We introduce Risk Density RRT$^*$ (RD-RRT$^*$), an extension of the widely-used RRT$^*$ algorithm \citep{karamanSamplingbasedAlgorithmsOptimal2011}, designed for high-level planning in environments with both deterministic and uncertain obstacles. The core of our method is a computationally tractable, closed-form edge cost derived from the risk density functional introduced in our previous work \citep{paiolaEvaluationCollisionProbability2024a}. This allows for the efficient and robust integration of a continuous risk measure directly into the planner's cost function, enabling the search for paths with low approximated collision probability or low user-defined risk.

We validate RD-RRT$^*$ against geometric, chance-constrained, BIT$^*$, and AIT$^*$ baselines in \cref{sec:setup} and \cref{sec:res}. We also evaluate planned paths in a more realistic setting using an actuated finite-footprint TidyBot model in the MuJoCo simulator \citep{todorovMuJoCoPhysicsEngine2012,wuTidyBotOpenSourceHolonomic}.

The paper is organized as follows. \Cref{sec:prelim} provides the probabilistic background. \Cref{sec:riskRRT} formulates the planning objective and introduces RD-RRT$^*$, while \cref{sec:extensions} describes its modular components. \Cref{sec:setup} presents the geometric, informed-planner, and MuJoCo validation protocols, and \cref{sec:res} reports their results. \Cref{sec:disc} discusses the findings and limitations, and \cref{sec:concl} concludes the paper. Detailed distributional and three-dimensional extensions are retained in the appendices and summarized in the main experimental narrative.

\section{Background}
\label{sec:prelim}
The Risk Density index and its motivation were introduced in previous work \citep{paiolaEvaluationCollisionProbability2024a}. We briefly summarize here the main ideas, limiting the discussion to the case of a workspace contained in $\mathbb{R}^2$. We guide the reader to the original paper for more details and a derivation in $\mathbb{R}^3$.

\subsection{Notation and Conventions}
\label{sub:notation}

\paragraph{Vector and Matrix Notations.}
Vectors are denoted by bold lowercase letters, e.g., $\mathbf{x} \in \mathbb{R}^n$. The zero vector in $\mathbb{R}^n$ is denoted by $\mathbf{0}_n$. The Euclidean norm of a vector is written as $\|\mathbf{x}\|$. Matrices are represented by bold uppercase letters, e.g., $\mathbf{\Sigma}$. The determinant of a matrix is denoted by $|\mathbf{\Sigma}|$. The Hadamard (element-wise) product of two vectors or matrices of the same dimension is denoted by the symbol $\odot$.

\paragraph{Set and Geometric Notations.}
Sets are denoted by calligraphic notation. The robot's configuration space is denoted by $\mathcal{C} \subseteq \mathbb{R}^n$. The obstacle-free subset is $\mathcal{C}_{\text{free}} \subseteq \mathcal{C}$. Geometric sets are represented by uppercase letters, e.g., $A, B \subset \mathbb{R}^n$.
\begin{itemize}[leftmargin=*]
    \item The \textbf{Minkowski sum} of two sets $A$ and $B$ is the set of all vector sums of their elements:
    \[ A \oplus B := \{ \mathbf{a} + \mathbf{b} \mid \mathbf{a} \in A, \mathbf{b} \in B \}. \]
    \item The \textbf{Minkowski difference} is defined as the Minkowski sum with the reflection of the second set:
    \[ A \ominus B := A \oplus (-B) = \{ \mathbf{a} - \mathbf{b} \mid \mathbf{a} \in A, \mathbf{b} \in B \}. \]
\end{itemize}
A set $A$ translated by a vector $\mathbf{d} \in \mathbb{R}^n$ is written as $A + \mathbf{d}$. The Borel $\sigma$-algebra on $\mathbb{R}^n$ is denoted by $\mathcal{B}(\mathbb{R}^n)$.

\paragraph{Probabilistic and Other Notations.}
All random variables (RVs) are defined on a common probability space $(\Omega, \mathcal{F}, P)$ and capital notation is used for these, e.g. $X: \Omega \to \mathbb{R}$; we write $x=X(\omega)$ for a realization. The Borel $\sigma$-algebra on $\mathbb{R}^n$, denoted $\mathcal{B}(\mathbb{R}^n)$, is the collection of all sets in $\mathbb{R}^n$ for which a probability measure can be defined; this includes all open and closed sets. A random set is a random variable that takes values in $\mathcal{B}(\mathbb{R}^n)$.
A Gaussian (Normal) distribution with mean $\boldsymbol{\mu}$ and covariance $\mathbf{\Sigma}$ is denoted by $\mathcal{N}(\boldsymbol{\mu}, \mathbf{\Sigma})$, and its probability density function (PDF) is written as $\mathcal{N}(\mathbf{x} | \boldsymbol{\mu}, \mathbf{\Sigma})$. Statistical independence between two RVs $X$ and $Y$ is denoted by $X \perp Y$. The negation of a logical condition or event is denoted by an overline, e.g., $\overline{C}$ is the event that $C$ does not occur. Finally, the standard error function is denoted by $\erf(x)$.

\subsection{Formalizing Collision Events}
Let the robot and an obstacle be represented by sets $\mathcal{X}^d_R, \mathcal{X}^d_O \subset \mathbb{R}^2$ that define the space they occupy when centered at the origin, while $\mathcal{X}^d_R(\cdot), \mathcal{X}^d_O(\cdot): \mathbb{R}^2  \to \mathcal{B}(\mathbb{R}^2)$ are their translated versions. A collision occurs if their intersection is non-empty. This condition can be equivalently expressed using the Minkowski difference, which provides a convenient geometric interpretation: a collision occurs if and only if the origin is contained within the Minkowski difference of the robot and obstacle sets.

Formally, for a robot at configuration $\mathbf{x}_R$ and an obstacle at $\mathbf{x}_O$, the collision event $C$ is defined as:
\begin{equation}
    C(\mathbf{x}_R, \mathbf{x}_O) := \mathcal{X}^d_R(\mathbf{x}_R) \cap \mathcal{X}^d_O(\mathbf{x}_O) \neq \emptyset \iff \mathbf{0}_2 \in \mathcal{X}^d_R(\mathbf{x}_R) \ominus \mathcal{X}^d_O(\mathbf{x}_O).
    \label{eq:minkowski_collision}
\end{equation}
This formulation centralizes the collision check around a single point, the origin.

Let the robot $R$ and obstacle $O$ have uncertain positions. Since their exact configurations are unknown, we model the bodies as random sets, $\mathcal{X}_R, \mathcal{X}_O: \Omega \to \mathcal{B}(\mathbb{R}^2)$. Assuming only translational uncertainty, we can decompose their occupied space into a known, deterministic shape centered at the origin and an uncertain translational component. For a robot and obstacle with Gaussian position uncertainty, their configurations are given by the random variables $R \sim \mathcal{N}(\mathbf{x}_R, \mathbf{\Sigma}_R)$ and $O \sim \mathcal{N}(\mathbf{x}_O, \mathbf{\Sigma}_O)$.
The occupied space is then:
\begin{align}
    \mathcal{X}_{R} = \mathcal{X}^d_R + \mathbf{W}_{R}(\omega),\\
    \mathcal{X}_{O} = \mathcal{X}^d_O + \mathbf{W}_{O}(\omega)
\end{align}
where $\mathbf{W}_R$ and $\mathbf{W}_O$ are zero-mean Gaussian random variables representing the translational uncertainty of the corresponding objects. The difference between their positions, $\mathbf{d}_{RO} = \mathbf{x}_R - \mathbf{x}_O$, is also a random variable whose uncertainty is captured by the difference $\mathbf{W}_d = \mathbf{W}_R - \mathbf{W}_O$. The resulting distribution is also Gaussian, $\mathbf{W}_d \sim \mathcal{N}(\mathbf{0}, \mathbf{\Sigma}_d)$, where $\mathbf{\Sigma}_d = \mathbf{\Sigma}_R + \mathbf{\Sigma}_O$ due to the independence of the robot and obstacle uncertainties.

The probability of collision for this static configuration is the probability that the random vector $\mathbf{W}_d$ falls within the Minkowski difference of the shapes, translated by the nominal distance vector $\mathbf{d}_{RO}$. This is computed by integrating the Gaussian PDF over this target set:
\begin{equation}
    P(C(R,O)) = \int_{\mathcal{S}_d + \mathbf{d}_{RO}} \mathcal{N}(\mathbf{x}|\mathbf{0}, \mathbf{\Sigma}_d) d\mathbf{x},
    \label{eq:probsingleevent}
\end{equation}
where $\mathcal{S}_d = \mathcal{X}^d_R \ominus \mathcal{X}^d_O$.

\begin{remark}
    This derivation requires that $\mathbf{W}_d = \mathbf{W}_R - \mathbf{W}_O$ admit an analytical form. This property is not unique to the Normal distribution family; in principle, we only require $\mathbf{W}_R$ and $\mathbf{W}_O$ to be defined by the same family of stable distributions \citep{crooksFieldGuideContinuous2019}.
\end{remark}

\subsection{Collision Probability along a Continuous Path}
We now extend the analysis from a static configuration to a continuous path. Let the robot's nominal trajectory be parametrized by a continuous curve $\mathbf{x}_R(s): [0,1] \to \mathbb{R}^2$, where $s$ denotes the normalized progress. The distance vector between the robot and a static obstacle at $\mathbf{x}_O$ becomes a function of $s$, denoted $\gamma_d(s) = \mathbf{x}_R(s) - \mathbf{x}_O$.

Consequently, the integration domain in \eqref{eq:probsingleevent} also becomes a function of $s$:
\begin{equation}
    \mathcal{D}_d(s) = \mathcal{S}_d + \gamma_d(s).
    \label{eq:parametrizedintegrationset}
\end{equation}
For each point $s$ along the path, there is a corresponding collision event $C_s$, defined as the set of outcomes $\omega \in \Omega$ where a collision occurs:
\begin{equation}
    C_s := \{ \omega \in \Omega \mid \mathbf{W}_d(\omega, s) \in \mathcal{D}_d(s) \}.
    \label{eq:collisioneventparametrized}
\end{equation}
The probability of this instantaneous event, assuming the underlying stochastic process $\mathbf{W}_d$ is independent and identically distributed for each $s$, is given by:
\begin{equation}
    P(C_s) = \int_{\mathcal{D}_d(s)} \mathcal{N}(\mathbf{x}|\mathbf{0},\Sigma_d) d\mathbf{x}.
    \label{eq:prob_Cs}
\end{equation}
The total collision event along the path, $C$, is the union of all such instantaneous events:
\begin{equation}
    C := \bigcup_{s \in [0,1]} C_s.
    \label{eq:unionofallprobabilities}
\end{equation}
Directly computing the probability $P(C)$ is generally intractable due to the non-trivial interdependence between the events $C_s$ for different values of $s$. To proceed, we must make simplifying assumptions about the nature of this interdependence.

\subsection{Hypotheses on Event Interdependence}
\begin{figure}
    \includegraphics[clip,width=0.9\linewidth]{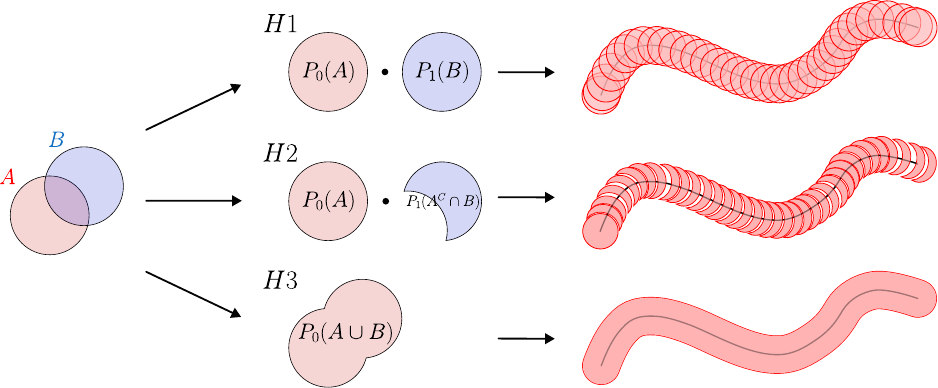}
    \caption{Visual representation of how considered assumptions work on distinct events. \textbf{Left:} The events $\{\omega \in \Omega \mid w_d(\omega,s_0)\in A\}$ and $\{\omega \in \Omega \mid w_d(\omega,s_1)\in B\}$, where the sets $A$ and $B$ are shown as red and blue disks. $P_0(\cdot)$ and $P_1(\cdot)$ stand respectively for the probability measure of the first and the second random variable to be inside a set. \textbf{Right:} Graphical illustration of the collision events considered respectively in \textbf{(H1)} \cref{eq:probH1}, \textbf{(H2)} \cref{eq:PH2abs} and \textbf{(H3)} \cref{eq:probH3}.}
    \label{fig:H1H2H3}
\end{figure}

As detailed in previous work \citep{paiolaEvaluationCollisionProbability2024a}, the relationship between collision events along a path can be modeled under several distinct hypotheses, which span a spectrum from complete independence to complete dependence. A summary of these models is depicted in \cref{fig:H1H2H3}.

\begin{enumerate}[leftmargin=1cm]
    \item[\textbf{(H1)}] \textbf{Complete Independence.} This is the most common simplification, assuming that the collision event at any point $s$ is statistically independent of the event at any other point $q \neq s$:
    \begin{equation*}
        \text{\textbf{(H1)}}: C_s \perp C_q, \quad \forall s \neq q \in [0,1].
    \end{equation*}
    Under this assumption, the total probability of no collision is the product of the probabilities of no collision at each point. This leads to the formulation:
    \begin{equation}
        P_{H1}(C) = 1 - \prod_{s \in [0,1]} (1 - P(C_s)).
        \label{eq:probH1}
    \end{equation}
    This model is intuitive for discrete waypoints but is ill-suited for continuous paths. As the number of points in the discretization of $[0,1]$ approaches infinity, $P_{H1}(C)$ invariably converges to 1, rendering the approximation useless for continuous settings.

    \item[\textbf{(H2)}] \textbf{Markov Property on Locally Swept Area.} A less restrictive model assumes a specific form of the Markov property, akin to a \enquote{settling dust} scenario. In this view, if a robot retraces its steps or loops, it re-accumulates risk. 

    To formalize this, consider a sequence of partitions of the path interval $\mathcal{P}_n = \{0=s_0 < s_1 < \dots < s_{N}=1\}$ such that the mesh size $\|\mathcal{P}_n\| \to 0$ as $n \to \infty$. Similarly to \cref{eq:unionofallprobabilities}, the collision event over a sub-interval $[s_i, s_{i+1}]$ is:
    \begin{equation}
        C_{[s_i, s_{i+1}]} = \bigcup_{\tau \in [s_i, s_{i+1}]} C_\tau.
    \end{equation}
    We assume that this event is conditionally independent of the trajectory prior to $s_i$, given only that the robot is safe at the instant $s_i$. The conditional probability is then determined by the \textit{local} volume swept during that interval:
    \begin{equation}
        P(C_{[s_i, s_{i+1}]} \mid \overline{C}_{s_i}) \approx P\left(\mathbf{W}_d(s_{i+1}) \in (\mathcal{D}_d(s_{i+1}) \setminus \mathcal{D}_d(s_i))\right).
        \label{eq:PH2local}
    \end{equation}
    The total probability of survival is the limit of the product of survival probabilities over the partition:
    \begin{equation}
        P_{H2}(C) = 1 - \lim_{\|\mathcal{P}\| \to 0} \prod_{i=0}^{N-1} \left(1 - P(C_{[s_i, s_{i+1}]} \mid \overline{C}_{s_i})\right). \label{eq:PH2abs}
    \end{equation}
    Crucially, because the set difference in \cref{eq:PH2local} is strictly local (relative only to the immediate predecessor $s_i$), overlapping paths contribute to the risk multiple times.

    \item[\textbf{(H3)}] \textbf{Stopped Process.} This hypothesis sits at the opposite extreme, assuming all uncertainty is resolved at the initial time $s_0 = 0$. The stochastic process describing the uncertainty is \enquote{stopped}, meaning its value remains constant for all $s > s_0$:
    \begin{equation*}
        \text{\textbf{(H3)}}: \mathbf{W}_d(\omega, s) := \mathbf{W}_d(\omega, \min(s_0, s)).
    \end{equation*}
    This model is appropriate for scenarios with static but unknown obstacles, where the uncertainty is in the initial knowledge of the environment. The total collision probability is the probability that the single random outcome causes a collision anywhere along the path. This corresponds to integrating the PDF over the union of all integration domains, a set known as the swept volume $\mathcal{D}_T$:
    \begin{equation}
        \label{eq:probH3}
        P_{H3}(C) = \int_{\mathcal{D}_T} \mathcal{N}(\mathbf{x}|\mathbf{0}, \Sigma_d) d\mathbf{x}, \quad \text{where} \quad \mathcal{D}_T = \bigcup_{s \in [0,1]} \mathcal{D}_d(s).
    \end{equation}
\end{enumerate}

\subsection{The Risk Density Functional}
While \textbf{(H1)} is unsuitable for continuous paths, both \textbf{(H2)} and \textbf{(H3)} can be formulated as integrals. Let us assume the robot and obstacle shapes are outer-bounded by balls, making their Minkowski difference $\mathcal{S}_d$ a disk of radius $r$. The swept volume $\mathcal{D}_T$ under \textbf{(H3)} becomes a tube-like shape, as shown in \Cref{fig:H1H2H3}. A general, though not always diffeomorphic, parametrization for this tube is given by the track coordinates:
\begin{equation}
    \Phi(s,\theta) = \mathbf{x}_R(s) + \theta \begin{bmatrix} 0 & -1 \\ 1 & 0 \end{bmatrix} \frac{d\mathbf{x}_R}{ds},
    \label{eq:areaparametrization}
\end{equation}
where $\theta \in [-r, r]$ is the orthogonal distance from the path's centerline. Using this, the probability under \textbf{(H3)} can be written as a double integral:
\begin{equation}
    P^a_{H3}(C) = \int_0^1 \int_{-r}^{r} \mathcal{N}(\Phi_d(s,\theta)|\mathbf{0},\Sigma_d)\Gamma(s,\theta) d\theta ds,
    \label{eq:intprobH3}
\end{equation}
where $\Phi_d(s,\theta) = \Phi(s,\theta) - \mathbf{x}_O$ and $\Gamma = |\text{det}(\nabla \Phi_d)|$ is the Jacobian determinant. 
Similarly, using the same parametrization, \cref{eq:PH2local} is rewritten as 
\begin{equation}
    \label{eq:PH2slice}
    \lim_{\Delta s \to 0} P^a(C_{[s, s + \Delta s]} \mid \overline{C}_{s}) = \int_{s}^{s+\Delta s}\int_{-r}^{r} \mathcal{N}(\Phi_d(u,\theta)|\mathbf{0},\Sigma_d)\Gamma(u,\theta) d\theta d u,
\end{equation}
which intuitively is the \enquote{moonlike slice} shown in the \textbf{(H2)} depiction in \cref{fig:H1H2H3}. 

Then, by plugging \cref{eq:PH2slice} into \cref{eq:PH2abs} and recognizing it as a product integral, its continuous form is written as
\begin{equation}
    P^a_{H2}(C) = 1-\exp\left(-\int_0^1\int_{-r}^{r} \mathcal{N}(\Phi_d(s,\theta)|\mathbf{0},\Sigma_d)\Gamma(s,\theta) d\theta d s\right).
    \label{eq:intprobH2}
\end{equation}
A key insight arises when we consider the sensitivity of these probabilities to the radius $r$ of the combined disk. 

Define the sensitivity functional $S_r(\gamma_d, r)$ as the derivative of the probability with respect to $r$:
\begin{equation}
     S_r^{H2/H3}(\gamma_d, r) := \frac{dP^a_{H2/H3}}{dr}.
     \label{eq:sensitivities}
\end{equation}
It can be shown that in the limit as $r \to 0$, the sensitivities under both hypotheses converge to the same functional:
\begin{equation}
    S_r^{H2}(\gamma_d, 0) = S_r^{H3}(\gamma_d, 0).
\end{equation}
We call this resulting functional the \emph{risk density}, $r_d(\gamma_d)$. It represents the first-order sensitivity of collision probability to changes in the object dimensions and is given by the line integral:
\begin{equation}
    r_d(\gamma_d) = 2\int_{0}^{1} \mathcal{N}(\gamma_d(s)|\mathbf{0},\Sigma_d)\left\|\frac{d \gamma_d}{d s}\right\| ds.
    \label{eq:riskdensity}
\end{equation}
This theoretical connection shows that for infinitesimally thin objects, the fundamentally different physical assumptions of \textbf{(H2)} and \textbf{(H3)} lead to the same measure of risk. This allows us to form a first-order Taylor approximation of the collision probability $P(C)$ that is valid under either assumption:
\begin{equation}
    \label{eq:taylorRD}
    P_a(C,r) \approx r_d(\gamma_d) r.
\end{equation}

For a set $\mathbb{O}$ of multiple uncertain obstacles, let $C_i$ denote the event that the robot collides with obstacle $O_i\in\mathbb{O}$. We assume that these obstacle-specific collision events are mutually independent:

\begin{equation}
    \label{eq:conditionalindependence}
    C_i \perp\!\!\!\perp C_j, \qquad \forall i\neq j.
\end{equation}

After some mathematical bookkeeping, detailed in previous work \citep{paiolaEvaluationCollisionProbability2024a}, we recover the familiar approximation
\begin{equation}
    P_a(C) = \sum_{j \in \text{ind}(\mathbb{O})} r_{d_j}(\gamma_d) r_j = \mathbf{r}_d^{\mathbb{O}}(\gamma_d)^\top \overline{\mathbf{r}},
    \label{eq:rdformultiple}
\end{equation}
where $\mathbf{r}_d^{\mathbb{O}}: C^1(\mathbb{R}^2) \to \mathbb{R}^{|\mathbb{O}|}$ is the vector form of the functional \eqref{eq:riskdensity} and $\overline{\mathbf{r}} \in \mathbb{R}^{|\mathbb{O}|}$ is the vector of combined radii.

As a first-order additive approximation, \eqref{eq:rdformultiple} is not intrinsically bounded by one. Values above one represent saturation of the corresponding physical probability rather than a probability greater than one; the implementation retains the raw additive value as the planning cost so that high-cost paths remain distinguishable.

We refer the reader to previous work \citep{paiolaEvaluationCollisionProbability2024a} for a more complete derivation of all the quantities discussed here.

\begin{remark}
In practical scenarios, the assumption in \cref{eq:conditionalindependence} maps to obstacles whose positions are independent of each other, i.e., physically separated objects, although their bounding balls may overlap. While the formulation supports decomposing a larger shape into multiple disks to maintain morphological characteristics, e.g., maintaining a corner shape instead of over-bounding it, \cref{eq:conditionalindependence} states that the decomposed obstacle will then act as multiple independent pieces instead of a coherent shape. This effect is partially studied in \cref{app:experiments}.
\end{remark}

\subsubsection{Approximation Characteristics}
The approximation in \eqref{eq:taylorRD} carries a clear geometric intuition. The risk density functional, $r_d(\gamma_d)$, can be interpreted as a line integral of the collision probability density along the path $\gamma_d$, weighted by the path's differential arc length $\|\frac{d \gamma_d}{d s}\|$. This functional yields a scalar value representing a \enquote{risk per unit of radius}. Multiplying this density by the specific radius $r$ of the combined object provides a first-order approximation of the total probability mass contained within a \enquote{tube} of radius $r$ centered on the path. This is the continuous analog to approximations that estimate collision probability of singular configurations by multiplying a PDF value by a small area at discrete points along a trajectory \citep{lambertFastMonteCarlo2008}.
 
Moreover, the approximation in \eqref{eq:rdformultiple} depends only on the intrinsic geometry of the problem (the path and the uncertainty models) and requires no external discretization parameters. 

\section{Risk Density \texorpdfstring{RRT$^*$}{RRT*}}
\label{sec:riskRRT}
\subsection{Problem Formulation}
Consider a robot whose configuration can be described by a point in a configuration space $\mathcal{C}$.
A set of static obstacles renders a set of robot configurations $\mathcal{C}_{obst}\subseteq \mathcal{C}$ infeasible. The robot position is known with sufficient precision to define, by subtraction, a static-obstacles-free region $\mathcal{C}_{free} = \mathcal{C}\setminus\mathcal{C}_{obst}$.
Assume, however, that a second set of uncertain obstacles $\tilde{\mathcal{C}}_{obst}(\omega)$ is present in the scenario, i.e., the uncertain set formed by the elements of $\mathbb{O}$.

Denoting $\mathbf{x}_{0}\in\mathcal{C}_{free}$ as the initial robot position and $\mathcal{X}_{goal}\subseteq\mathcal{C}_{free}$ as a goal region, the objective of collision-probability-minimizing path planning is to find a path $\gamma_d:[0,1]\to \mathbb{R}^2$ that satisfies the constraints
\begin{align}
    \label{eqs:pathplanningconstraints}
    & \gamma_d(0)=\mathbf{x}_{0}, \ \gamma_d(1)\in \mathcal{X}_{goal} \\
    & \gamma_d(s)\in\mathcal{C}_{free},\ \forall s\in[0,1] &
\end{align}
while minimizing the probability of collision $P(C)$.

Our approach relies on approximating the collision probability $P(C)$ with \eqref{eq:rdformultiple}, where $C$ is \eqref{eq:unionofallprobabilities} for $\tilde{\mathcal{C}}_{obst}(\omega)$. This, as shown in the next subsection, lets us efficiently compute a proxy for $P(C)$ along the edges of a graph: an advantage that we leverage in \cref{sub:algorithm} to propose an extension to the RRT$^*$ algorithm \citep{karamanSamplingbasedAlgorithmsOptimal2011}, denoted \emph{Risk Density RRT$^*$} (RD-RRT$^*$).

\begin{remark}
    Notice how our problem definition does not concern the usual geometric cost, such as the length of $\gamma_d$ or the time taken to reach $\mathcal{X}_{goal}$, but instead focuses on searching for paths with low collision probability. 
\end{remark}

\begin{table}[ht]
\centering
\caption{Assumptions of the Problem.}
\label{tab:assumptions}
\begin{tabular}{lp{8cm}}
\hline
\textbf{Category} & \textbf{Assumption} \\ \hline
\textbf{Space} & The task space maps directly to the Configuration Space (Euclidean). No complex manifold constraints are considered. \\
\textbf{Dynamics} & The system is purely kinematic with no dynamic constraints. \\
\textbf{Obstacles} & Obstacles are outer-bounded by balls. \\
\textbf{Independence} & Obstacle distributions are conditionally independent of each other \cref{eq:conditionalindependence}. \\
\textbf{Distributions} & Uncertainty distributions must belong to a stable family (e.g., Gaussian, Cauchy). \\
\textbf{Robot State} & Robot position uncertainty is considered negligible and is ignored during planning. \\
\textbf{Interdependence} & Collision events follow either the H2 or H3  hypothesis. \\ \hline
\end{tabular}
\end{table}

\subsection{A Computationally Tractable Edge Cost}
\label{sub:closed_form}
The practical application of RD-RRT$^*$ hinges on the ability to efficiently compute the cost for a vast number of potential edges in the search graph. The integral form of the risk density \eqref{eq:riskdensity} is not directly suitable for this high-frequency task. However, for a straight-line path segment, which forms the edges in our RRT$^*$ graph, this integral admits a closed-form solution.

Consider a straight-line path $\gamma_d^i(s)$ from a state $\mathbf{x}_i$ to $\mathbf{x}_{i+1}$ relative to an obstacle $O_j$ at nominal position $\mathbf{x}_O^j$. The path is parametrized over $s \in [0, s_f]$ as:
\[
\gamma_d^i(s) = (\mathbf{x}_i - \mathbf{x}_O^j) + s(\mathbf{x}_{i+1} - \mathbf{x}_i).
\]
To solve the integral in \eqref{eq:riskdensity}, we analyze its two main components. First, the differential arc length is the magnitude of the path derivative, which for a straight line is constant:
\[
\left\|\frac{d \gamma_d^i}{d s}\right\| = \|\mathbf{x}_{i+1} - \mathbf{x}_i\|.
\]
Second, the argument of the Gaussian PDF, $\mathcal{N}(\gamma_d^i(s)|\mathbf{0},\Sigma_d^j)$, contains a quadratic form in its exponent, $-\frac{1}{2} (\gamma_d^i(s))^\top (\Sigma_d^j)^{-1} \gamma_d^i(s)$. Substituting the path parametrization yields a simple quadratic polynomial in the variable $s$:
\[
(\gamma_d^i(s))^\top (\Sigma_d^j)^{-1} \gamma_d^i(s) = cs^2 + 2bs + a,
\]
where the coefficients $a, b, c$ depend only on the segment's endpoints and the obstacle's uncertainty:
\begin{align*}
    a &= (\mathbf{x}_i - \mathbf{x}_O^j)^\top (\Sigma_d^j)^{-1} (\mathbf{x}_i - \mathbf{x}_O^j), \\
    b &= (\mathbf{x}_i - \mathbf{x}_O^j)^\top (\Sigma_d^j)^{-1} (\mathbf{x}_{i+1} - \mathbf{x}_i), \\
    c &= (\mathbf{x}_{i+1} - \mathbf{x}_i)^\top (\Sigma_d^j)^{-1} (\mathbf{x}_{i+1} - \mathbf{x}_i).
\end{align*}
With these substitutions, the risk density integral becomes the integral of a function proportional to a 1D Gaussian PDF, which can be solved analytically using the error function, $\erf(\cdot)$. For a single obstacle $O_j$, the risk density for the segment is:
\begin{equation}
    \label{eq:closedformRD}
    r_d^j(\gamma_d^i)|_0^{s_f} = \frac{\|\mathbf{x}_{i+1} - \mathbf{x}_i\|}{\sqrt{2\pi c |\Sigma_d^j|}} e^{\frac{b^2-ac}{2c}} \left[ \erf\left(\frac{b+cs_f}{\sqrt{2c}}\right) - \erf\left(\frac{b}{\sqrt{2c}}\right) \right].
\end{equation}
This closed-form solution allows for the rapid forward computation of edge costs.

Consequently, using \eqref{eq:rdformultiple} with $s_f=1$, for each segment $\gamma_d^i$ the probability can be approximated with 
\begin{equation}
    \label{eq:probabilityeachsegment}
    P(C_{\gamma_d^i}) \approx \mathbf{r}_d^{\mathbb{O}}(\gamma_d^i)^\top \overline{\mathbf{r}}.
\end{equation}
Because \eqref{eq:riskdensity} is a linear functional, any path from $\mathbf{x}_0$ to a leaf $\mathbf{x}_n$ can be expressed as the concatenation of segments $\gamma_d^i$ for $i=0,\ldots,n-1$. Therefore, \eqref{eq:rdformultiple} for a path $\gamma_d$ is
\begin{equation}
    \label{eq:probabilityconcatenationpaths}
    P(C)\approx P(C_{\gamma_d}) = \sum_{i=0}^{n-1} \mathbf{r}_d^{\mathbb{O}}(\gamma_d^i)^\top \overline{\mathbf{r}}.
\end{equation}

\subsection{Core Algorithm and Primitives}
\label{sub:algorithm}
The general structure of RD-RRT$^*$ traces the outline of the RRT$^*$ algorithm \citep{karamanSamplingbasedAlgorithmsOptimal2011}, modifying its core subroutines to operate on the probabilistic cost function defined by \eqref{eq:probabilityconcatenationpaths}. Our implementation, written in Julia \citep{Julia-2017}, was initially an extension of \texttt{MotionPlanning.jl} \citep{jansonMonteCarloMotion2015}, but has since been modernized and is now self-contained. The code can be found at \url{https://github.com/IIT-SoftBots/RiskPlanning}.

A single iteration of the main algorithm loop, shown in \cref{alg:RDRRT*}, consists of sampling a random state, finding the nearest node in the existing tree, steering towards the sample to create a new state, and then performing the \texttt{ChooseParent} and \texttt{Rewire} operations to integrate the new state in a cost-optimal way. The key primitives and subroutines are detailed below.

\begin{algorithm}
\caption{RD-RRT$^*$ Main Loop}\label{alg:RDRRT*}
        \KwData{$x_{0}$, $N$, $\epsilon_{steer}$, $\epsilon_{n}$, $b$, $\mathbb{O}$, Sampler, Finder}
        \KwResult{Tree $\mathcal{T}$}
        $\mathcal{T}\leftarrow\texttt{InitializeTree}(x_0)$\;
        $\text{Finder}.\texttt{Build}(\mathcal{T})$\;
        \For{$i=1,...,N$}{
            $x_{rand}\leftarrow\text{Sampler}.\texttt{Sample}(\mathcal{T})$\;
            $x_{nearest}\leftarrow\text{Finder}.\texttt{Nearest}(\mathcal{T}, x_{rand})$\;
            $x_{new}, c_{edge} \leftarrow \texttt{RDSteer}(x_{nearest}, x_{rand}, \epsilon_{steer}, b)$\;
            \If{\texttt{CollisionFree}$(x_{nearest}, x_{new})$}{
                \tcp{Find neighbors in a fixed Euclidean radius (heuristic)}
                $X_{near} \leftarrow \text{Finder}.\texttt{NeighborsInRange}(\mathcal{T}, x_{new}, \epsilon_n)$\;
                $x_{min} \leftarrow \texttt{ChooseParent}(X_{near}, x_{new}, x_{nearest}, \epsilon_{steer}, b, \mathcal{T})$\;
                $\mathcal{T}\leftarrow\texttt{InsertNode}(\mathcal{T}, x_{new}, x_{min})$\;
                $\text{Finder}.\texttt{Update}(\mathcal{T}, x_{new})$\;
                $\mathcal{T}\leftarrow\texttt{Rewire}(X_{near}, x_{new}, \epsilon_{steer}, b, \mathcal{T})$\;
            }
        }
        \Return $\mathcal{T}$
\end{algorithm}

\subsubsection{Primitives.}
\begin{itemize}[leftmargin=*]
    \item \texttt{Cost} and \texttt{c}: The \texttt{Cost} function computes the cumulative cost $P(C_{\gamma_d})$ of a path from the root node $\mathbf{x}_0$ to a given node $\mathbf{x}_i$ using \eqref{eq:probabilityconcatenationpaths}. The incremental function \texttt{c} computes the cost of a single edge using \eqref{eq:probabilityeachsegment}. Both routines depend on the set of uncertain obstacles $\mathbb{O}$, which summarizes the collection of parameters $[\mathbf{x}_O^j, \Sigma_d^j, r_j]$ for each obstacle.
    
    \item \texttt{Sample}: This procedure,  handled by a modular \texttt{Sampler} component, outputs a sample state $\mathbf{x}_{rand}$ from the collision-free space $\mathcal{C}_{free}$ in a pseudo-random way. The strategies for sampling are discussed in \cref{sec:extensions}.
    
    \item \texttt{CollisionFree}: This standard routine checks for collisions with \emph{deterministic} obstacles along a straight-line segment and returns \texttt{true} if the segment is in $\mathcal{C}_{free}$.
    
    \item \texttt{RDNear}: Conceptually, this primitive handles all spatial queries related to the tree. In our modular implementation, this responsibility is delegated to a \texttt{Finder} component, which performs two distinct operations:
    \begin{itemize}[leftmargin=1cm]
        \item \textbf{Nearest Neighbor:} Finding the single node $x_{nearest}$ in the tree that is closest (in Euclidean distance) to a given sample $\mathbf{x}_{rand}$.
        \item \textbf{Neighbor Search:} Finding the set of all neighboring nodes $X_{near}$ for potential connection and rewiring. Formally, this set is defined as $X_{near} := \{ \mathbf{x} \in \mathcal{T} \mid \|\mathbf{x} - \mathbf{x}_{new}\| \leq \epsilon_{n} \}$, where $\epsilon_n$ is a search radius. The use of a Euclidean ball for this search is a key heuristic, as discussed in \cref{subsec:theory_and_heuristics}.
    \end{itemize}

    \item \texttt{RDSteer}: This function generates a new state $x_{new}$ by taking a step from an existing node $x_{near}$ towards a sample $x_{rand}$. The step is constrained to fulfill both a maximum Euclidean length, $\|\mathbf{x}_{new} - \mathbf{x}_{near}\| < b$, and a maximum probabilistic cost, $\texttt{c}(\mathbf{x}_{near}, \mathbf{x}_{new}) < \epsilon_{steer}$. This requires solving a Boundary Value Problem (BVP), as detailed in \cref{subsec:theory_and_heuristics}.
\end{itemize}

\subsubsection{Subroutines.} The core logic of the RRT$^*$ algorithm is encapsulated in the \texttt{ChooseParent} and \texttt{Rewire} subroutines, shown in \cref{alg:chooseRD} and \cref{alg:rewireRD}.
\begin{itemize}[leftmargin=*]
    \item \texttt{ChooseParent}: For a new state $\mathbf{x}_{new}$, this subroutine iterates through the neighborhood $X_{near}$. For each neighbor, it computes the total path cost if that neighbor were to be the parent of $\mathbf{x}_{new}$. It returns the neighbor that results in the lowest cumulative cost.

    \item \texttt{Rewire}: After a new node $\mathbf{x}_{new}$ has been added to the tree, this subroutine examines each of its neighbors in $X_{near}$. It considers whether routing the path to a neighbor \emph{through} $\mathbf{x}_{new}$ would result in a lower total cost. The cost of this potential new edge is found by a direct evaluation, $\texttt{c}(\mathbf{x}_{new}, \mathbf{x}_{neighbor})$, as both nodes already exist in the tree. If a better path is found, the neighbor's parent is updated to be $\mathbf{x}_{new}$, improving the overall quality of the solution tree.
\end{itemize}

\begin{algorithm}[t]
\caption{ChooseParent}\label{alg:chooseRD}
        \KwData{$X_{near}$, $x_{new}$, $x_{nearest}$ , $\epsilon_{steer}$, $b$, $\mathcal{T}$}
        \KwResult{$x_{min}$}
        $x_{min}\leftarrow x_{nearest}$\;
        $c_{min}\leftarrow \texttt{Cost}(x_{nearest}, \mathcal{T}) + \texttt{c}(x_{nearest}, x_{new})$\;
            \For{$x_{near} \in X_{near}$}{
                \If{\texttt{CollisionFree}$(x_{near}, x_{new})$}{
                    $c_{new}\leftarrow \texttt{Cost}(x_{near}, \mathcal{T}) + \texttt{c}(x_{near}, x_{new})$\;
                    \If{$c_{new} < c_{min}$}{
                        $c_{min}\leftarrow c_{new}$\;
                        $x_{min}\leftarrow x_{near}$\;
                    }
                }
            }
        \Return $x_{min}$
\end{algorithm}

\begin{algorithm}[t]
\caption{Rewire}\label{alg:rewireRD}
        \KwData{$X_{near}$, $x_{new}$, $\epsilon_{steer}$, $b$, $\mathcal{T}$}
        \KwResult{Updated $\mathcal{T}$}
        \For{$x_{near} \in X_{near}$}{
            \If{\texttt{CollisionFree}$(x_{new}, x_{near})$}{
                \If{$\texttt{Cost}(x_{new}, \mathcal{T}) + \texttt{c}(x_{new}, x_{near}) < \texttt{Cost}(x_{near}, \mathcal{T})$}{
                    $\mathcal{T}\leftarrow \texttt{UpdateParent}(x_{near}, x_{new})$ 
                }
            }
        }
        \Return $\mathcal{T}$
\end{algorithm}

\begin{algorithm}[t]
    \caption{\texttt{RDSteer}}\label{alg:steerprob}
    \KwData{$x_{from}$, $x_{to}$, $\epsilon_{steer}$, $b$}
    \KwResult{$x_{new}, c_{edge}$}
    \uIf{$\texttt{c}(x_{from}, x_{to}) > \epsilon_{steer} \text{ or } \|x_{from} - x_{to}\| > b$}{
        $x_{new}, c_{edge} \leftarrow \texttt{SolveBVP}(x_{from}, x_{to}, \epsilon_{steer}, b)$\;
    }
    \uElse{
        $x_{new} \leftarrow x_{to}$\;
        $c_{edge} \leftarrow \texttt{c}(x_{from}, x_{to})$\;
    }
    \Return $x_{new}, c_{edge}$
\end{algorithm}

\subsection{Cost Function Variants}
The flexibility of the framework allows the cost function \eqref{eq:probabilityeachsegment} to be easily modified to achieve different planning objectives.

\subsubsection{Risk-Based \texorpdfstring{RRT$^*$}{RRT*}}
Inspired by the classical definition of \emph{risk} = \emph{probability} $\times$ \emph{severity} \citep{ISO310002018en}, we introduce a positive scalar $\xi_j$ to denote the severity of colliding with each specific obstacle $O_j$. This allows the user to express preferences, for instance, by assigning a higher severity to a collision with a critical piece of infrastructure compared to a generic obstacle. We modify the cost \eqref{eq:probabilityeachsegment} to be:
\begin{equation}
    \label{eq:severitycost}
    \texttt{c}_{r}(\gamma_d^i) = \mathbf{r}_d^{\mathbb{O}}(\gamma_d^i)^{\top} (\overline{\mathbf{r}} \odot \overline{\boldsymbol{\xi}}),
\end{equation}
where $\overline{\boldsymbol{\xi}}\in \mathbb{R}_+^{|\mathbb{O}|}$ is the vector of all severities. All subroutines are modified to use this cost.

\subsubsection{Hybrid Risk-Length \texorpdfstring{RRT$^*$}{RRT*}}
To regularize the algorithm and prevent the generation of solutions that circumvent stochastic obstacles via arbitrarily long paths, we propose a variant that creates a hybrid cost function. This cost is a weighted sum of the original probability cost and the Euclidean path length:
\begin{equation}
    \texttt{c}_h(x_i, x_{i+1}, \lambda) = \texttt{c}(x_i, x_{i+1}) + \lambda \|x_i - x_{i+1}\|.
\end{equation}
All subroutines, including the cost check within the \texttt{RDSteer} function, are updated to use this new cost function $\texttt{c}_h$. 

\subsection{Theoretical Properties and Implementation Heuristics}
\label{subsec:theory_and_heuristics}

The RD-RRT$^*$ algorithm is designed within the structural framework of RRT$^*$ \citep{karamanSamplingbasedAlgorithmsOptimal2011}, which offers guarantees of asymptotic optimality. However, this guarantee hinges on the cost function satisfying specific properties that are implicitly assumed by the RRT$^*$ proof. We discuss these properties and the necessary heuristic choices for a feasible implementation.

\subsubsection{The Semimetric Nature of Risk.}
Consider the endpoint function obtained by evaluating \eqref{eq:probabilityconcatenationpaths} only on the straight segment connecting two states. This direct straight-edge cost is non-negative and symmetric, and it is strictly positive between distinct states for non-degenerate Gaussian distributions. It can, however, violate the triangle inequality: a direct segment through a high-uncertainty region can cost more than two segments forming a longer, safer detour. Under the convention used here, this straight-edge endpoint function is therefore a \textbf{semimetric}. This statement concerns the local connection evaluated by the implementation, rather than the minimum path distance obtained by optimizing over every possible connecting curve.

\begin{figure}[t!]
    \centering
    \begin{subfigure}{0.4\textwidth}
        \centering
        \includegraphics[width=\linewidth]{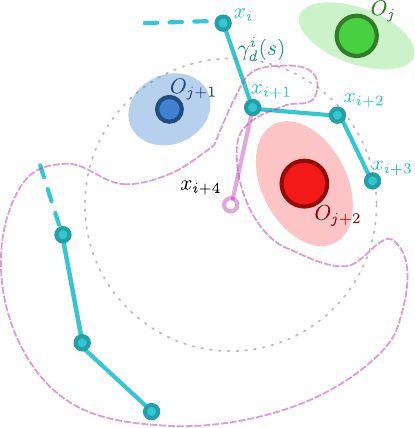} %
        \caption{Heuristic Neighbor Search}
        \label{sfig:heuristic_mismatch}
    \end{subfigure}
    \hfil
    \begin{subfigure}{0.48\textwidth}
        \centering
        \includegraphics[width=\linewidth]{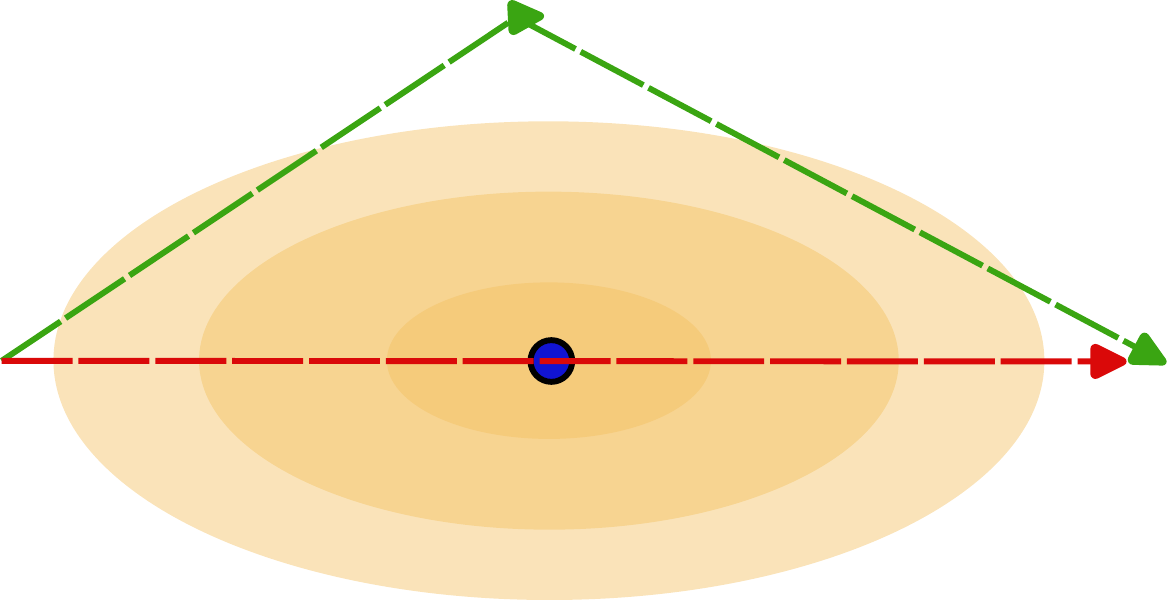} %
        \caption{The Semimetric Property}
        \label{sfig:semimetric}
    \end{subfigure}
    \caption{(\textbf{a}) The \texttt{RDNear} routine searches for neighbors of a new node (e.g., $x_{i+4}$) within a Euclidean ball (dashed gray line). This neighborhood may exclude a useful parent that is farther away in Euclidean distance but lies inside a low-cost sublevel set of the direct straight-edge RD cost (dashed violet line). Such a sublevel set can be non-convex or disconnected in the Euclidean state space. (\textbf{b}) The direct straight edge (red) through the high-risk region has greater RD cost than the longer detour (green), so the direct straight-edge endpoint cost violates the triangle inequality: $c(\text{red}) > c(\text{green})$.}
    \label{fig:theory_and_heuristics}
\end{figure}

\subsubsection{Consequences for Optimality.}
The distinction between the direct straight-edge cost and the induced path distance is important. The endpoint cost evaluated on a prescribed straight segment is a semimetric under the convention used here because it can violate the triangle inequality. By contrast, taking the infimum of the RD functional over all curves connecting two states produces an induced path distance that satisfies the triangle inequality by path concatenation; for a continuous strictly positive risk density on a bounded domain, this is a weighted geodesic metric.

The standard RRT$^*$ proof covers a robust optimal path using Euclidean state-space balls, not sublevel sets of the objective \citep{karamanSamplingbasedAlgorithmsOptimal2011}. Our practical difficulty is therefore not the compactness of an RD-cost ball. It is that the Euclidean neighborhoods used by the implementation need not match the potentially non-convex sublevel sets of the direct straight-edge RD cost. This mismatch can affect finite-sample parent selection and rewiring. We do not establish an asymptotic-optimality result here for the combination of the modified objective, steering rule, and implemented radius schedules.

This theoretical impasse forces us to adopt computationally feasible heuristics:
\begin{itemize}[leftmargin=*]
    \item \textbf{Euclidean Neighbor Search:} The \texttt{RDNear} subroutine searches for neighbors within a \textbf{Euclidean ball} of radius $\epsilon_n$, not a cost-space ball. This creates a mismatch between the search geometry and the optimization objective, meaning the \texttt{Rewire} step is not guaranteed to find the truly optimal connections.
    \item \textbf{Radius Schedule:} The standard decaying Euclidean radius, $\epsilon_n \propto (\log(n)/n)^{1/d}$, can be applied as a heuristic schedule. However, using it does not by itself transfer the original RRT$^*$ asymptotic-optimality guarantee to the modified objective and steering procedure considered here. Our default implementation instead employs a fixed Euclidean radius, which is likewise heuristic in this setting.
\end{itemize}

\subsubsection{The \texttt{RDSteer} Inverse Problem.}
To enforce the maximum edge cost $\epsilon_\text{steer}$, the algorithm must solve an inverse problem: finding the fractional distance $s_f \in [0,1]$ along a segment that yields a cost exactly equal to $\epsilon_\text{steer}$. This requires root-finding on the equation:
\[
\texttt{c}(\mathbf{x}_\text{from}, \mathbf{x}_\text{from} + s_f(\mathbf{x}_\text{to} - \mathbf{x}_\text{from})) - \epsilon_\text{steer} = 0.
\]
This is a frequent operation, for which we use Halley's method \citep{pal2024nonlinearsolve} for its fast (cubic) convergence.

\begin{remark}
    The Halley implementation uses automatic differentiation tools to programmatically obtain the gradients of the cost function; analytically computing the derivatives could improve performance. While this is not the main computational bottleneck, a maximum number of iterations is supplied to the solver; when exceeded, the last feasible $s$ is returned.
\end{remark}

\subsubsection{Rewiring Schedule.}
As previously discussed, we do not establish that the standard RRT$^*$ optimality guarantee carries over to the modified objective, steering rule, and radius schedules used here. While our default implementation uses a fixed Euclidean radius for simplicity, we also evaluate the standard shrinking schedule as a heuristic, together with other alternatives, to understand their practical behavior. We consider four schedules for a given base radius $\epsilon_n$ in a $d$-dimensional space:
\begin{itemize}[leftmargin=*]
    \item \texttt{Constant}: A fixed radius $\gamma(i) = \epsilon_n$, used as a baseline.
    \item \texttt{Shrinking}: The theoretically motivated schedule \citep{karamanSamplingbasedAlgorithmsOptimal2011}, adapted as a heuristic:
    \[ \gamma(i) = \epsilon_n \left( \frac{\log(i+1)}{i+1} \right)^{1/d} \]
    \item \texttt{Linear Decay}: A linear interpolation from $\gamma$ to a minimum factor $\epsilon_{\min}$ over $N$ total samples:
    \[ \gamma(i) = \epsilon_n \left( 1 - \frac{i}{N}(1 - \epsilon_{\min}) \right) \]
    \item \texttt{Exponential Decay}: An exponential decay from $\gamma$ towards the minimum, controlled by a rate $k$:
    \[ \gamma(i) = \epsilon_n \left( \epsilon_{\min} + (1 - \epsilon_{\min}) e^{-k \cdot i/N} \right) \]
\end{itemize}
As will be shown in the results, the choice of schedule presents a clear trade-off between computation time and solution quality.

In summary, RD-RRT$^*$ trades the formal optimality guarantee of RRT$^*$ for the ability to handle an expressive, non-metric risk cost in a computationally tractable manner. It is therefore presented as a powerful heuristic algorithm designed to efficiently find high-quality, low-risk paths.

\section{Generalizations and Modular Components}
\label{sec:extensions}
Our implementation separates the core planning logic from the specific strategies used for spatial queries and search guidance. This allows for the flexible composition of different components to tailor the algorithm's behavior. The two primary modular components are the Neighbor Finder and the Sampler.

The same separation also supports different stable uncertainty families and workspace dimensions. In addition to the two-dimensional Gaussian closed form used in the core presentation, we derive edge costs for two-dimensional Cauchy uncertainty and for Gaussian and Cauchy models in three dimensions. Cauchy tails decay polynomially and therefore retain a risk signal farther from obstacle modes, while the three-dimensional extension applies the same planner structure with dimension-specific edge costs. The radius scaling of the probability approximation changes with dimension: the 2D approximation is $P^a(C,r)\approx r_d(\gamma_d)r$, whereas the 3D swept cross-section gives $P^a(C,r)\approx (\pi/2)r_d(\gamma_d)r^2$. Thus, under the Risk Density normalization used here, the 3D density edge cost is multiplied by $\pi r^2/2$, rather than by $r$. The main empirical findings are summarized in \cref{sec:res}; the 3D scaling and closed forms are detailed in \cref{app:radius_scaling_3d,app:additional}, with secondary benchmarks in \cref{app:experiments}.

\subsection{Sampling Strategies}
The choice of Sampler provides distinct behaviors, allowing the user to tailor the search heuristic based on prior knowledge of the environment. The optimal strategy is often problem-dependent.
\begin{itemize}[leftmargin=*]
    \item \texttt{UniformSampler}: This is the baseline strategy, which draws samples uniformly from $\mathcal{C}_{free}$. It is unbiased but can be inefficient in environments with narrow passages, where its exploration is not guided toward the goal.

    \item \texttt{DispersedSampler}: An exploration-focused strategy designed to rapidly cover the search space. At each iteration, it generates $k$ uniform candidate samples and returns the one that is farthest (in Euclidean distance) from any existing node in the tree, actively pushing the search into large, unexplored regions.

    \item \texttt{FrontierSampler}: An exploitation-focused strategy that aims to accelerate convergence. It maintains a priority queue of \enquote{frontier} nodes (the leaves of the RRT$^*$ tree), prioritized by a heuristic cost $f(n) = g(n) + \lambda h(n)$, where $g(n)$ is the known cost-to-come and $h(n)$ is a heuristic cost-to-go (Euclidean distance to the goal). It preferentially samples near the most promising frontier node, concentrating the search effort.

    \item \texttt{MahalanobisFrontierSampler}: A risk-aware exploitation strategy that extends the \texttt{FrontierSampler}. Its heuristic incorporates a penalty term inversely proportional to the Mahalanobis distance to the nearest uncertain obstacle. This actively guides the search towards frontiers that are not only close to the goal in terms of path cost but are also situated in demonstrably safer regions, farther from the high-probability areas of uncertain obstacles.
\end{itemize}

\subsection{Neighbor-Finding Strategies}
The efficiency of an RRT$^*$-based algorithm is heavily dependent on the speed of its nearest-neighbor and range queries. We implement two strategies to highlight this:
\begin{itemize}[leftmargin=*]
    \item \texttt{LinearFinder}: A naive baseline implementation that performs a linear scan through all existing nodes in the tree for every query. Its computational complexity is $O(N)$ for a tree with $N$ nodes. 
    \item \texttt{NtreeFinder}: Organizes nodes in an N-tree (a quadtree in 2D) and supports incremental updates. RRT$^*$ uses the incremental N-tree finder, while the batch-based BIT$^*$ and AIT$^*$ implementations use a bucket N-tree variant through the same neighbor-query interface.
\end{itemize}
The Finder.\texttt{Update} call in \cref{alg:RDRRT*} signifies the addition of a leaf to the finder tree.

\section{Numerical Validation}
\label{sec:setup}
\subsection{Experimental Setup}
\label{subsec:experimental_setup}
The experiments are organized into a sequence of benchmarks designed to systematically evaluate the algorithm. Unless otherwise specified, all experiments use the following default parameters: $\epsilon_n = 0.3$, $\epsilon_{steer} = 0.3$, $b=0.3$, $N=5000$, and the \texttt{UniformSampler}. The rewiring schedule default for $\epsilon_n$ is the \texttt{Constant} one.

\subsubsection{Benchmark 1: Foundational Performance and Accuracy.}
This benchmark establishes the core performance of RD-RRT$^*$ and the fidelity of its underlying approximation in the \enquote{Default Scenario}. We compare the primary cost variants (Prob, Risk, and Hybrid) against RRT$^*$ with obstacle radii enlarged by $\chi=1.05$, the best value found for inflated RRT$^*$ in terms of Monte Carlo path probability in \cref{tab:inflated-rrt}, and Chance-Constrained RRT$^*$ (CC-RRT$^*$) \citep{ludersRobustSamplingbasedMotion2013} with a per-step threshold $\delta=0.2$. We additionally compare implementations of BIT$^*$ \citep{gammellBatchInformedTrees2015a} and AIT$^*$ \citep{strubAdaptivelyInformedTrees2020a}. RRT$^*$, BIT$^*$, and AIT$^*$ optimize Euclidean path length in the inflated-obstacle problem. RD-RRT$^*$ optimizes the pure RD cost, while the configurations denoted BIT$^*$-RD and AIT$^*$-RD use the hybrid RD cost with $\lambda=10^{-8}$; see \cref{subsec:regularization} for the rationale.

The information available to BIT$^*$ and AIT$^*$ depends on the objective. For the Euclidean inflated-obstacle problem, both planners use exact Euclidean edge and goal lower bounds; after finding a solution, they sample from the corresponding prolate hyperspheroid. AIT$^*$ additionally updates an adaptive cost-to-go estimate through reverse graph search. For the RD term, all admissible lower-bound heuristics are set to the uninformative null bound. The minimally regularized RD variants can use the exact hybrid cost when an edge is processed and the weak bounds $\lambda\|x-y\|$ for cost-to-come and cost-to-go, but no direct informed sampler is available for the combined objective. Their batches therefore remain uniform, with one explicit goal sample per batch. In particular, AIT$^*$'s reverse search receives no informative lower-bound signal from the RD component.

All configurations are run 100 times, use 5000 planning samples, and validate the final path with 10,000 Monte Carlo samples to compute the reference value of $P(C)$. They share the same straight-line steering and deterministic collision checker, use a search-radius factor of $0.5$, and reject evaluated edges above the cost-steering threshold $0.2$. RRT$^*$ uses the incremental N-tree finder and a goal bias of $0.01$. BIT$^*$ and AIT$^*$ use bucket N-tree finders, one explicit goal sample per batch, and batch sizes of 1000 and 500, respectively. A maximum planning time of 300 seconds is imposed, and every configuration achieved 100 successful runs.

\subsubsection{Benchmark 2: Robustness in Challenging Environments.}
This benchmark evaluates the planner's ability to navigate scenarios where geometric simplicity and safety are in direct conflict. We use two specialized environments:
\begin{itemize}[leftmargin=*]
    \item The \textbf{Local Minimum Scenario} is designed to probe the consequences of the heuristic-metric mismatch detailed in \cref{subsec:theory_and_heuristics}. Its purpose is to induce and analyze the "rewiring catastrophe" behavior and demonstrate the effectiveness of the Hybrid variant's regularization term.
    \item The \textbf{Risk Trap Scenario} tests the planner's ability to escape deceptive local minima in the risk landscape, comparing RD-RRT$^*$'s pathwise accumulation of RD cost against the more myopic or locally constrained behaviors of the baseline planners.
\end{itemize}

\subsubsection{Benchmark 3: Scalability and Search Guidance.}
This benchmark assesses the algorithm's efficiency. First, we evaluate its computational \textit{scalability} by measuring the increase in planning time as the number of uncertain obstacles grows within \enquote{Complex Random Scenarios}. Second, we analyze the impact of \textit{search guidance} by comparing the anytime performance and convergence rates of different sampling strategies (Uniform, Dispersed, Frontier-based) across multiple test scenarios.

\subsubsection{Benchmark 4: Parameter Sensitivity Analysis.}
The final experiment characterizes the algorithm's sensitivity to its key internal parameters. We analyze how solution quality and computation time are affected by variations in the rewiring search radius $\epsilon_n$ (including different decay schedules as described in \cref{subsec:theory_and_heuristics}) and the maximum steering cost $\epsilon_{steer}$.

Additional benchmarks are provided in \cref{app:experiments}.

\subsubsection{Methodological Note.}
Based on the findings from Benchmark 2 regarding robustness, the \textbf{Hybrid RD-RRT$^*$ variant (with $\lambda=10^{-8}$) is selected as the standard planner} for the subsequent analyses of sampling strategies, scalability, and parameter sensitivity to ensure consistent and robust performance.

\subsection{Test Scenarios}
\label{subsec:scenarios}
\begin{figure}[t]
    \centering
    \newcommand{\scenimgH}{0.26\textheight}

    \begin{subfigure}[t]{0.48\linewidth}
        \centering
        \includegraphics[width=\linewidth,height=\scenimgH,keepaspectratio]{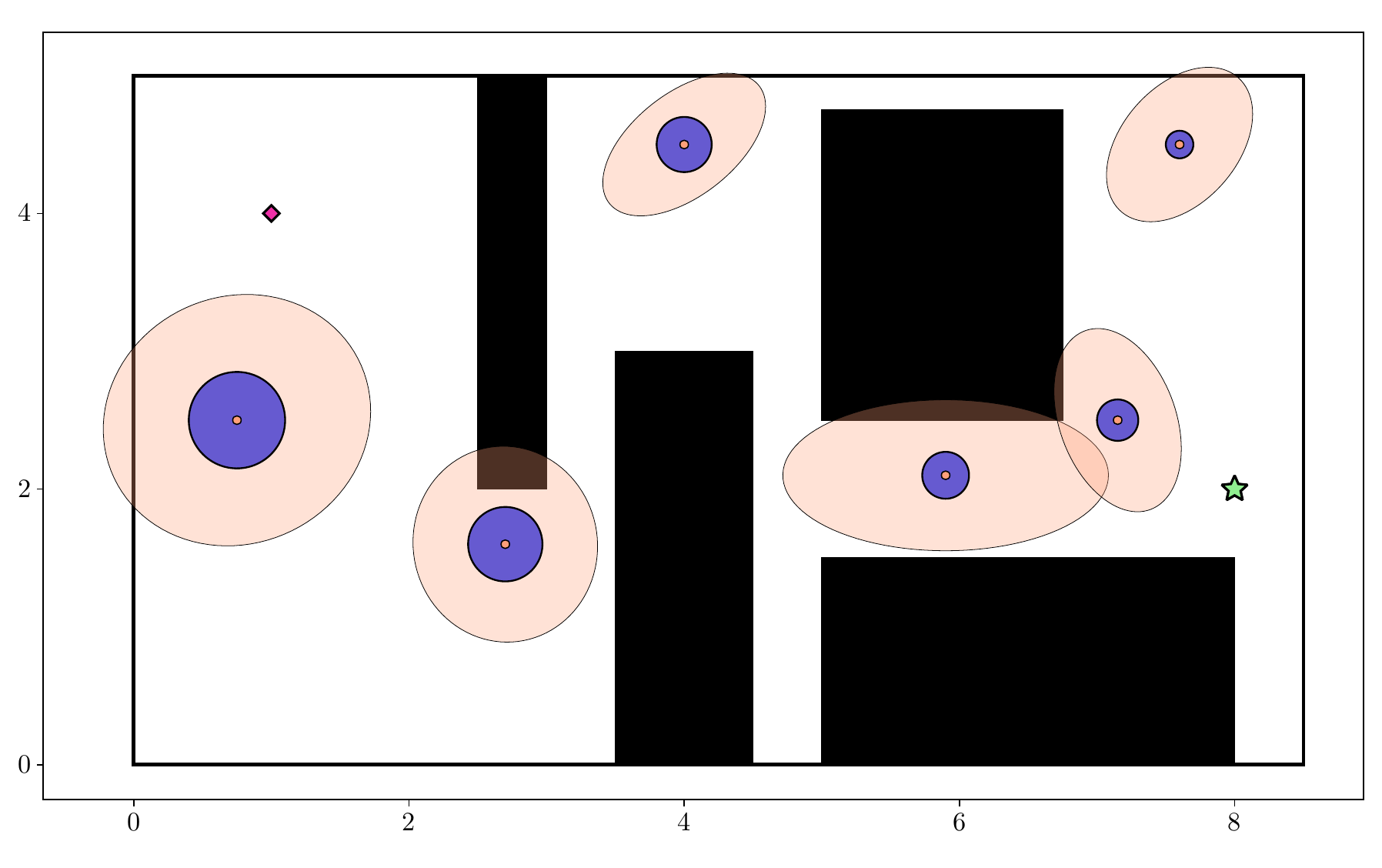}
        \caption{Default Scenario.}
        \label{subfig:defaultscenario}
    \end{subfigure}\hfill
    \begin{subfigure}[t]{0.48\linewidth}
        \centering
        \includegraphics[width=\linewidth,height=\scenimgH,keepaspectratio]{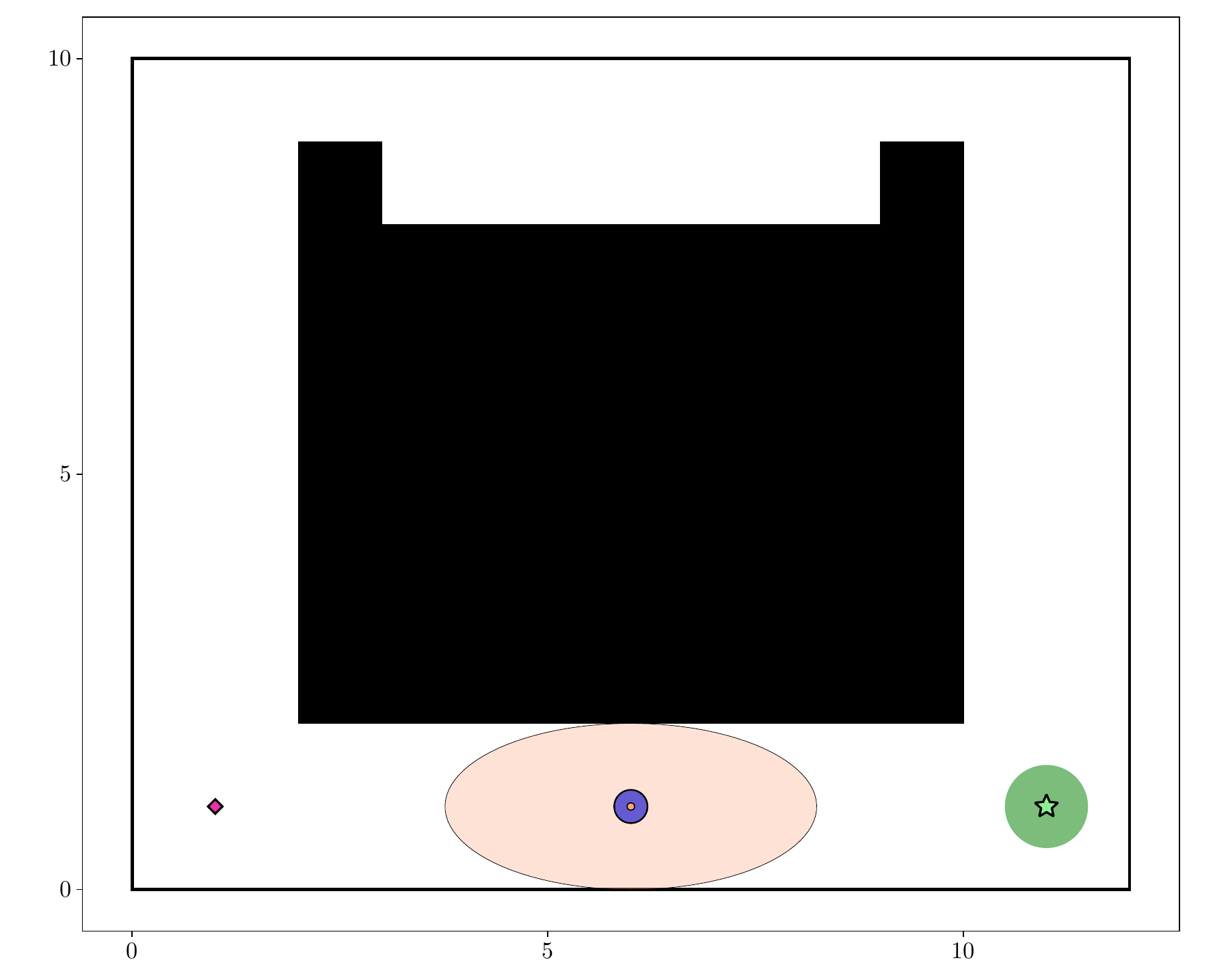}
        \caption{Local Minimum Scenario.}
        \label{subfig:localminimum}
    \end{subfigure}

    \vspace{0.6em}

    \begin{subfigure}[t]{0.48\linewidth}
        \centering
        \includegraphics[width=\linewidth,height=\scenimgH,keepaspectratio,angle=-90,origin=c]{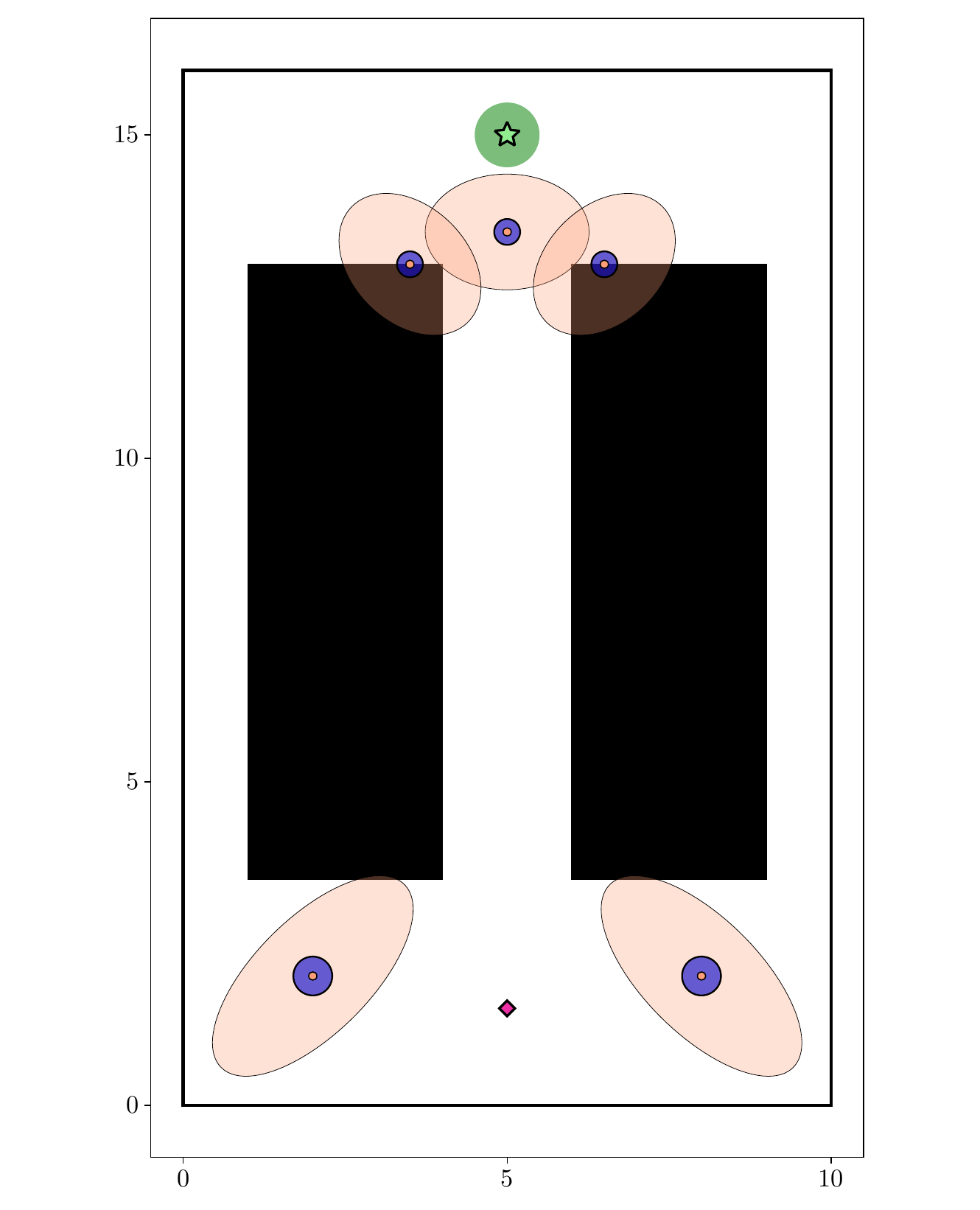}
        \caption{Risk Trap Scenario.}
        \label{subfig:risktrap}
    \end{subfigure}\hfill
    \begin{subfigure}[t]{0.48\linewidth}
        \centering
        \includegraphics[width=\linewidth,height=\scenimgH,keepaspectratio]{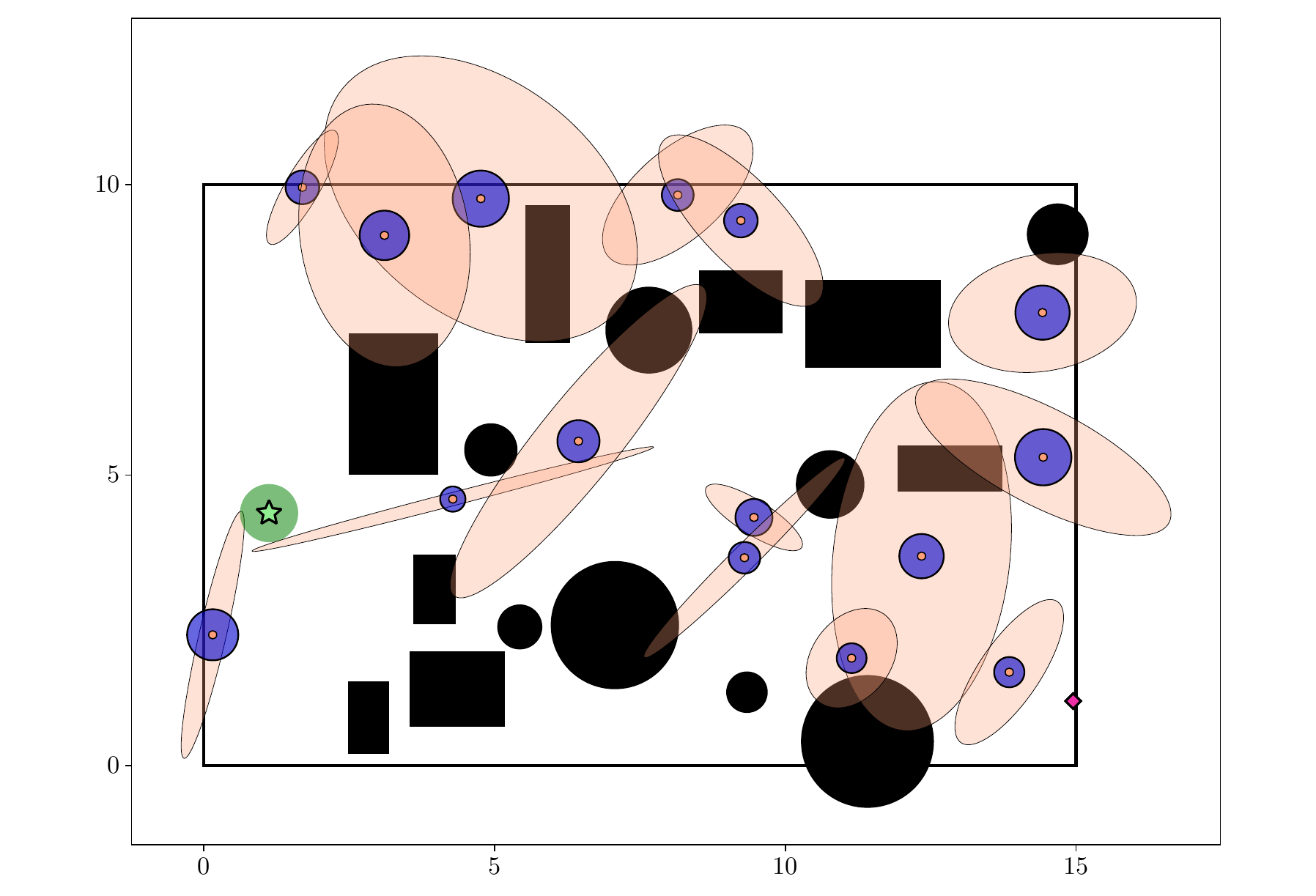}
        \caption{Example of Complex Scenario.}
        \label{subfig:complex}
    \end{subfigure}
    \caption{In all scenarios, deterministic obstacles are black, whereas uncertain obstacles are blue disks with light-orange ellipses showing two standard deviations about their means. The start and goal are, respectively, a pink diamond and a green star.}
    \label{fig:scenarios}
\end{figure}

\subsubsection{Default Scenario}
This environment, in \cref{subfig:defaultscenario}, serves as a consistent baseline for validating core performance claims and comparing results against previous findings.
\subsubsection{Local Minimum Scenario}
This scenario is specifically designed to challenge the planner by creating a conflict between geometric distance and probabilistic cost. It features a short, direct path to the goal that is obstructed by a high-risk uncertain obstacle, and a significantly longer, but safer, alternative route, as shown in \cref{subfig:localminimum}.
\subsubsection{Risk Trap Scenario}
This environment tests the planner's ability to escape deceptive local cost minima. It features a \enquote{funnel} of deterministic obstacles that guides the search towards a narrow passage blocked by a cluster of high-risk uncertain obstacles, see \cref{subfig:risktrap}. The optimal path requires avoiding the funnel entirely, demanding a more global exploration of the search space.
\subsubsection{Complex Random Scenario}
Used for general robustness and scalability testing, these scenarios are generated procedurally with a high density of both deterministic obstacles (of mixed shapes) and uncertain obstacles. The start and goal locations are chosen to ensure a non-trivial planning problem. One of these randomly generated environments is shown in \cref{subfig:complex}.

\subsection{Ground Truth via Monte Carlo Simulation}
\label{subsec:mc_methodology}
To serve as ground truth for collision probability, we employ multithreaded Monte Carlo (MC) simulations. This allows for an unbiased estimate of the true collision probability for a given path under the two primary hypotheses of event interdependence, H2 and H3, as discussed in \cref{sec:prelim}. The simulations use all available cores, with 10,000 trials per simulation.

\subsubsection{H3 (Stopped Process)}
This hypothesis models a static but unknown environment. For each MC trial, the positions of all uncertain obstacles are sampled once from their respective Gaussian distributions at the start of the trial. A collision is then recorded if any segment of the robot's path intersects with any of the obstacles in this fixed configuration. The final probability is the fraction of trials in which a collision occurred.

\subsubsection{H2 (Markov Property on Newly Swept Area)}
This hypothesis models a sequential process where uncertainty is resolved as the robot moves. The simulation proceeds segment-by-segment along the path. For each MC trial, an initial world state is sampled for the first path segment. For each subsequent segment, the position of each obstacle is conditionally re-sampled if its previous position would have caused a collision with the segment just traversed. This is achieved via rejection sampling on an obstacle-by-obstacle basis, leveraging their statistical independence. A collision check is then performed on the current segment using the newly-conditioned world state. A trial is marked as a failure if a collision occurs on any segment.

\subsection{Physics-Based Execution Validation}
\label{subsec:mujoco_methodology}
To evaluate the planned paths beyond the point-robot geometric model, we execute them in MuJoCo \citep{todorovMuJoCoPhysicsEngine2012} using the Stanford TidyBot holonomic mobile-base model \citep{wuTidyBotOpenSourceHolonomic}. Planning is performed against conservative disk and box abstractions, whereas validation uses the finite-footprint robot and native three-dimensional collision meshes of the MuJoCo model and obstacle instances.

The base is operated in holonomic mode. A PD controller sends planar position and heading references to the model's native position actuators. The heading reference follows the path tangent. To avoid reacting only to individual corners of the piecewise-linear path, the controller samples ten look-ahead points and orients the robot according to a front-loaded weighted mean of their tangents. MuJoCo then integrates the controlled motion and resolves contacts against the native collision geometries.

We consider procedurally generated complex scenarios containing 15 deterministic and eight uncertain obstacles. The planner seed is fixed, while each scenario seed defines one environment and one point in the approximation-quality analysis. For each scenario, 1000 obstacle-pose samples are generated (in accordance with \textbf{H3}). Every sampled world is checked both with the geometric collision model against the cylinders bounding the sampled uncertain obstacles, and through a complete MuJoCo rollout of the same planned path. The two empirical collision frequencies are then compared with the RD approximation.

\subsection{Implementation Details}
All experiments were conducted on a machine with an Intel i7-12700H CPU and 32 GB of RAM. The planner and benchmark suite are implemented in the Julia programming language. To ensure accurate timings, individual planning runs are executed serially on a single core. Each data point was evaluated 50--100 times and averaged to reduce variability in the results.

\section{Results}
\label{sec:res}
\subsection{Performance Against Baselines}
\begin{figure}[t]
    \centering
    \includegraphics[width=\linewidth, clip, trim = 0 4.5cm 0 0]{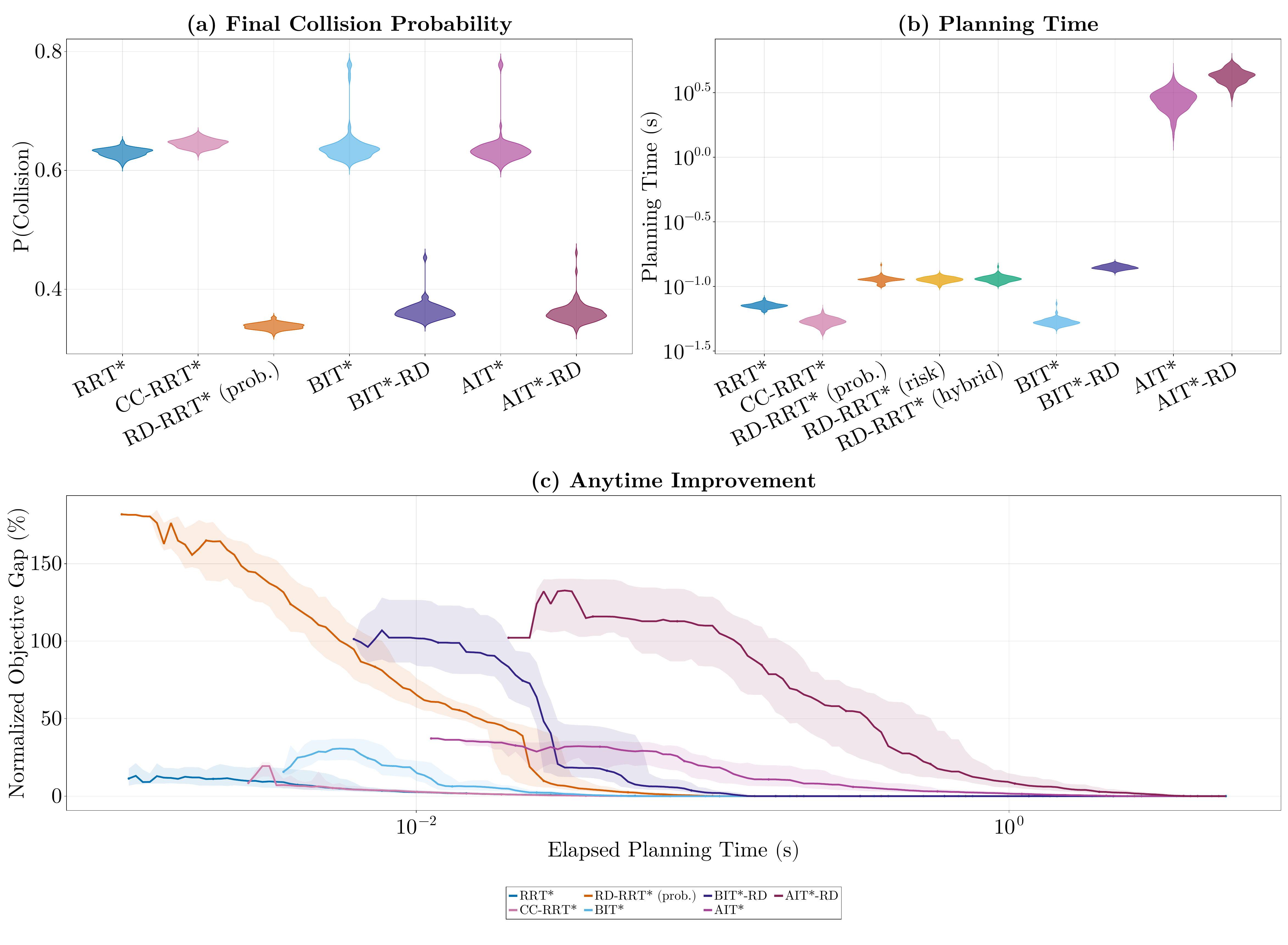}
    \caption{Benchmark 1 over 100 runs. \textbf{(a)} Final H3 Monte Carlo collision probability. \textbf{(b)} Planning-time distribution. \textbf{(c)} Anytime convergence of each planner's objective cost; lines show the median and shaded bands span the interquartile range across successful runs.}
    \label{fig:baseline_comparison}
\end{figure}

\Cref{fig:baseline_comparison} shows a clear separation in path safety. The final H3 collision probability $P(C)$ is estimated independently for every returned path using 10,000 Monte Carlo samples. RD-RRT$^*$ produces a distribution centered near $0.34$, whereas the geometric and chance-constrained configurations remain near $0.63$--$0.65$. The BIT$^*$-RD and AIT$^*$-RD variants also produce substantially lower Monte Carlo collision probabilities than their Euclidean counterparts, indicating that the principal safety improvement follows from optimizing the RD objective rather than from changing the search strategy. Among the tested RD configurations, the simpler RD-RRT$^*$ returns the lowest final Monte Carlo estimates.

\begin{figure}[t]
    \centering
    \includegraphics[width=\linewidth]{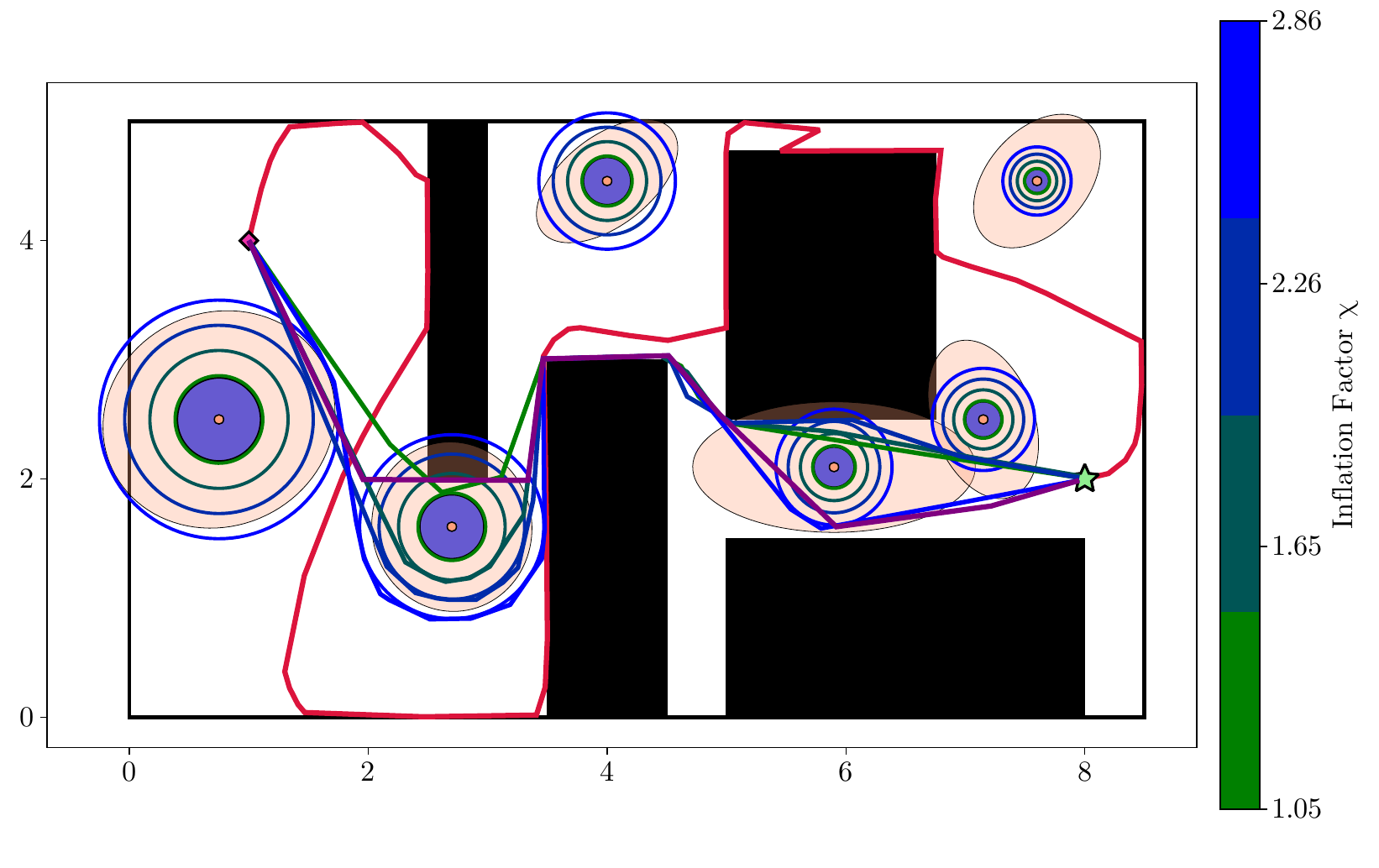}
    \caption{Representative final paths in the Default Scenario. The RD-RRT$^*$ path is red and the CC-RRT$^*$ path is purple. Inflated RRT$^*$ paths and obstacle contours are colored by their inflation coefficient $\chi$.}
    \label{fig:baseline_paths}
\end{figure}

The representative solutions in \cref{fig:baseline_paths} illustrate this difference. Geometric and locally constrained planning favor the shorter passage through the uncertain-obstacle field, while RD-RRT$^*$ accepts a longer detour to reduce cumulative path risk. The numerical data in \cref{tab:inflated-rrt} further show that increasing $\chi$ produces longer paths but does not monotonically decrease collision risk. Likewise, \cref{tab:cc-rrt} shows that enforcing a local per-step threshold does not directly control cumulative path risk. In contrast, \cref{fig:hybrid-rrt} shows that Hybrid RD-RRT$^*$ can vary the trade-off between path length and safety through $\lambda$.

The lower Monte Carlo collision probability is obtained with a moderate increase in planning time: RD-RRT$^*$ has a median planning time of $0.113$ s, compared with $0.071$ s for RRT$^*$ and $0.053$ s for BIT$^*$. BIT$^*$-RD has a similar median time of $0.140$ s, whereas both AIT$^*$ configurations are substantially slower in this inexpensive 2D problem. All configurations succeed in every run. The anytime panel tracks each algorithm's own objective cost rather than Monte Carlo $P(C)$. At each elapsed-time grid point, its normalized objective gap is $100(C_t/C_{\mathrm{final}}-1)\%$, where $C_t$ is the incumbent cost and $C_{\mathrm{final}}$ is the final cost of the same run; negative gaps are clipped to zero. The shaded regions span the interquartile range across successful runs. Because the objectives differ, these curves measure within-run convergence rather than safety on a common scale.

\begin{table}[!t]
    \centering
    \caption{Performance of CC-RRT$^*$ vs. the per-step probability threshold $\delta$. The planner failed to find solutions for $\delta < 0.2$.}
    \label{tab:cc-rrt}
    \footnotesize
    \setlength{\tabcolsep}{5pt}
    \begin{tabular}{l|rrrr}
      \toprule
      $\delta$ & 0.2 & 0.25 & 0.3 & 0.35 \\ \midrule
      \cref{eq:probabilityconcatenationpaths} & 0.853 & 0.971 & 1.045 & 1.104 \\
      (H2) [\%] & 63.80 & 68.99 & 71.96 & 74.27 \\
      (H3) [\%] & 63.81 & 69.04 & 71.99 & 74.28 \\
      Length & 12.23 & 11.37 & 11.02 & 10.75 \\
      \bottomrule
    \end{tabular}
\end{table}

\begin{table}[!t]
    \centering
    \caption{Performance of RRT$^*$ with inflated obstacles vs. safety factor $\chi$. The lowest Monte Carlo collision frequencies are shown in bold.}
    \label{tab:inflated-rrt}
    \setlength{\tabcolsep}{3pt}
    \footnotesize
    \begin{adjustbox}{max width=\linewidth}
    \begin{tabular}{l|rrrrrrrrrrrrrr}
      \toprule
      $\chi$ & 1.05 & 1.19 & 1.33 & 1.47 & 1.61 & 1.75 & 1.89 & 2.03 & 2.16 & 2.30 & 2.44 & 2.58 & 2.72 & 2.86 \\\midrule
      \cref{eq:probabilityconcatenationpaths} & 0.868 & 0.872 & 0.946 & 1.176 & 1.172 & 1.157 & 1.140 & 1.120 & 1.088 & 1.020 & 0.973 & 0.934 & 0.895 & 0.852 \\
      (H2) [\%] & \textbf{63.12} & 63.20 & 66.40 & 76.35 & 76.17 & 75.66 & 74.94 & 74.14 & 72.83 & 70.56 & 68.68 & 67.11 & 65.52 & 63.53 \\
      (H3) [\%] & \textbf{63.01} & 63.15 & 66.31 & 76.46 & 76.20 & 75.57 & 74.97 & 74.09 & 72.91 & 70.55 & 68.72 & 67.07 & 65.43 & 63.53 \\
      Length & 9.01 & 9.02 & 9.33 & 10.31 & 10.42 & 10.52 & 10.61 & 10.72 & 10.86 & 11.15 & 11.33 & 11.47 & 11.63 & 11.94 \\
      \bottomrule
    \end{tabular}
\end{adjustbox}
\end{table}

\begin{figure}[!t]
    \centering
    \includegraphics[width=0.8\linewidth]{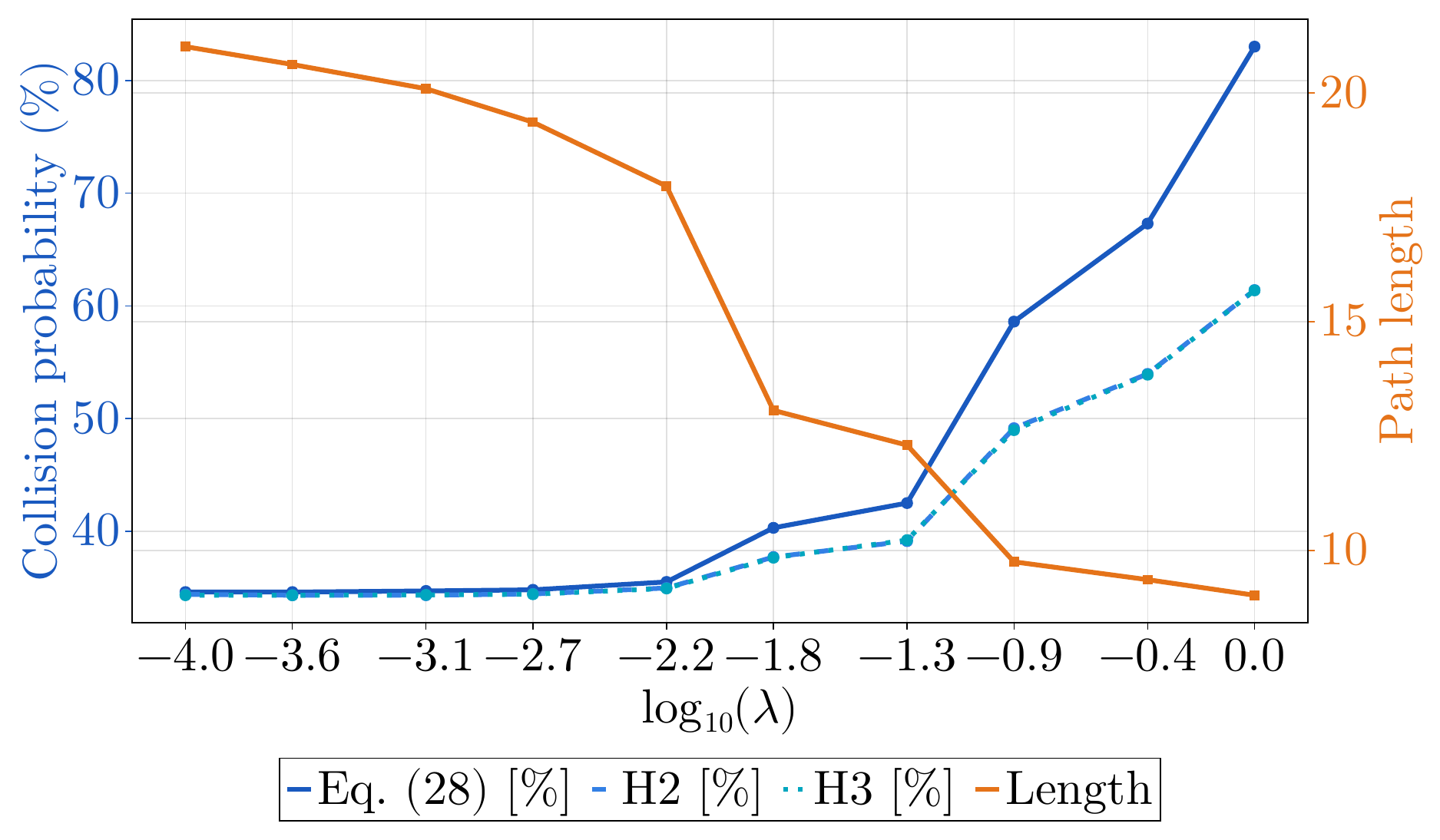}
    
    \vspace{1em} %

    \footnotesize
    \setlength{\tabcolsep}{4pt}
    \begin{adjustbox}{max width=\linewidth}
    \begin{tabular}{l|rrrrrrrrrr}
      \toprule
      $\lambda$ & $10^{-4}$ & $10^{-3.6}$ & $10^{-3.1}$ & $10^{-2.7}$ & $10^{-2.2}$ & $10^{-1.8}$ & $10^{-1.3}$ & $10^{-0.9}$ & $10^{-0.4}$ & $1$ \\\midrule
      \cref{eq:probabilityconcatenationpaths} & 0.346 & 0.346 & 0.347 & 0.348 & 0.355 & 0.403 & 0.425 & 0.586 & 0.673 & 0.830 \\
      (H2) [\%] & 34.39 & 34.32 & 34.32 & 34.43 & 34.96 & 37.71 & 39.12 & 49.14 & 53.96 & 61.40 \\
      (H3) [\%] & 34.32 & 34.32 & 34.35 & 34.41 & 34.92 & 37.65 & 39.21 & 48.97 & 53.89 & 61.37 \\
      Length & 21.01 & 20.62 & 20.09 & 19.36 & 17.96 & 13.06 & 12.30 & 9.75 & 9.36 & 9.02 \\
      \bottomrule
    \end{tabular}
    \end{adjustbox}

    \caption{Performance of the Hybrid RD-RRT$^*$ variant vs. length weight $\lambda$.}
    \label{fig:hybrid-rrt}
\end{figure}

\subsection{Accuracy of the Risk-Density Approximation}
\begin{SCfigure}
    \centering
    \includegraphics[clip,trim={0 0cm 0 0cm},width=0.6\linewidth]{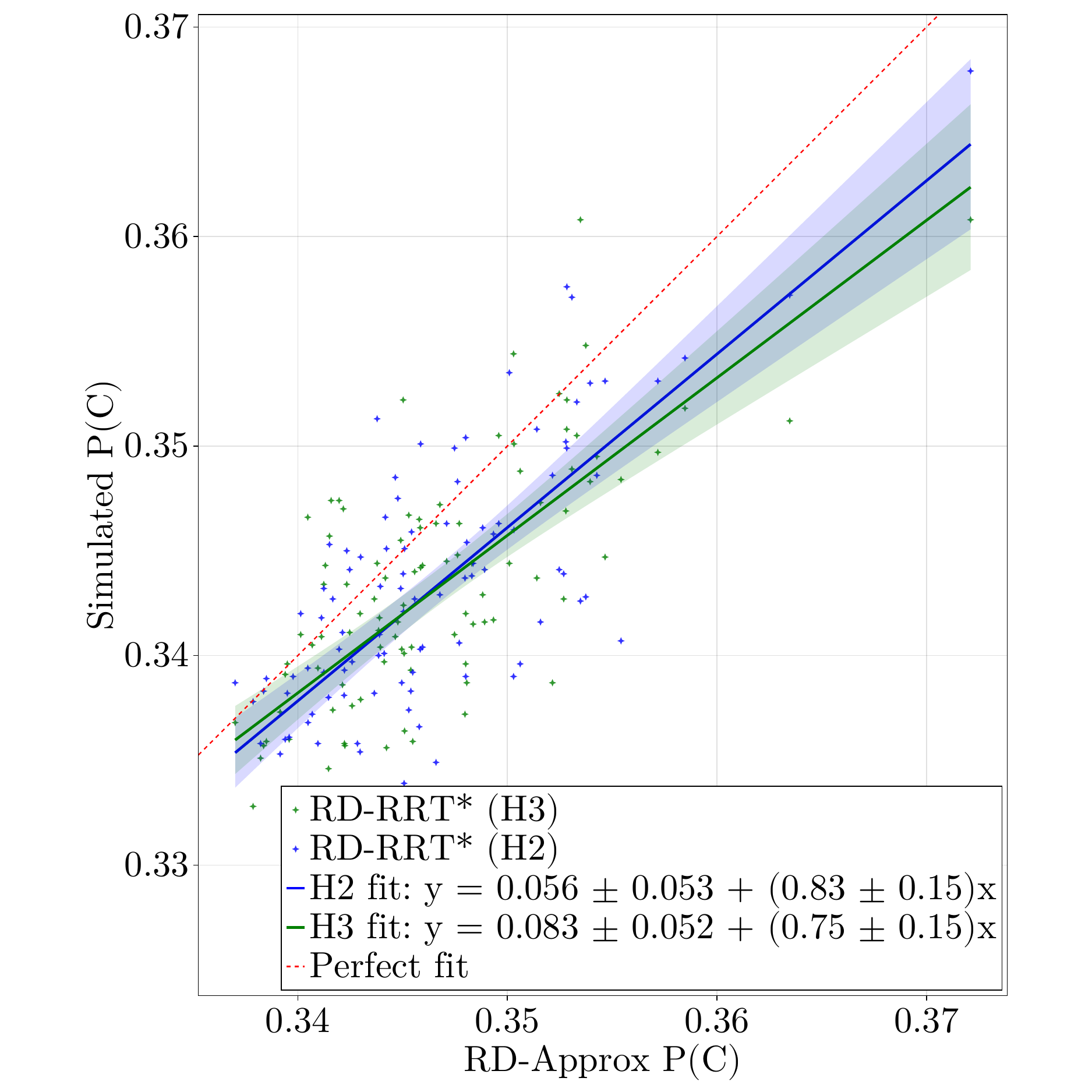}
    \caption{Risk-density approximation against Monte Carlo collision probability for successful paths in the Default Scenario. Shaded regions are 95\% confidence bands for the mean fitted regression relationships.}
    \label{fig:approximation_quality}
\end{SCfigure}

A crucial element of RD-RRT$^*$ is the fidelity of the cost function \eqref{eq:rdformultiple} as a proxy for the true collision probability. \Cref{fig:approximation_quality} plots the approximated probability from \cref{eq:probabilityconcatenationpaths} against the ground-truth probability obtained from MC simulations for all successful paths generated by \cref{alg:RDRRT*}. The results show a strong linear correlation between the approximation and the simulation for both the H2 and H3 interdependence hypotheses, along with a very tight clustering for the two hypotheses.

\subsection{Physics-Based MuJoCo Validation}
\begin{figure}[t]
    \centering
    \begin{minipage}[c]{0.48\linewidth}
        \centering
        \includegraphics[width=\linewidth, clip]{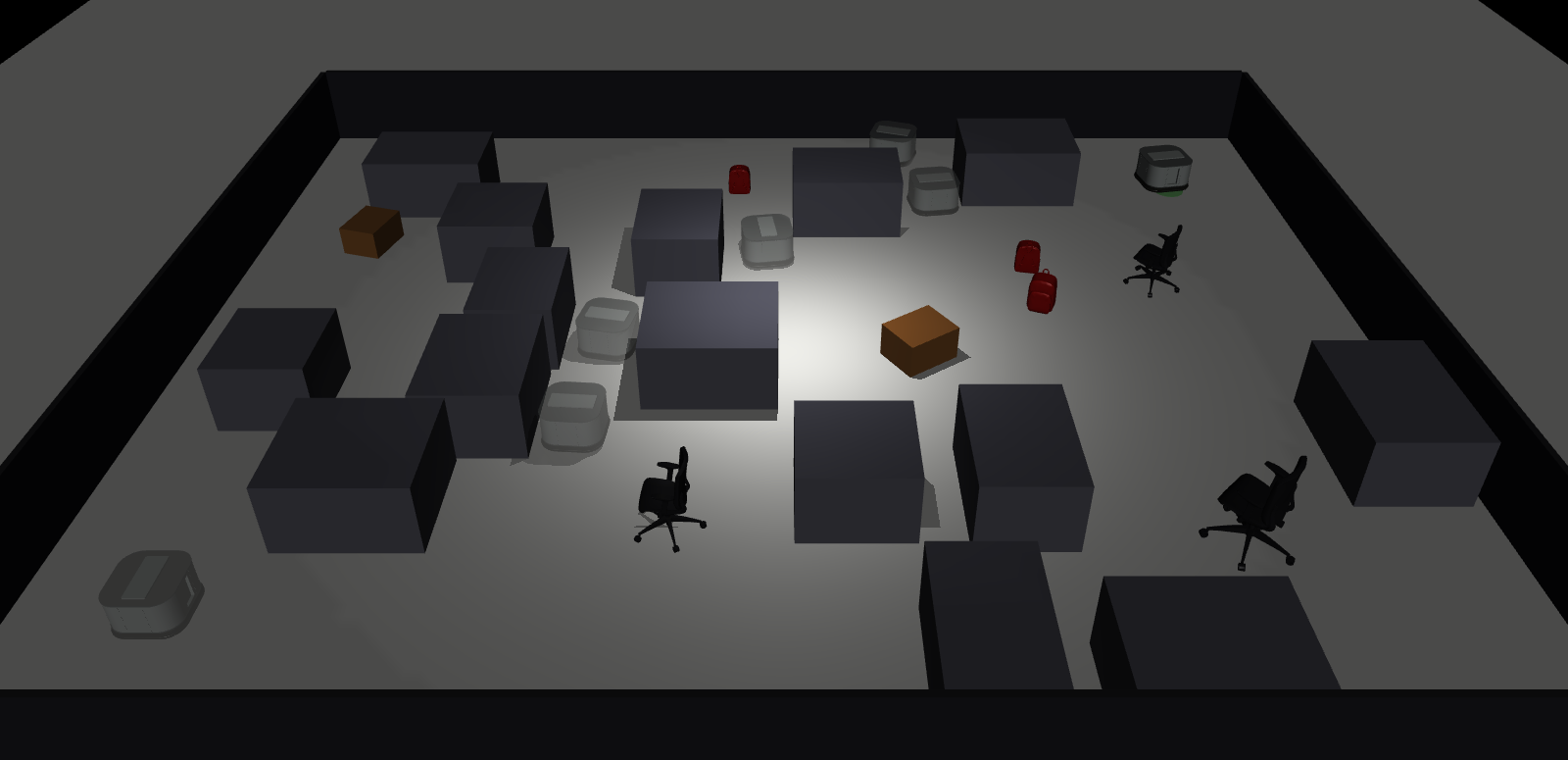}

        \vspace{0.1em}

        \includegraphics[width=\linewidth, clip, trim = 0 0 4cm 1cm]{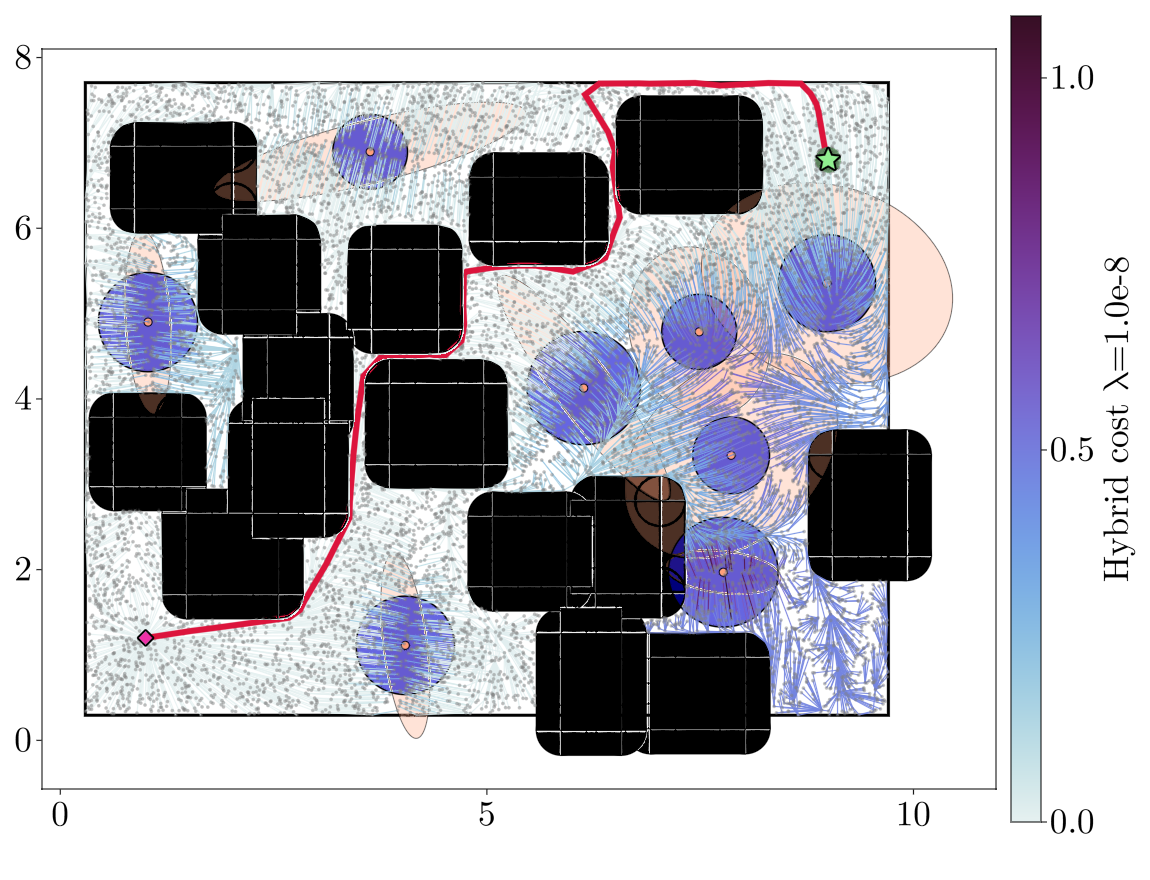}
        
    \end{minipage}\hfill
    \begin{minipage}[c]{0.48\linewidth}
        \centering
        \includegraphics[width=\linewidth]{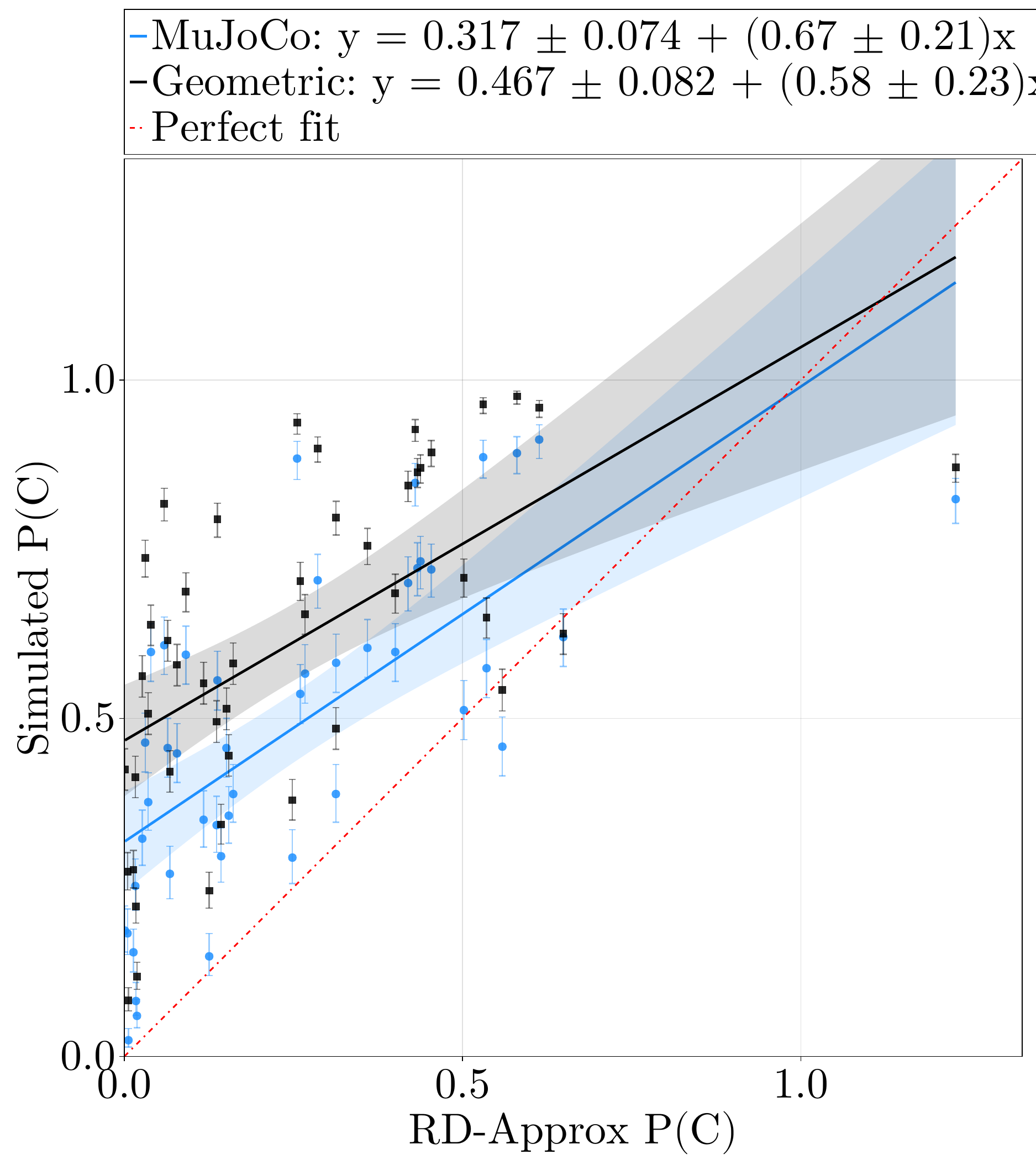}
    \end{minipage}
    \caption{Physics-based validation. \textbf{(Bottom left)} RD-RRT$^*$ plan with inflated obstacles in a sampled complex scenario. \textbf{(Top left)} Overlaid TidyBot configurations at successive times during the corresponding closed-loop MuJoCo rollout. The obstacle poses are sampled from the template shown at the bottom. \textbf{(Right)} RD approximation against geometric and MuJoCo collision frequencies. Shaded regions are 95\% confidence bands for the mean fitted regressions; vertical error bars are 95\% Wilson binomial intervals for the collision-rate estimates.}
    \label{fig:mujoco_validation}
\end{figure}

\Cref{fig:mujoco_validation} compares the RD approximation with collision frequencies obtained from the geometric checker and from complete MuJoCo rollouts. As described in \cref{subsec:mujoco_methodology}, each point represents one scenario seed, and the two empirical frequencies are computed from the same 1000 sampled obstacle configurations. These scenarios are substantially more complex than the Default Scenario, containing 15 deterministic and eight uncertain obstacles. The approximation is consequently less tightly calibrated than in \cref{fig:approximation_quality}, but it remains in the same broad probability range and retains a positive relationship with both empirical estimates. 

The MuJoCo estimates differ from the purely geometric checks because execution includes the finite TidyBot footprint, native three-dimensional collision meshes, contact resolution, and closed-loop tracking of a piecewise-linear reference. The agreement is therefore not expected to be exact. The results nevertheless place the simulated execution risk in a comparable range to both the geometric estimate and the RD approximation. The left panels of \cref{fig:mujoco_validation} show how the planned path is realized by the holonomic TidyBot in one representative sampled scenario.

\subsection{Sensitivity Analysis} 
\begin{figure}[!ht]
    \centering
    \includegraphics[width=\linewidth]{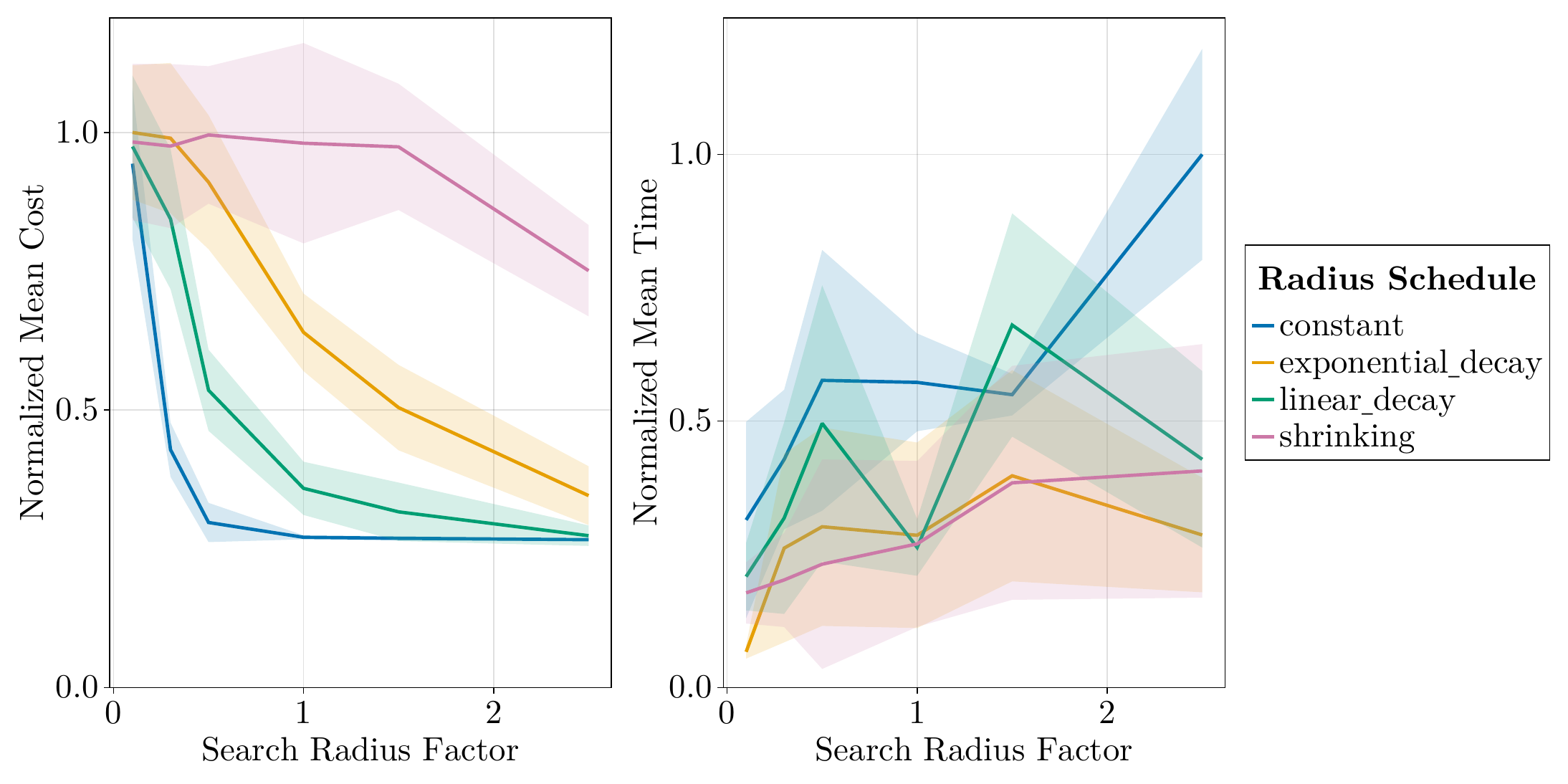}
    \caption{Sensitivity of the Hybrid RD-RRT$^*$ to the search radius factor and decay schedule in the \enquote{Paper Default} scenario. \textbf{(Left)} Larger radii generally lead to lower-cost solutions. \textbf{(Right)} This improvement in solution quality comes with an increase in planning time. Shaded bands show the mean plus or minus one standard deviation across runs.}
    \label{fig:sensitivity_radius}
\end{figure}

The planner's sensitivity to the search radius factor and radius schedule was analyzed.
As shown in \cref{fig:sensitivity_radius}, there is a clear trade-off: larger search radii $\epsilon_n$, particularly with constant or linear decay schedules, lead to better (lower-cost) solutions at the expense of increased computation time. 
The standard RRT$^*$ shrinking schedule is fastest but consistently yields the highest-cost paths. The algorithm showed little to no sensitivity to $\epsilon_{steer}$, so no plot is included.

\subsection{Regularization for Robustness}
\label{subsec:regularization}
\begin{figure}[!ht]
    \centering
    \begin{subfigure}[t]{0.48\textwidth}
        \centering
        \includegraphics[width=\linewidth]{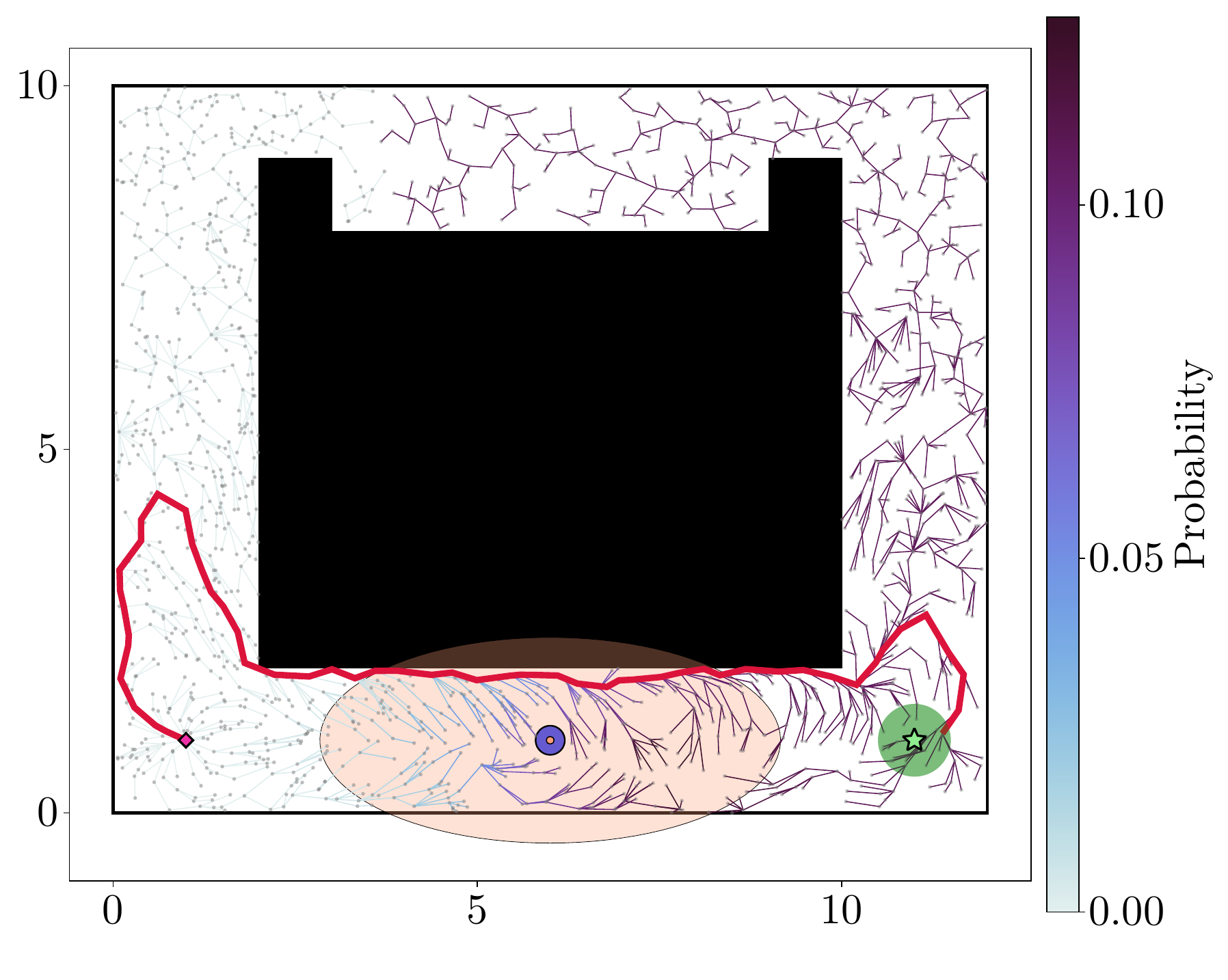} 
        \caption{Failed Planning.} \label{sfig:heuristicfail}
    \end{subfigure}
    \hfill
    \begin{subfigure}[t]{0.48\textwidth}
        \centering
        \includegraphics[width=\linewidth]{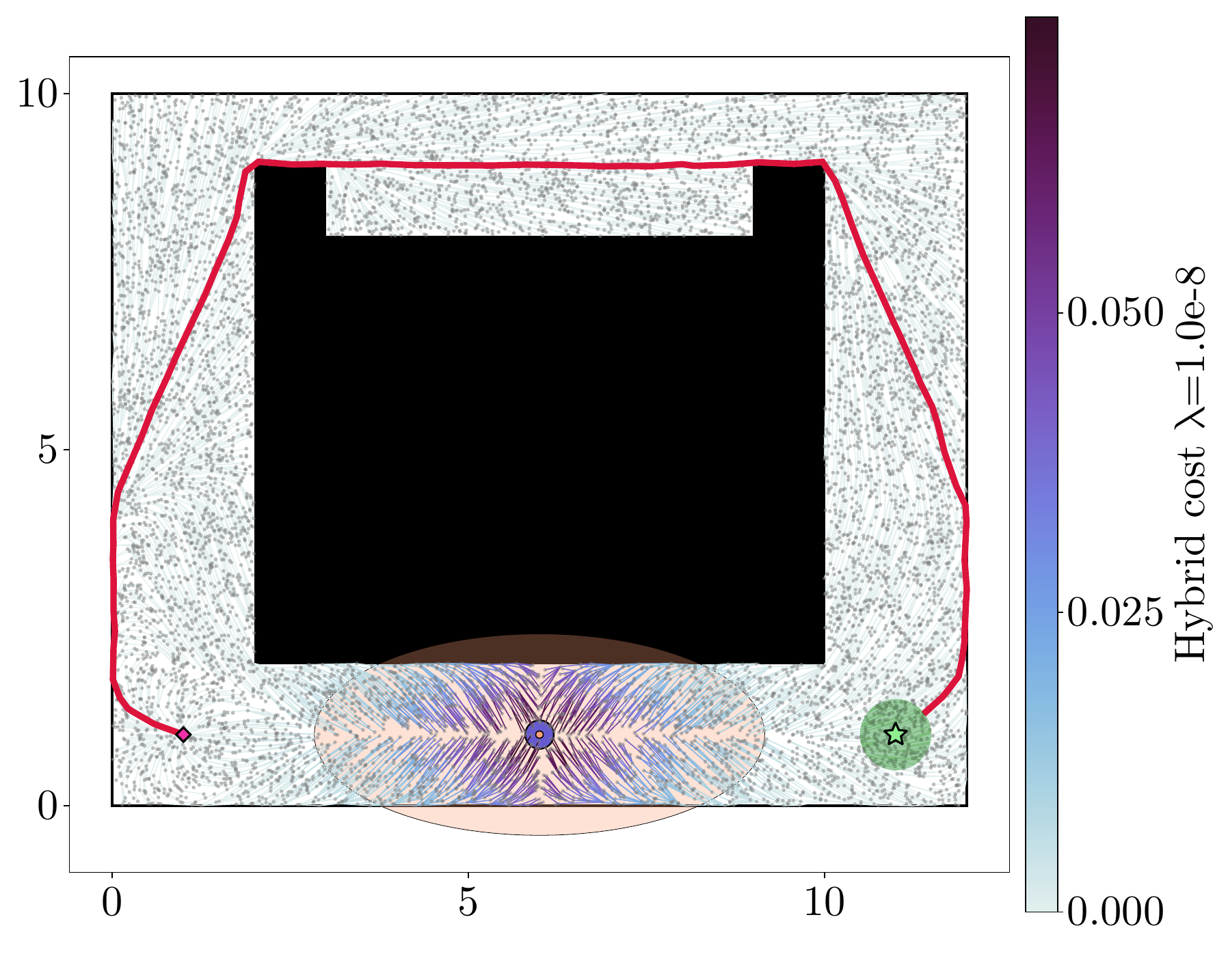}
        \caption{Regularized problem.} \label{sfig:regularized}
    \end{subfigure}
    \caption{\textbf{(Left)} Using the pure probability cost function can stall exploration in low-risk regions due to a heuristic-metric mismatch, resulting in two separate \enquote{branches} close in Euclidean space but with dramatically different costs. \textbf{(Right)} Adding a small length regularization term ($\lambda=10^{-8}$) restores robust exploration, allowing the planner to find a lower-probability path.}
    \label{fig:rewiring_stall}
\end{figure}

A key practical finding relates to the semimetric property of the pure probability cost. In regions where the risk density is very low and flat, the mismatch between the Euclidean neighbor search and the probabilistic cost can cause \enquote{rewiring catastrophes}. There the planner effectively stops in its tracks due to the very large number of rewiring operations to be computed when the two \enquote{branches} of the global tree find one another, as shown in \cref{fig:rewiring_stall}. The introduction of a small Euclidean length penalty in the Hybrid RD-RRT$^*$ variant acts as a crucial regularizer. It ensures the cost function does not encounter regions in which it has no information, restoring robust exploration and rewiring: indeed, at the top of the deterministic obstacle in \cref{fig:rewiring_stall} (left), the exponentially decaying tail of $\mathcal{N}$ makes $\mathit{c}\approx 0$, making the algorithm erratic. This justifies the use of a small regularization parameter $\lambda$ and the choice of the Hybrid as the default variant for robustness testing.

\subsection{Robustness in Deceptive Environments}
\begin{figure}[!ht]
    \centering

    \begin{subfigure}[t]{0.3\textwidth}
        \centering
        \includegraphics[width=\linewidth]{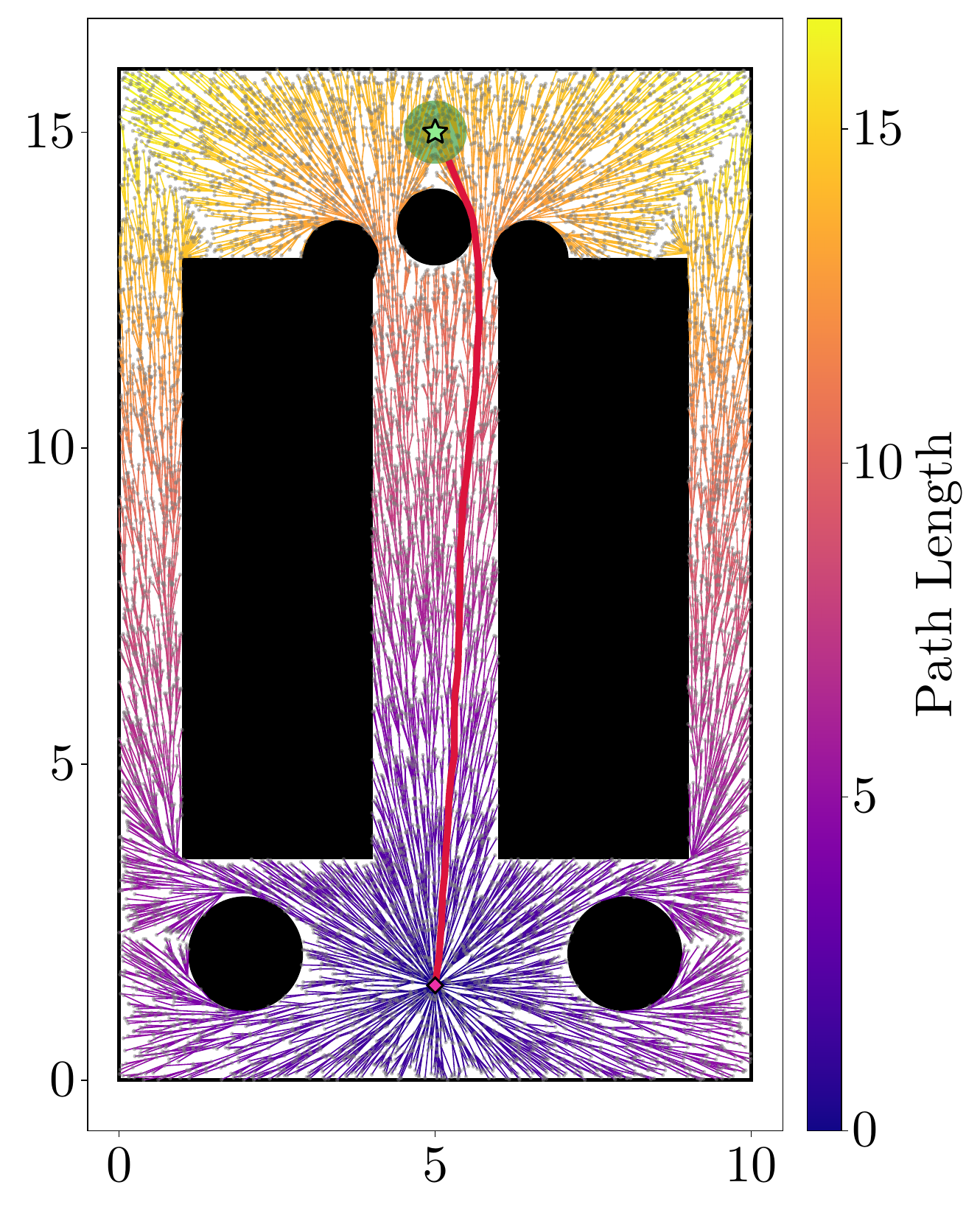} 
        \caption{RRT$^*$ with $\chi = 3$.} \label{sfig:funnelRRT}
        
    \end{subfigure}
    \hfil
    \begin{subfigure}[t]{0.3\textwidth}
        \centering
        \includegraphics[width=\linewidth]{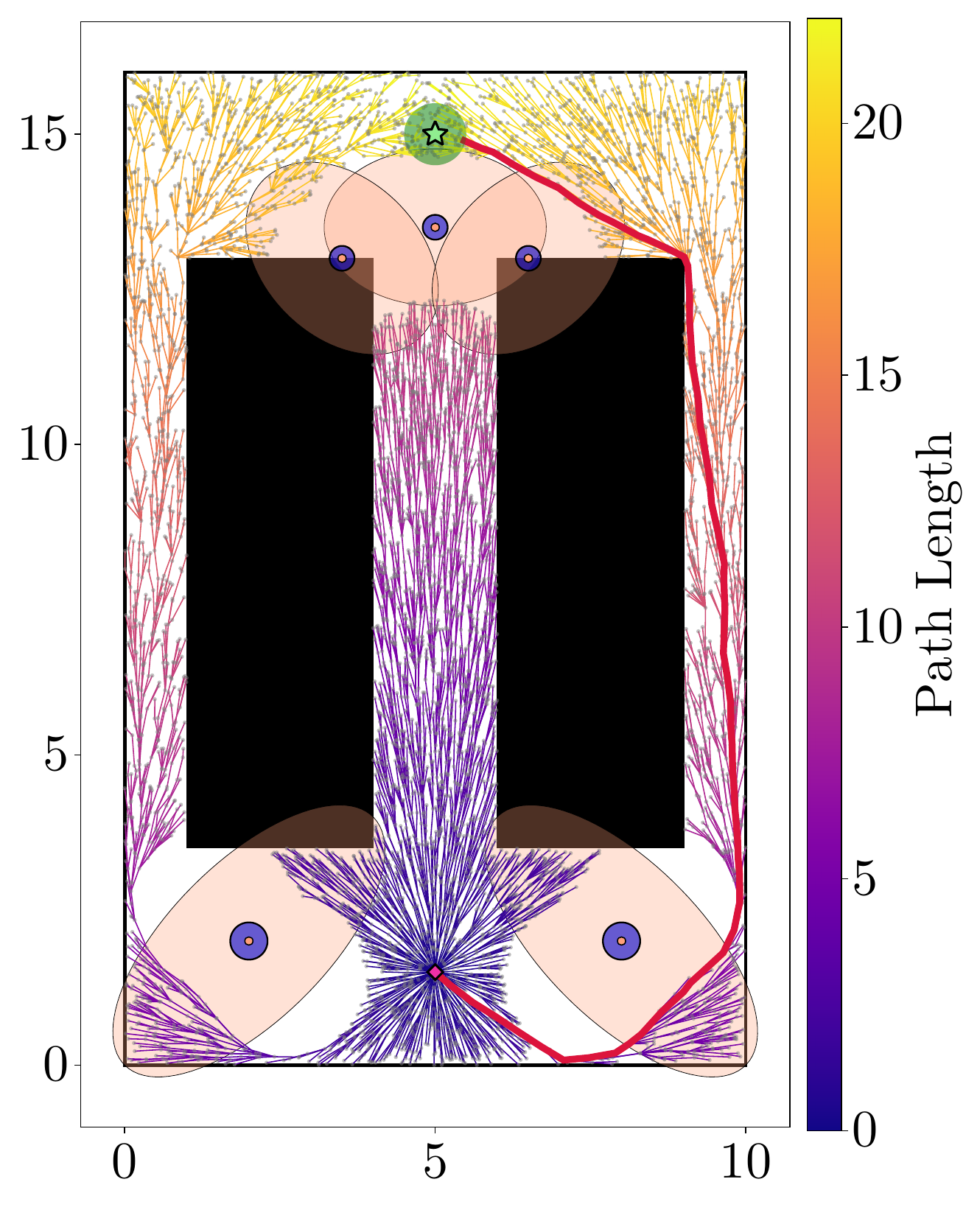} 
        \caption{CC-RRT$^*$, $\delta = 0.1$.} \label{sfig:funnelCCRRT}
    \end{subfigure}
    \hfil
    \begin{subfigure}[t]{0.3\textwidth}
        \centering
        \includegraphics[width=\linewidth]{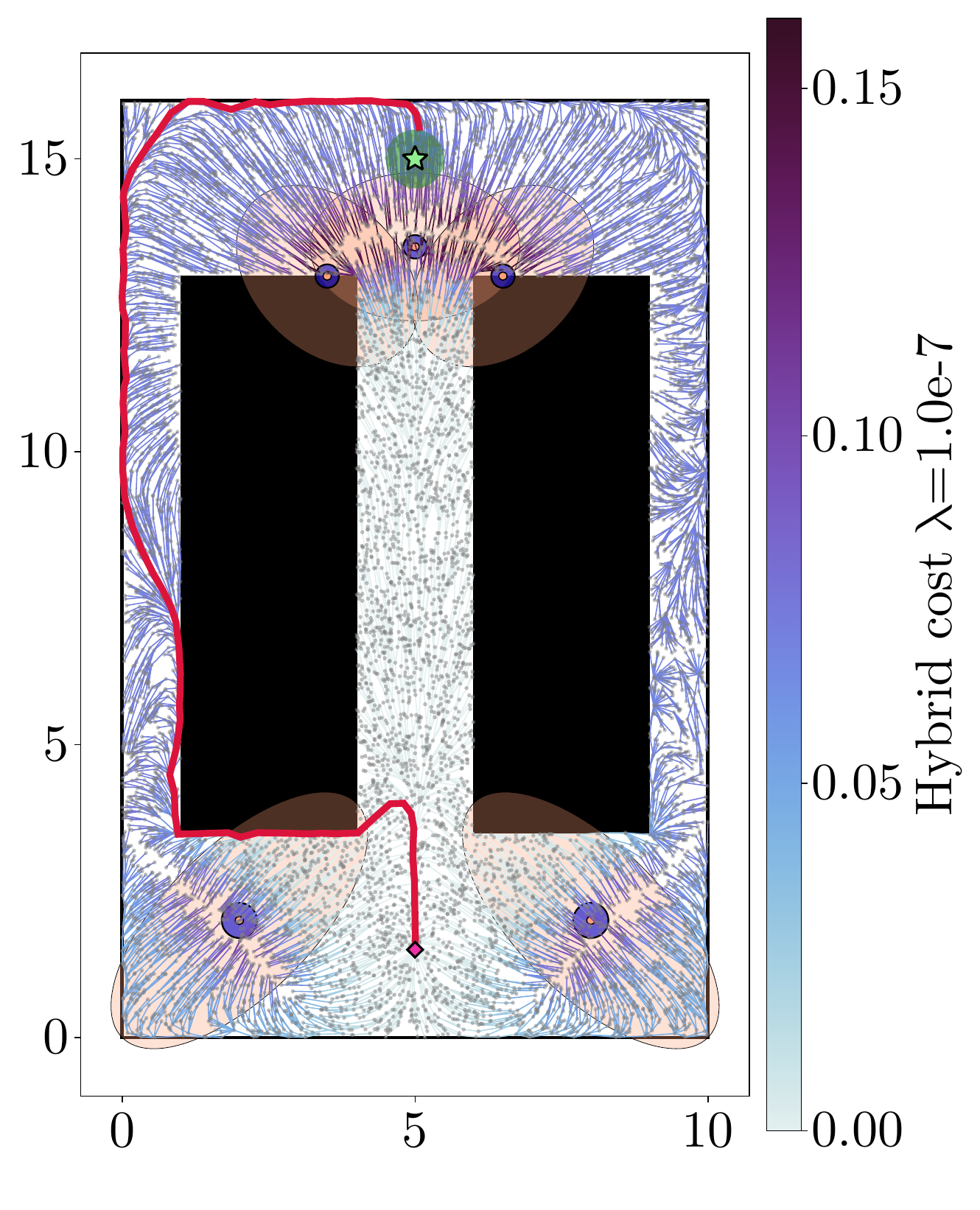} 
        \caption{RD-RRT$^*$.} \label{sfig:funnelRD}
    \end{subfigure}

    \caption{Planner behavior in the \enquote{Risk Trap Scenario}.}
    \label{fig:funnel_trap}
\end{figure}

The \enquote{Risk Trap Scenario} tests the planner's ability to escape deceptive local cost minima. The results, shown in \Cref{fig:funnel_trap}, highlight the fundamental differences between geometric, constraint-based, and global cost-optimization approaches.
\begin{itemize}[leftmargin=*]
    \item \textbf{Inflated RRT$^*$} exhibits myopic behavior. Guided by path length, it is drawn into the geometrically short funnel, finding a path with a length of only 13.2 but a collision probability of $\approx 24\%$. It fails to identify the safer global alternative.
    \item \textbf{CC-RRT$^*$} ($\delta=0.1$, the lowest tested threshold that produced a solution in this scenario) is not as easily trapped. It identifies the high risk within the funnel and correctly navigates around the central trap. However, its path represents a compromise, finding a route with a length of 22.4 and a collision probability of $\approx 15\%$. It is safer than the geometric path but has a higher estimated collision probability than the RD-RRT$^*$ path.
    \item \textbf{RD-RRT$^*$} finds the lowest estimated collision probability among the compared planners in this scenario. By integrating the RD proxy along the complete path, it favors the wide detour and returns a path with a collision probability of only $\approx 7\%$, albeit with the longest path length (23.7).
\end{itemize}

\subsection{Scalability} 
\begin{figure}[!ht]
    \centering
    \includegraphics[width=0.9\linewidth]{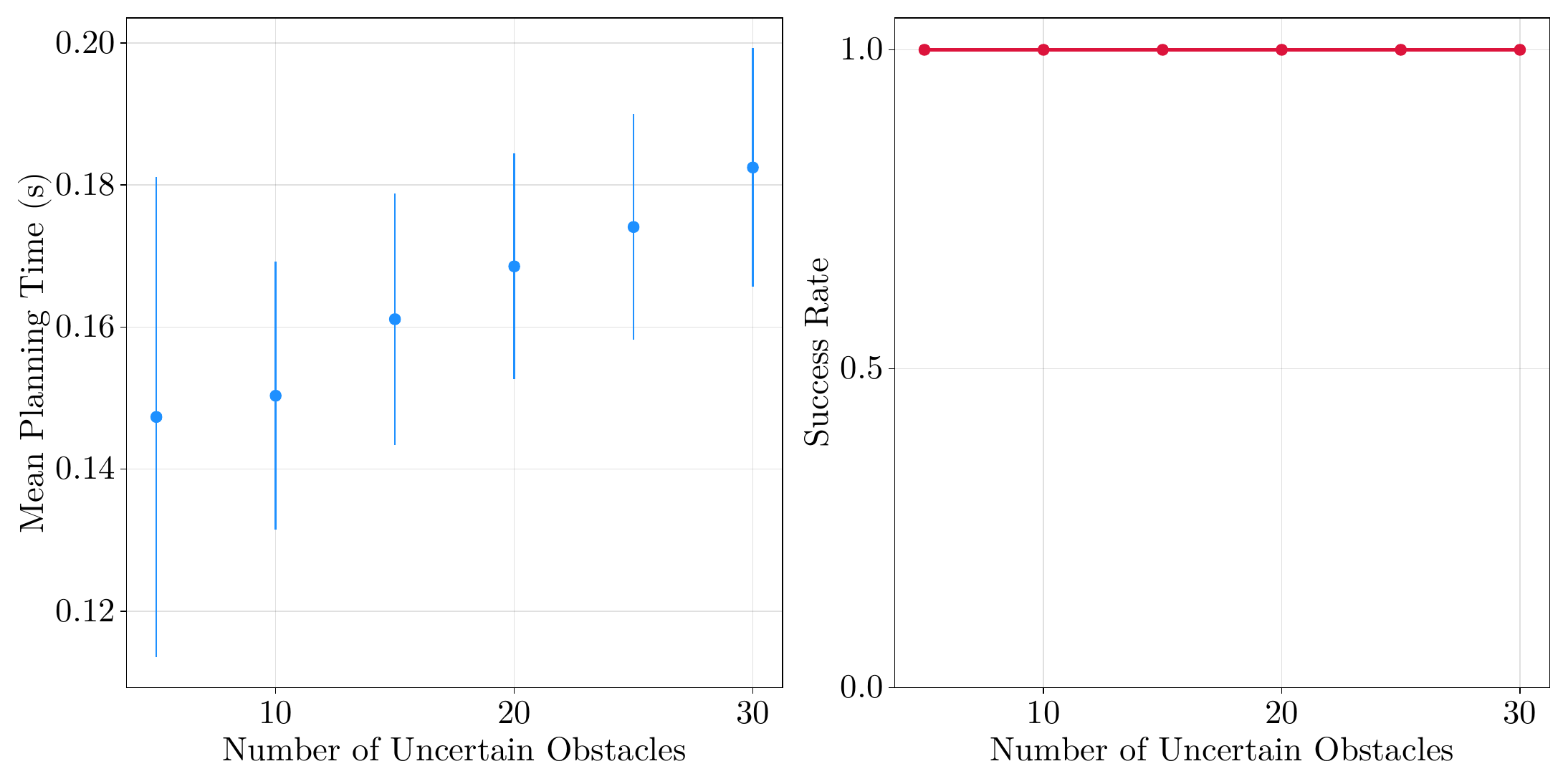}
    \caption{Mean planning time (\textbf{left}) and success rate (\textbf{right}) vs. the number of uncertain obstacles. The planner exhibits a favorable, near-linear time complexity and 100$\%$ success rate.}
    \label{fig:scalability1}
\end{figure}

The algorithm's computational performance was evaluated on complex random scenarios with an increasing number of uncertain obstacles. \Cref{fig:scalability1} shows that the mean planning time increases in a near-linear fashion, while the success rate remains at 100\%. This demonstrates that the algorithm scales robustly as the probabilistic complexity of the environment grows.

\begin{SCfigure}
    \centering
    \includegraphics[width=0.6\linewidth]{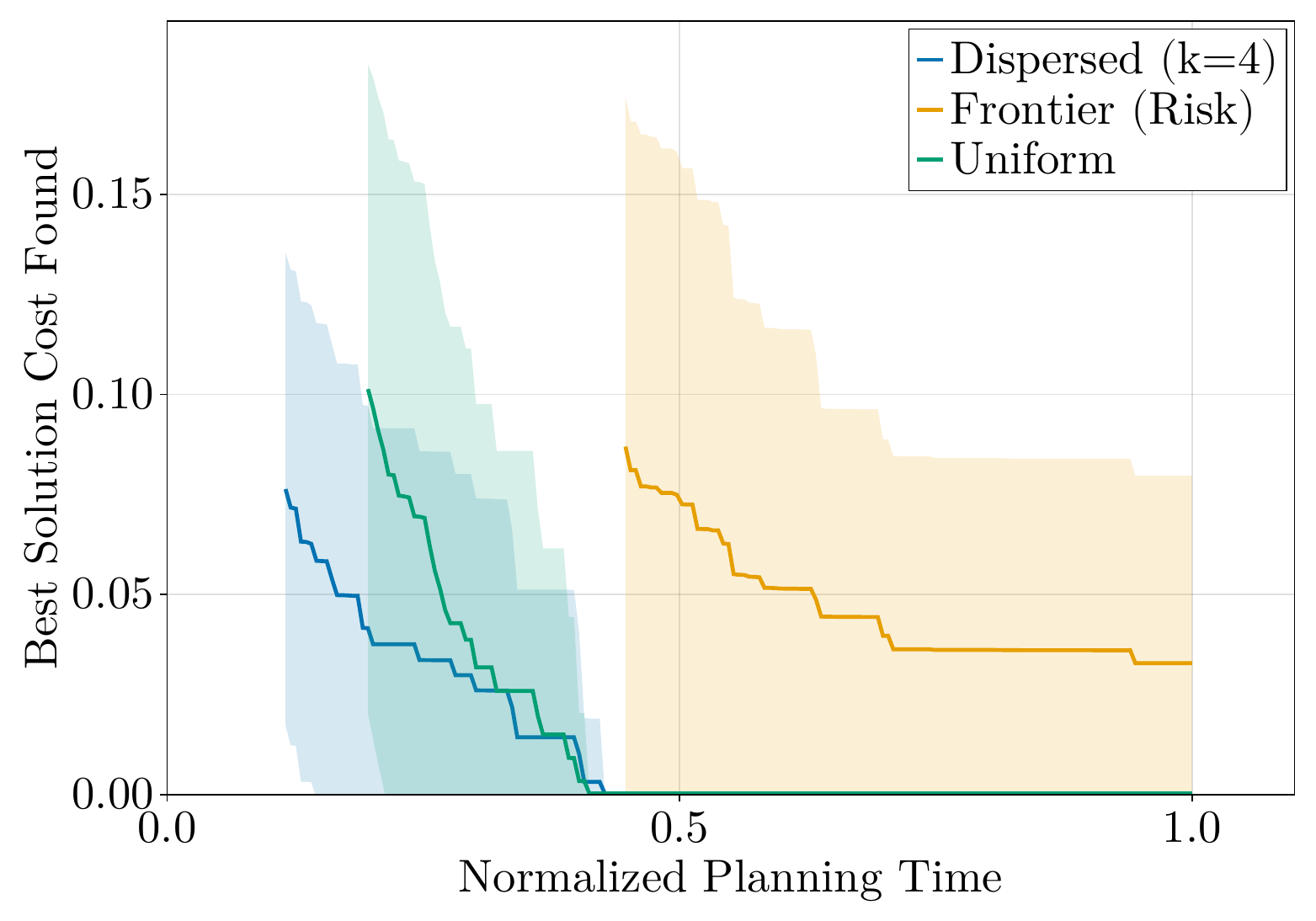}
    \caption{Anytime performance (best solution cost vs. normalized planning time) for different samplers in the \enquote{Local Minimum} scenario. Exploration-focused samplers (\texttt{Uniform}, \texttt{Dispersed}) converge faster and to better solutions than the risk-aware frontier sampler (here dubbed Frontier (Risk)). Shaded bands show the mean plus or minus one standard deviation across runs.}
    \label{fig:scalability2}
\end{SCfigure}

\subsection{Search Guidance} The effectiveness of different sampling strategies was compared across different scenarios. The results, exemplified by the convergence plot in \cref{fig:scalability2}, consistently show that simple, exploration-focused strategies like \texttt{UniformSampler} and \texttt{DispersedSampler} outperform the more complex, exploitation-focused \texttt{MahalanobisFrontierSampler}. The risk-heuristic sampler is often misled by the non-metric nature of the cost space and incurs significant computational overhead, leading to slower convergence and inferior solutions.

\subsection{Tree Topology} 
\begin{figure}[ht]
    \centering 
    \begin{subfigure}[t]{0.3\textwidth}
        \centering
        \includegraphics[height=\linewidth,angle=-90]{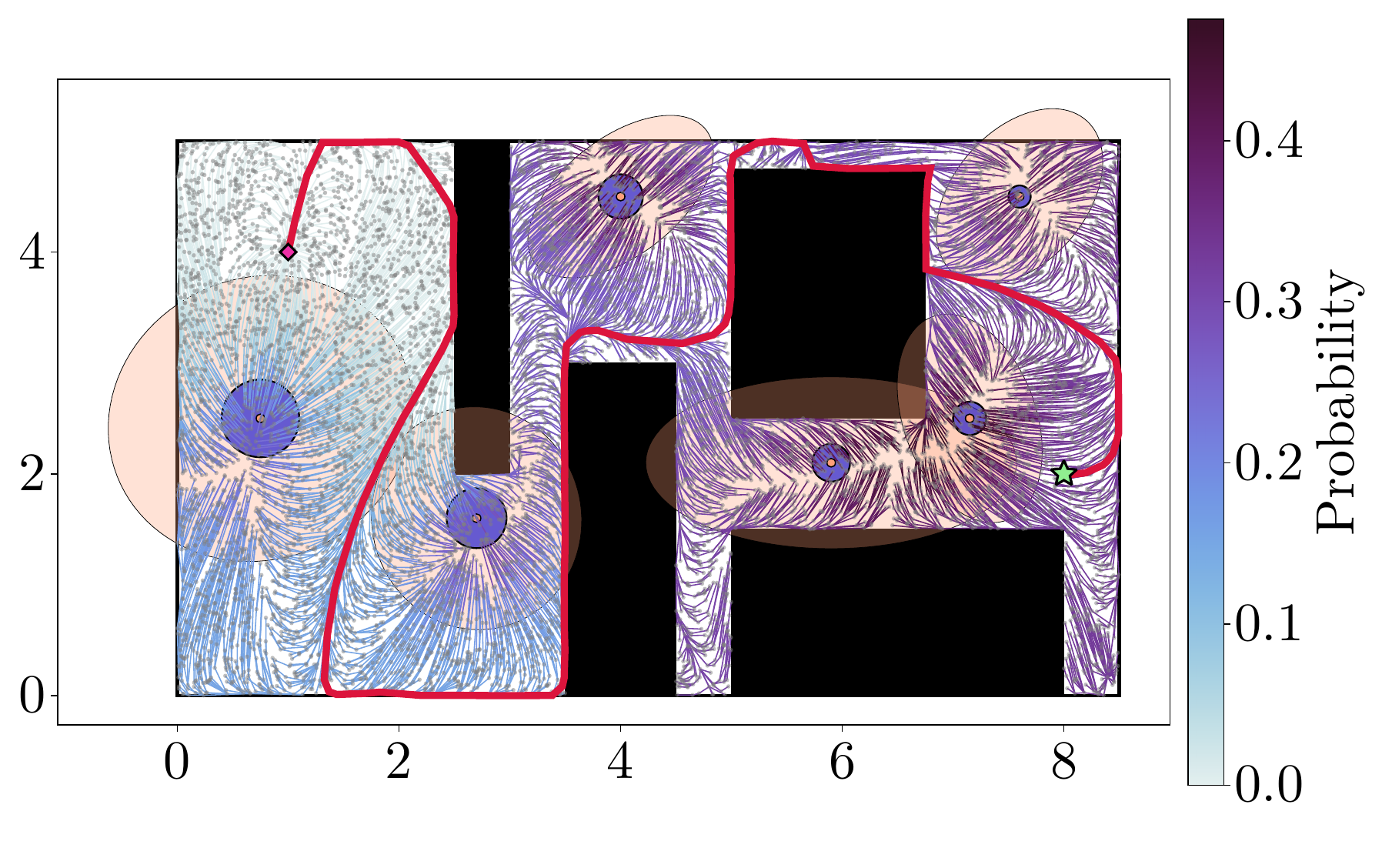}
        \caption{RD-RRT$^*$.} \label{sfig:defaultRD}
    \end{subfigure}
    \hfil
    \begin{subfigure}[t]{0.3\textwidth}
        \centering
        \includegraphics[height=\linewidth,angle=-90]{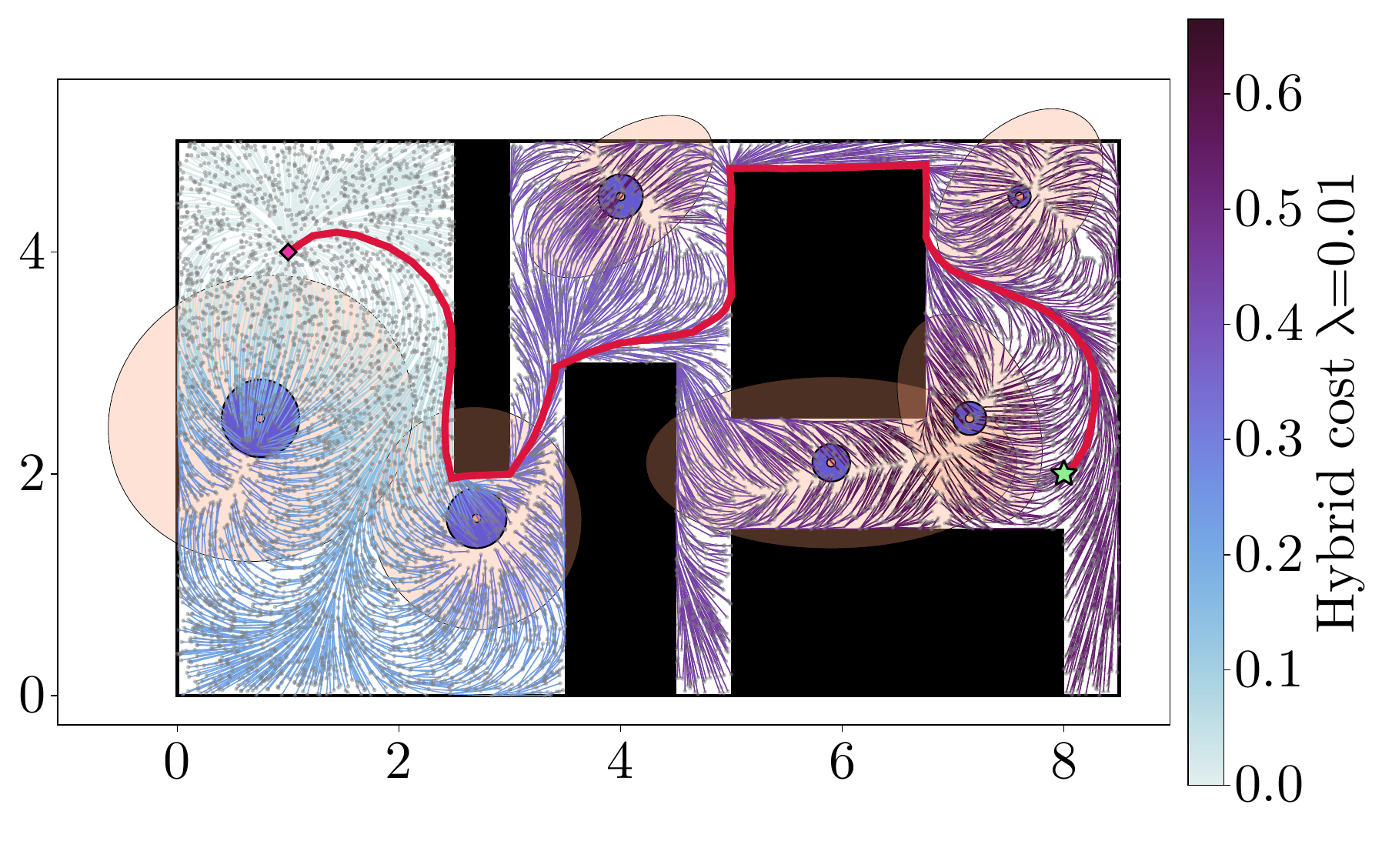}
        \caption{Hybrid Variant, $\lambda = 10^{-2}$.} 
        \label{sfig:defaultHybrid}
    \end{subfigure}
    \hfil
    \begin{subfigure}[t]{0.3\textwidth}
        \centering
        \includegraphics[height=\linewidth,angle=-90]{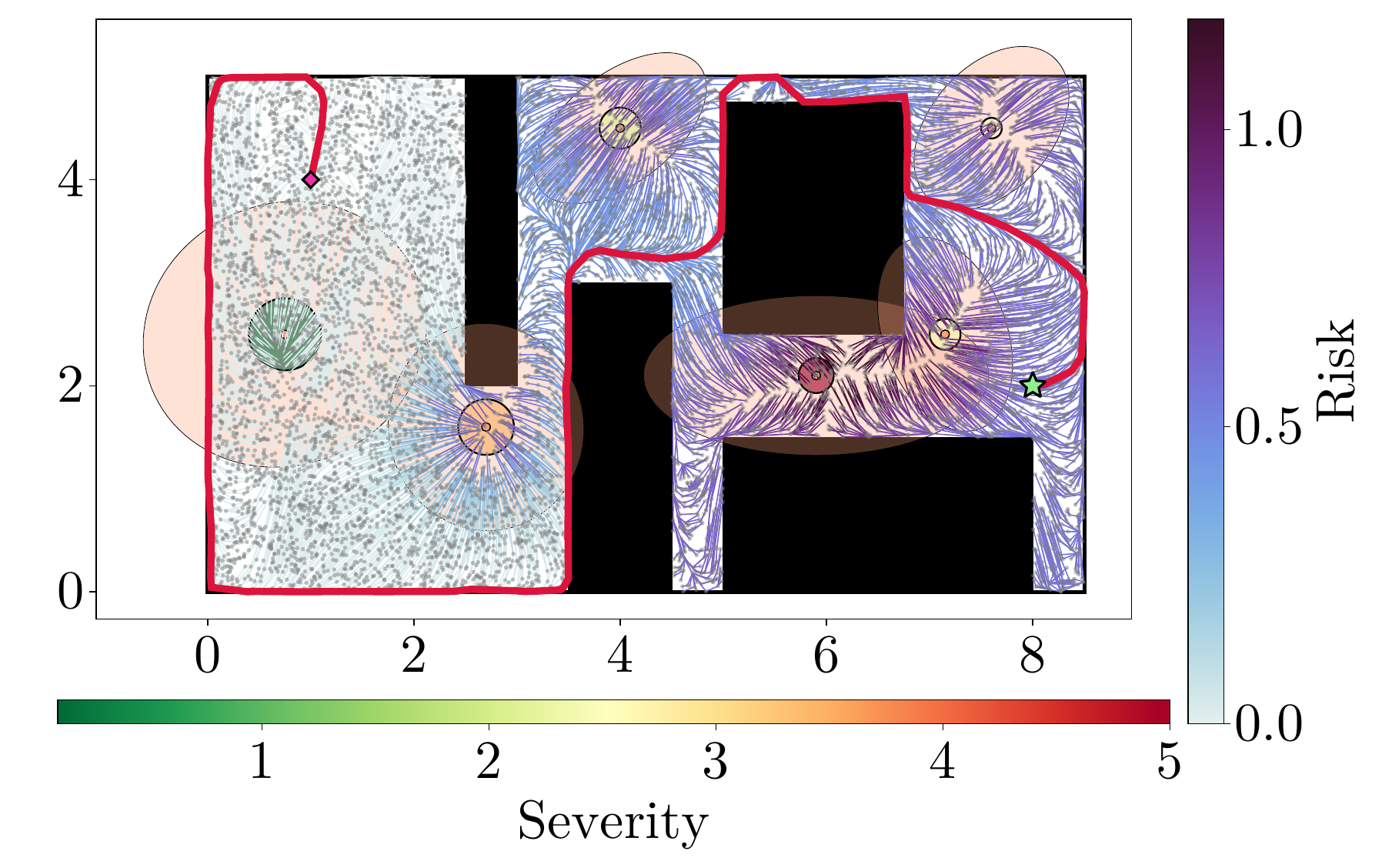}
        \caption{Risk Variant.} 
        \label{sfig:defaultRisk}
    \end{subfigure}
    \caption{Output $10^4$-node tree of the three variants of \cref{alg:RDRRT*} on the \enquote{Default Scenario}. The solutions are in red. The edges are colored with their corresponding costs. In the case of the Risk variant, the severities $\xi$ are colored according to the secondary colorbar.}
    \label{fig:Default}
\end{figure}

The tree constructed by \cref{alg:RDRRT*} varies greatly with the variant that constructs it, as visualized in \Cref{fig:Default}. The pure probability variant (\Cref{sfig:defaultRD}) explores diffusely in safe, low-cost regions, creating a structure that sharply avoids the high-cost areas around obstacles. The Hybrid variant (\Cref{sfig:defaultHybrid}), with its added length penalty, regularizes this exploration; it produces a more directed tree that prunes unnecessarily long paths in favor of shorter, more direct connections. Finally, the Risk variant (\Cref{sfig:defaultRisk}) demonstrates how user-defined severities create an anisotropic cost landscape. The tree is shaped by these preferences, steering more aggressively away from high-severity obstacles and thus reflecting a qualitative, context-dependent model of risk.

\subsection{Generalization Beyond 2D Gaussian Uncertainty}
\begin{figure}[ht]
    \centering 
    \begin{subfigure}[t]{0.36\textwidth}
        \includegraphics[clip,width=\linewidth]{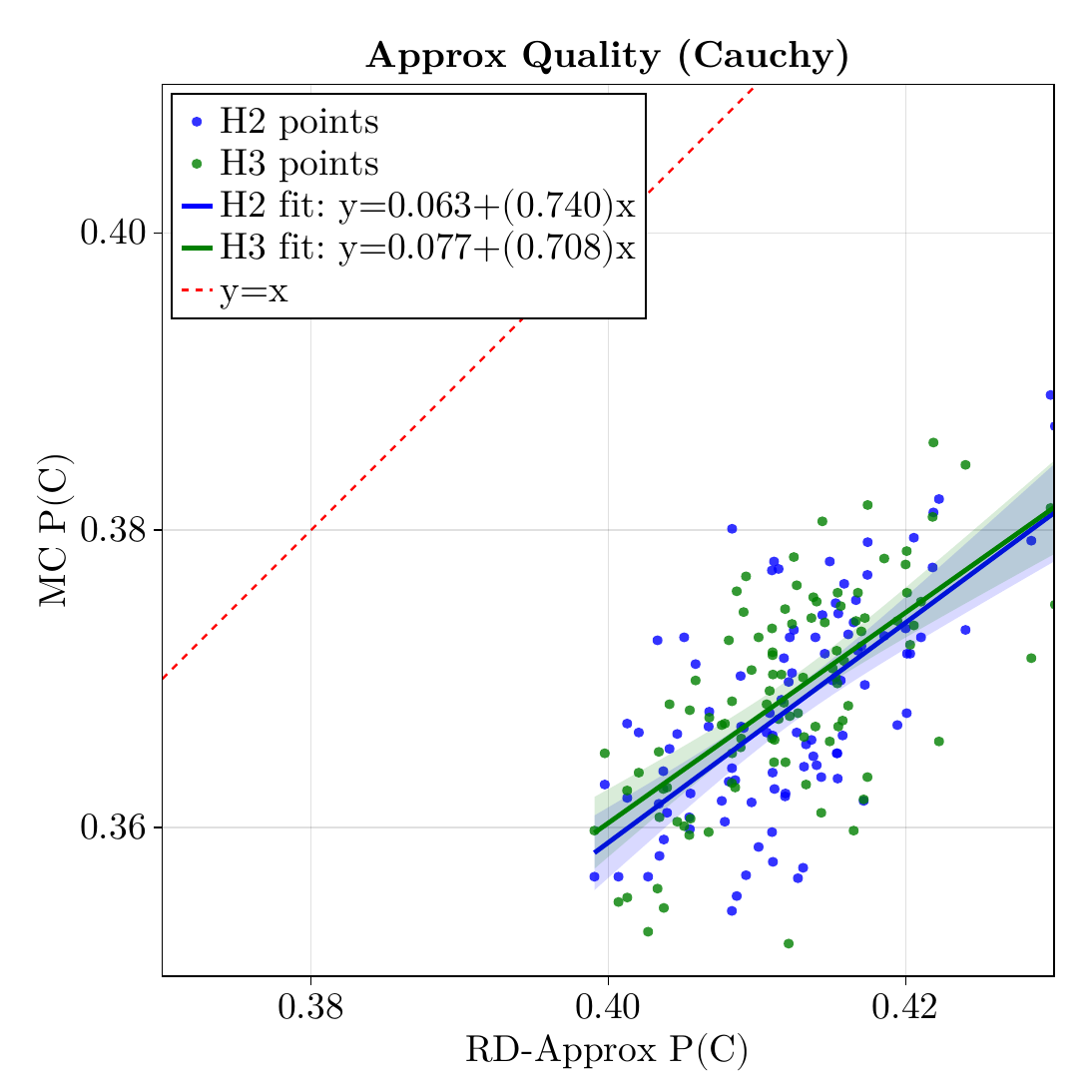}
        \label{sfig:cauchyaccuracy}
    \end{subfigure}
    \begin{subfigure}[t]{0.6\textwidth}
        \includegraphics[clip,width=\linewidth]{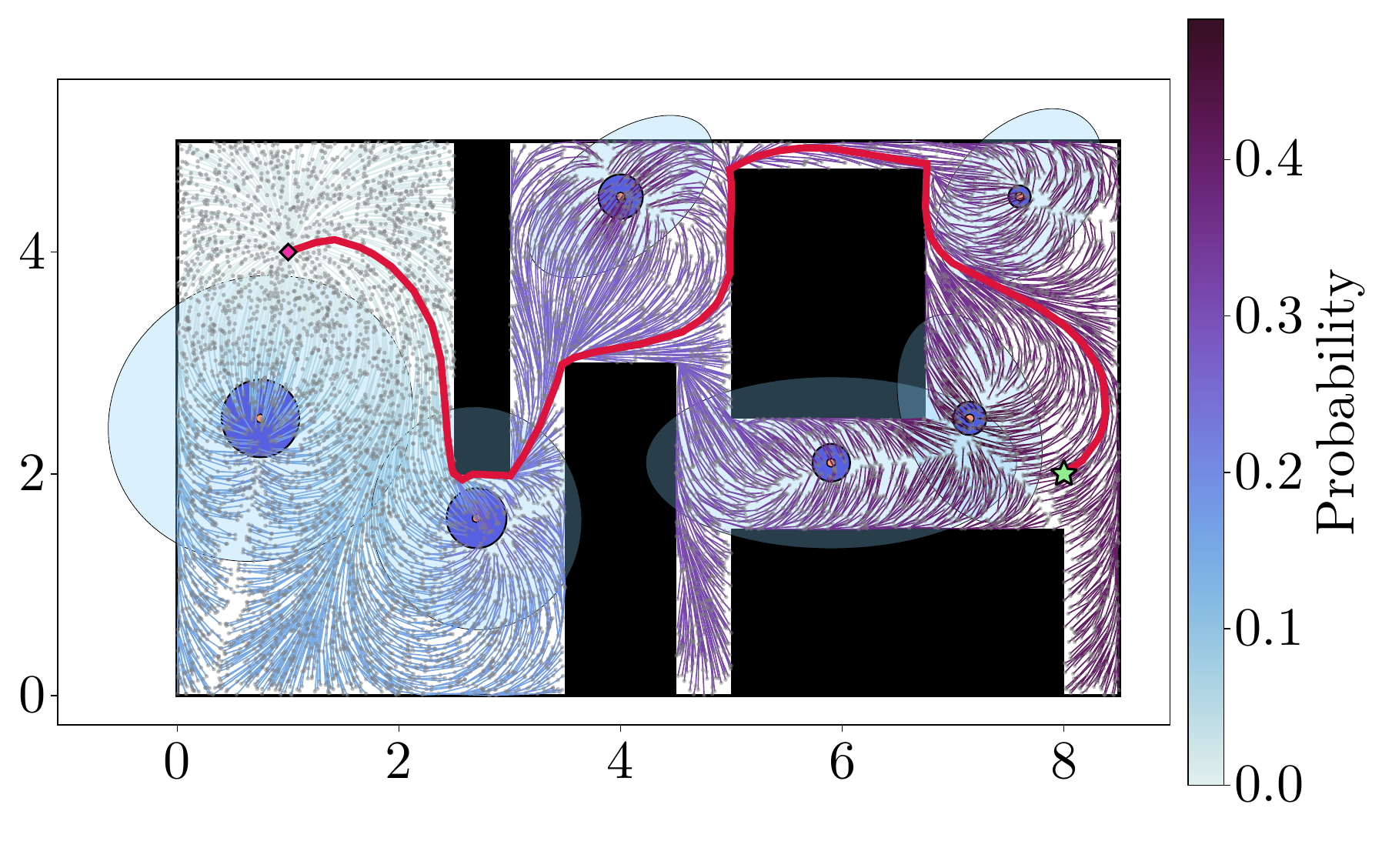}
        \label{sfig:cauchydefault}
    \end{subfigure}
    \caption{Cauchy-distribution benchmark. \textbf{(Left)} RD approximation against Monte Carlo collision probability under H2 and H3; shaded regions are 95\% confidence bands for the mean fitted regressions. \textbf{(Right)} Representative RD-RRT$^*$ planning result.}
    \label{fig:cauchy}
\end{figure}

The additional benchmarks evaluate Cauchy uncertainty, three-dimensional workspaces, obstacle decomposition, and the computational budget. \Cref{fig:cauchy}, moved into the main text, compares the Cauchy RD approximation with Monte Carlo collision probability under H2 and H3 and shows a representative planning result. The approximation study follows the Benchmark 1 conditions, while the representative tree uses 10,000 planning samples. Cauchy uncertainty preserves the applicability of the planner and provides guidance farther into the distribution tails, although its probability approximation is less tightly calibrated than the Gaussian counterpart. The detailed Cauchy regularization study remains in \cref{fig:fourcauchy} in the appendix.

\begin{figure}[ht]
    \centering 
    \begin{subfigure}[t]{0.48\textwidth}
        \includegraphics[clip,width=\linewidth]{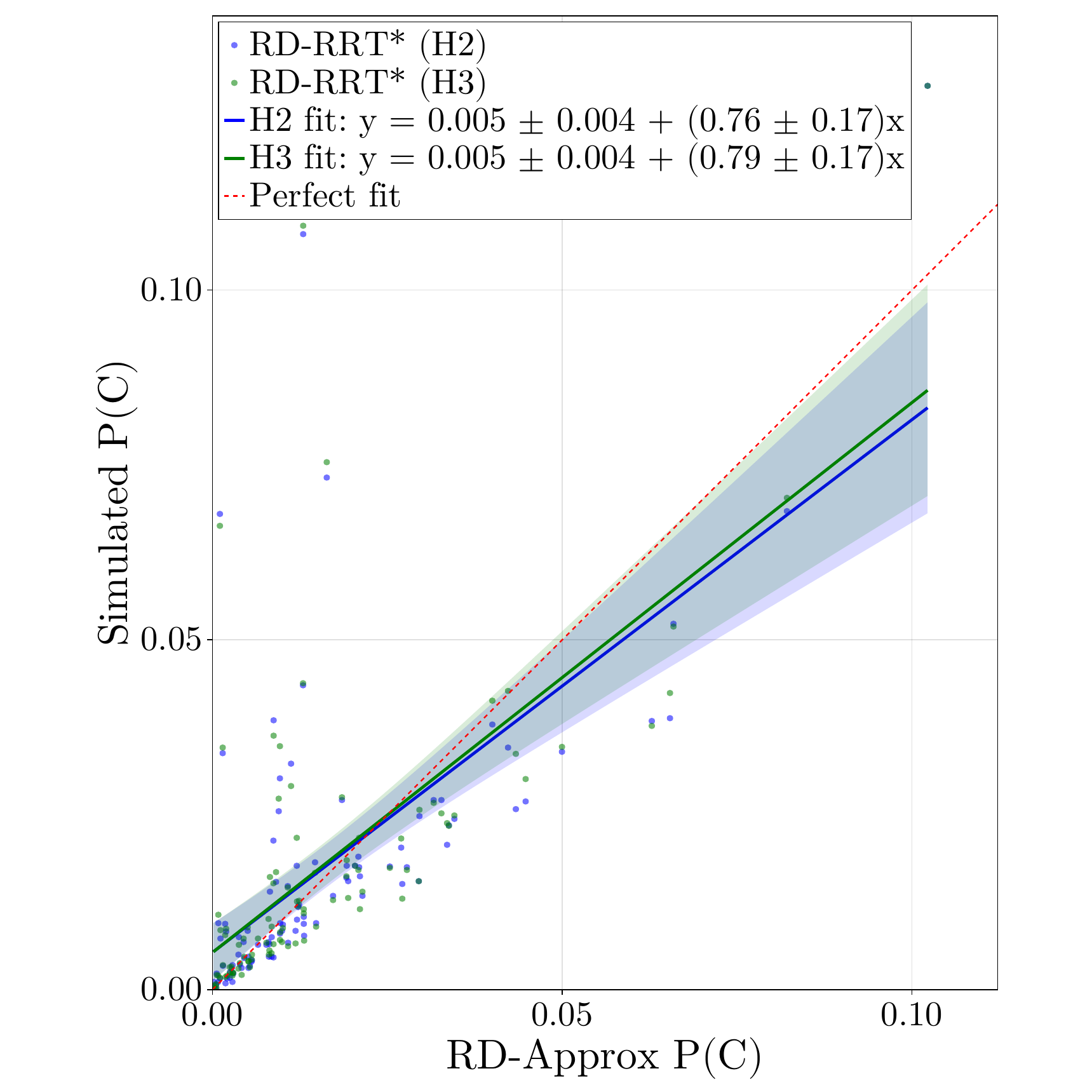}
        \label{sfig:3Daccuracy}
    \end{subfigure}
    \begin{subfigure}[t]{0.48\textwidth}
        \includegraphics[clip,width=\linewidth]{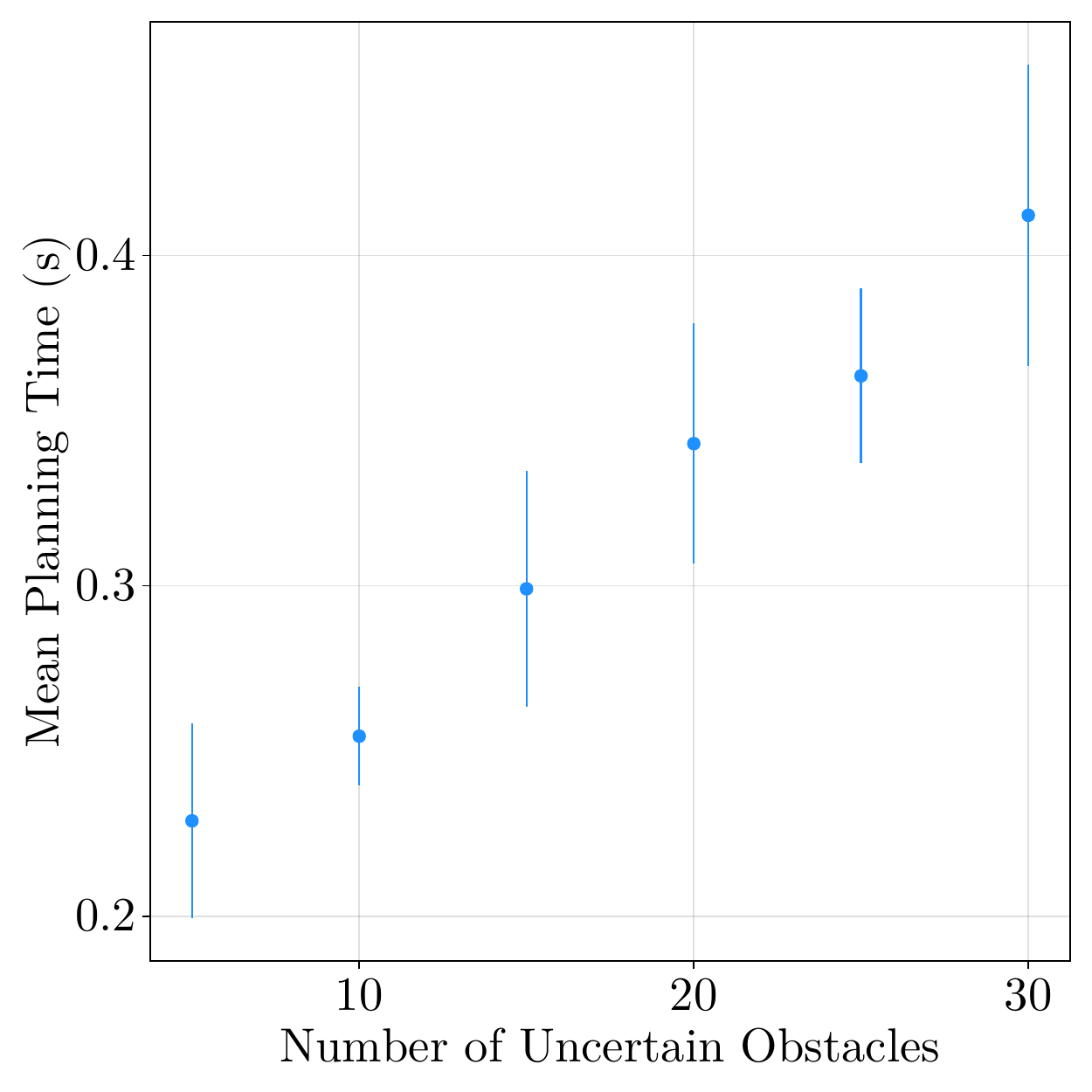}
        \label{sfig:3Dtime}
    \end{subfigure}
    \caption{Three-dimensional Gaussian benchmark. \textbf{(Left)} RD approximation against Monte Carlo collision probability; shaded regions are 95\% confidence bands for the mean fitted regressions. \textbf{(Right)} Mean planning time as the number of uncertain obstacles increases.}
    \label{fig:3Dbench}
\end{figure}

\begin{figure}
    \centering 
    \begin{subfigure}[t]{0.48\textwidth}
        \includegraphics[clip,width=\linewidth]{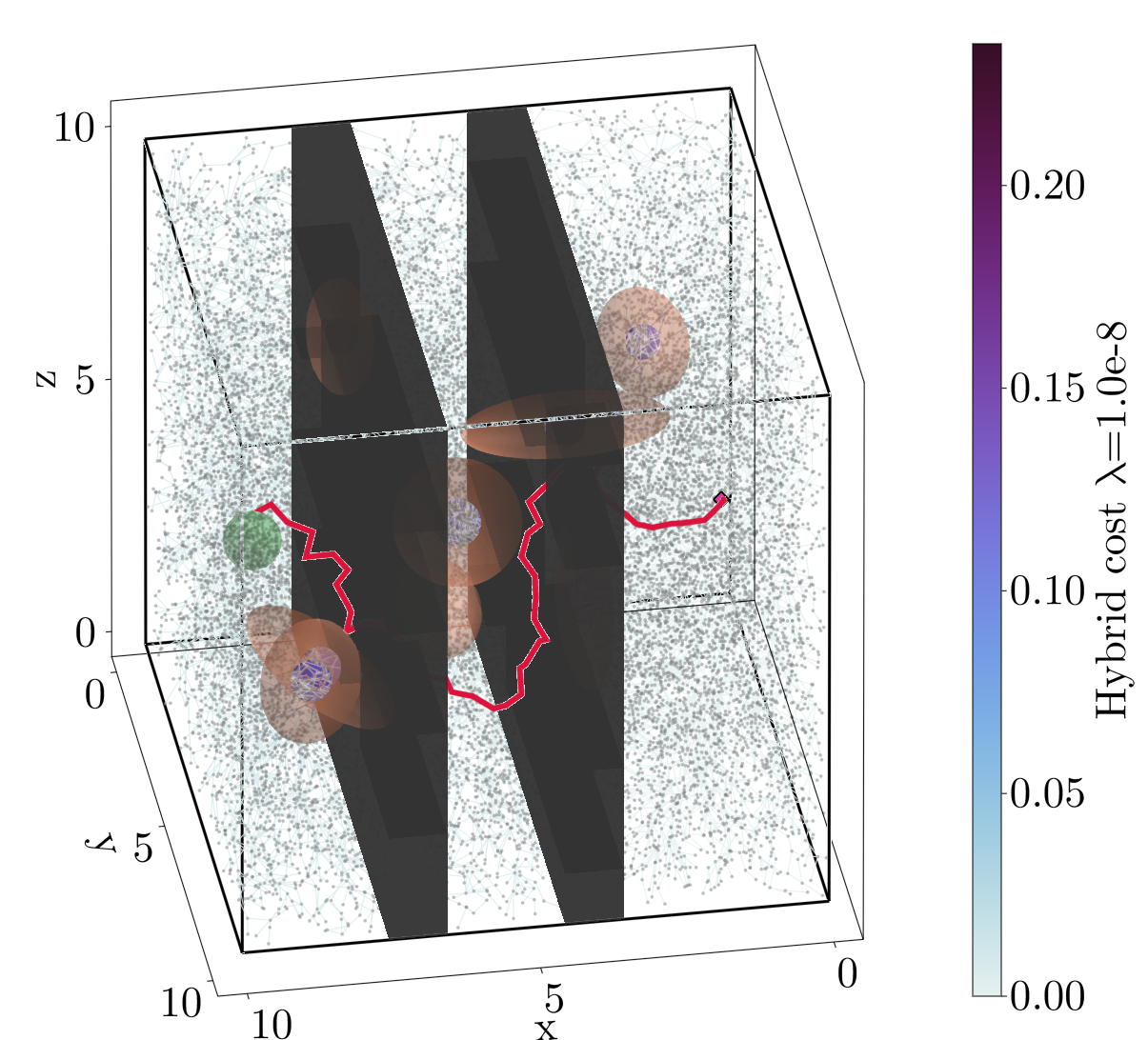}
        \label{sfig:3Dstandard}
    \end{subfigure}
    \begin{subfigure}[t]{0.48\textwidth}
        \includegraphics[clip,width=\linewidth]{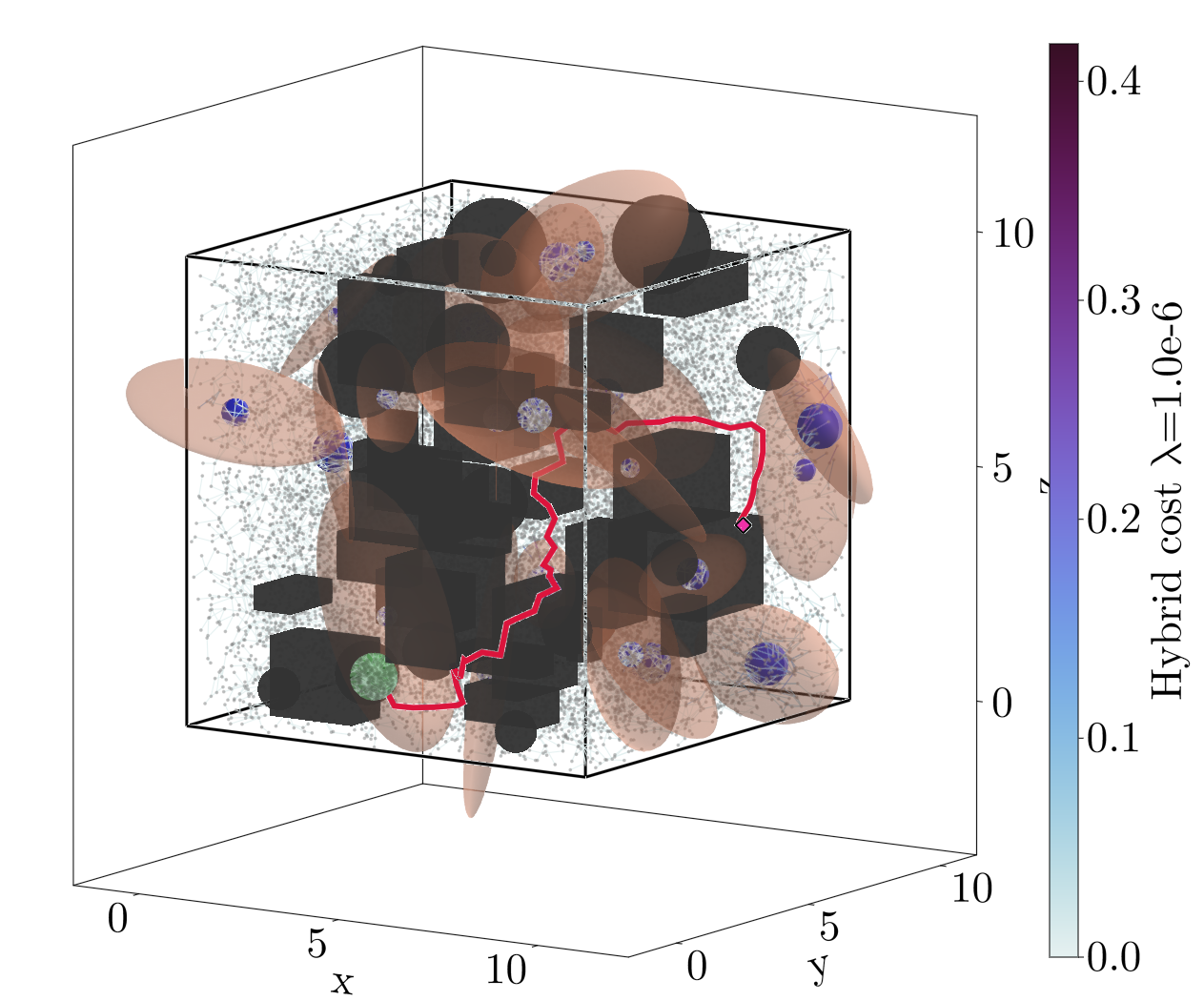}
        \label{sfig:3Dcomplex}
    \end{subfigure}
    \caption{\textbf{(Left:)} Result of the proposed algorithm on a \enquote{double slit} scenario. \textbf{(Right:)} Result on a randomly generated 3D scenario.}
    \label{fig:3Dscenarios}
\end{figure}

We evaluate the 3D Gaussian variant on randomly generated scenarios, with 100 runs for each tested number of uncertain obstacles. \Cref{fig:3Dbench} reports both approximation quality and planning-time scaling. Planning time remains approximately linear in the number of uncertain obstacles over the tested range. The linear relationship between the RD approximation and Monte Carlo probability is also retained, but with greater dispersion than in 2D. Near the origin, the approximation using \cref{eq:rd3D} more frequently underestimates $P(C)$, as indicated by points above the perfect-fit line. Representative 3D planning results are shown in \cref{fig:3Dscenarios}.

The obstacle-decomposition study and computational-budget details are provided in \cref{app:experiments} and \cref{app:computational_budget}, respectively.

\section{Discussion}
\label{sec:disc}
The empirical results highlight several key properties of the RD-RRT$^*$ algorithm and provide broader insights into risk-aware planning. We discuss the effectiveness of the approximation, the philosophical differences between planning paradigms, and practical design implications.

\subsection{The Efficacy of the Risk-Density Approximation}
A key finding is the high fidelity of the risk-density cost function (\Cref{fig:approximation_quality}). This effectiveness can be attributed to a self-reinforcing property of the algorithm. The cost, derived from a first-order Taylor expansion, is most accurate in the tails of the Gaussian obstacle distributions, where the probability density function is flatter. The planner's objective is to minimize the integral of this cost, which inherently penalizes paths that pass near the high-curvature peaks of the distributions (i.e., the obstacle means). Consequently, the search for a low-cost path naturally drives solutions into the very regions where the first-order approximation is most valid. 

\subsection{On the Empirical Convergence of Uncertainty Models}
A key finding from the simulations is the consistent agreement between the H2 (sequentially resolved uncertainty) and H3 (static uncertainty) ground-truth probabilities, as shown in \cref{tab:inflated-rrt,tab:cc-rrt} and \cref{fig:hybrid-rrt}. Notably, this holds true not only for the low-risk paths found by RD-RRT$^*$ but also for the high-risk paths generated by the baseline methods, which exhibit cumulative failure rates approaching 70\%.

We hypothesize that this occurs because the procedural difference between the H2 and H3 simulations—the conditional re-sampling in H2—is rarely triggered in practice. For any continuous path generated by a planner, even one with high cumulative risk, the probability of collision on any single, preceding segment remains low. This is because cumulative risk is typically not distributed uniformly; it is concentrated in \enquote{pinch points} or regions close to obstacles. Since planners generate paths composed of short, connected segments, the segment leading into a high-risk region is, by geometric necessity, located in a region of lower risk than the hotspot itself. Since the re-sampling in the H2 model is conditioned on such a low-probability event, it seldom occurs, and the simulation's behavior effectively defaults to that of the simpler H3 model.

This empirical convergence supports the theoretical foundation of the risk density functional. It demonstrates that $r_d$, which is derived from the point of theoretical equivalence between the models, serves as a robust measure of risk that is not overly sensitive to the specific choice between these two common stochastic process models.

\subsection{Global Risk Optimization vs. Local Filtering}
The \enquote{Risk Trap Scenario} highlights again a fundamental distinction between planning philosophies. Inflated RRT$^*$ treats risk as a simple geometric problem, which can perform poorly when geometric distance and safety are in conflict. Its optimization of path length leads it into the funnel in this scenario. CC-RRT$^*$ treats safety as a local, hard constraint---a filter on feasible edges. A series of locally \enquote{safe} decisions need not yield the lowest cumulative path risk because the method does not directly optimize that quantity from the root. In contrast, RD-RRT$^*$ treats its risk proxy as a pathwise soft cost. This allows the planner to compare risk trade-offs over complete candidate paths and, in the tested scenario, produces the path with the lowest estimated collision probability despite its greater geometric length.

\subsection{Practical Implications for Algorithm Design}
The experiments reveal two important practical considerations. First is the \textbf{necessity of regularization}. The pure probability cost is a semimetric, and in regions where the risk is low and flat, the mismatch between the Euclidean neighbor search and the cost can lead to \enquote{rewiring catastrophes} that stall exploration (\Cref{fig:rewiring_stall}). The hybrid cost's small Euclidean penalty acts as a crucial regularizer. Moreover, this regularization gives \enquote{guidance} to the algorithm where it may not have access to any, restoring robust exploration. Indeed, this phenomenon is evident in \Cref{fig:rewiring_stall}: without a hybrid formulation, the edges at the top of the deterministic obstacles evaluate to a cost within the floating-point rounding-error range because of the Gaussian's exponentially decaying tail, ultimately making exploration and wiring erratic.

Second is the \textbf{sufficiency of simple exploration}. Our results show that simple samplers like \texttt{UniformSampler} and \texttt{DispersedSampler} consistently outperform more complex, exploitation-focused heuristics (\Cref{fig:scalability2}). This suggests that for this problem, the quality of the cost function is more critical than the sophistication of the sampling strategy. 

\subsection{Physics-Based Execution Validation}
The MuJoCo study tests a different source of model mismatch. The planner uses conservative planar abstractions, while execution introduces a finite robot footprint, three-dimensional collision meshes, contact dynamics, and closed-loop path tracking. The reduced approximation quality in the 15-deterministic and 8-uncertain-obstacle random scenarios is unsurprising given both their greater geometric complexity and the difference between planning and execution models. Nevertheless, the geometric and MuJoCo collision frequencies remain in the same broad range as the RD predictions. This supports the use of RD-RRT$^*$ as a high-level planner while leaving perception uncertainty, dynamic planning, and physical-robot validation as future work.

\subsection{Expressiveness of the Risk Variant}
While the hybrid variant provides robustness, the risk-based variant using cost $\texttt{c}_r$ offers crucial expressiveness. By assigning different severities $\xi_j$ to obstacles, a user can encode context-dependent preferences. For example, a collision with a critical piece of infrastructure can be weighted more heavily than a collision with a transient, less important object. As shown in \Cref{sfig:defaultRisk}, this modification directly influences path selection, guiding the planner toward solutions with a lower severity-weighted RD proxy. This makes the planner adaptable to specific operational contexts and mission priorities.

\subsection{The Topology of Risk}
The trees generated by the RD-RRT$^*$ variants (\Cref{fig:Default}) offer insight into the topology induced on $\mathcal{C}_{free}$ by the risk-based cost functions. Each solution path is, by construction, the lowest-cost path within the explored graph. While the term \enquote{geodesic} is not formally justified since the semimetric cost violates the triangle inequality, the concept is powerful. The branches of the RRT$^*$ tree can be viewed as approximations of geodesics on a hypothetical Riemannian manifold whose metric would encode the risk. The search for a principled way to define this \enquote{risk-space} is a compelling direction for future work. Modeling such a space could eliminate the algorithm's dependence on heuristics to compute the neighborhood, and would give a more fundamental understanding of the geometry of this stochastic planning problem instance.

\subsection{Limitations}
The proposed algorithm has three primary limitations. First, we do not establish that the asymptotic-optimality guarantee of the original RRT$^*$ algorithm carries over to the combination of the modified objective, steering rule, and radius schedules used here. The standard decaying Euclidean radius can be applied as a heuristic, but doing so does not by itself provide that guarantee. Second, the current formulation does not account for the propagation of uncertainty through robot dynamics over time. Third, the planner implementation is currently limited to Euclidean spaces in which the task and the configuration space correspond. Extending it to non-Euclidean spaces is a key objective for future work.

\section{Conclusions}
\label{sec:concl}

This paper proposed a novel variant of RRT$^*$, the sampling-based planning algorithm RD-RRT$^*$, which searches for paths with low approximated collision probability. By leveraging a tractable, closed-form cost function derived from the risk density functional, the algorithm effectively navigates environments with both deterministic and uncertain obstacles.

The numerical validation suggests that directly optimizing the RD objective can produce paths with lower estimated collision probability than geometric or locally constrained alternatives in the tested scenarios. Closed-loop MuJoCo rollouts with a finite-footprint holonomic TidyBot further show that the predicted and simulated collision frequencies remain in a comparable range in more complex scenarios, although approximation quality is reduced relative to the simpler geometric benchmarks.

Future work will focus on implementation efficiency and on the current modeling limitations. Key directions include incorporating robot dynamics and time-varying uncertainty, adapting the method to higher-dimensional and non-Euclidean spaces, and extending the local planner to curved motion primitives, for example by using precomputed Gauss--Legendre quadrature nodes with curve-aware steering and collision checking. Integrating online observations and validating the complete planning and control stack on a physical mobile robot are also necessary steps beyond the MuJoCo study. In particular, physical-robot experiments and evaluations in high-dimensional articulated configuration spaces, such as those of robot arms, are needed to assess whether the benefits observed in the present simulations carry over to these settings. Finally, we plan to investigate the geometry induced by the RD path cost.

\section*{Acknowledgments}
\label{sec:ack}
\subsection*{Funding}
The author(s) declared that financial support was received for this work and/or its publication. This work was supported by the Italian Ministry of Research, under the complementary actions to the NRRP ``Fit4MedRob-FitforMedicalRobotics'' Grant(\#PNC0000007). Funded by the European Union under the Horizon Europe Programme through the FlexCycle project (Grant Agreement No. 101189600). This work was also supported by the EU Project euROBIN (Grant ID 101070596). Views and opinions expressed are, however, those of the author(s) only and do not necessarily reflect those of the European Union or the European Commission. Neither the European Union nor the granting authority can be held responsible for them.

\subsection*{Declaration of conflicting interests}
The author(s) declared no potential conflicts of interest with respect to the research, authorship, and/or publication of this article.

\subsection*{Declaration of generative AI use}
During the preparation of this work, the authors used OpenAI Codex to assist with manuscript drafting and editing and with the development, revision, and debugging of research software. These tools may introduce errors, bias, or omissions. All AI-assisted text and code remain subject to author review and verification. The authors take full responsibility for the final manuscript, including its accuracy, originality, and citations, and for the research software and the results produced using it.

\bibliographystyle{styles/SageH}
\bibliography{RiskRRTbib}

\appendix
\section{Generalized Uncertainty Models and 3D Extensions}
\label{app:additional}
In the main text, we derived the Risk Density (RD) cost and the associated RD-RRT$^*$ algorithm assuming a 2D workspace and Gaussian uncertainty.

Here we extend the Risk Density formulation to include the Cauchy distribution and derive closed-form edge costs for both Gaussian and Cauchy distributions in 3D workspaces.

\subsection{Radius Scaling in Three Dimensions}
\label{app:radius_scaling_3d}
In 2D, the swept cross-section is an interval, and the first-order approximation scales as $P^a(C,r)\approx r_d(\gamma_d)r$. In 3D, the swept cross-section is a disk: the first derivative with respect to $r$ vanishes at $r=0$, and the leading nonzero sensitivity is second order. With the Risk Density normalization used here, the corresponding approximation is
\begin{equation}
    P^a(C,r)\approx \frac{\pi}{2} r_d(\gamma_d)r^2.
    \label{eq:rd_3d_radius_scaling}
\end{equation}
Consequently, the 3D closed forms below evaluate the density functional $r_d=2\int_{\gamma_d}f(\mathbf{x})\,d\ell$ along an edge, while conversion to the collision-probability approximation uses the factor $\pi r^2/2$ for each spherical obstacle rather than the 2D factor $r$. For the full derivation see previous work \citep{paiolaEvaluationCollisionProbability2024a}.

\subsection{The Cauchy Distribution for Robust Planning}
While Gaussian models are ubiquitous in robotics, they decay exponentially, potentially underestimating risk in the \enquote{tails} of the distribution. The Cauchy distribution offers a heavy-tailed alternative that decays polynomially ($ \propto x^{-(k+1)} $), providing a more conservative risk estimate at greater distances from obstacles.

Independent elliptically contoured multivariate Cauchy variables remain within the same elliptical family under addition or subtraction when their shape matrices are proportional. Specifically, if $\Gamma_R=\sigma_R^2\Gamma_0$ and $\Gamma_O=\sigma_O^2\Gamma_0$, the difference of their centered random components, $\mathbf{W}_{d} = \mathbf{W}_R - \mathbf{W}_O$, satisfies
\begin{equation}
    \mathbf{W}_{d} \sim \mathcal{C}_a(\mathbf{0}, \Gamma_T)
\end{equation}
with $\Gamma_T=(\sigma_R+\sigma_O)^2\Gamma_0$ \citep{nolanMultivariateStableDistributions2024}. For non-proportional shape matrices, the difference remains a symmetric multivariate $1$-stable variable but is not generally elliptical, so it cannot in general be expressed using the density below. Under our planning assumption that robot-position uncertainty is negligible, $\sigma_R\to0$ and hence $\Gamma_T\to\Gamma_O$, recovering the Risk Density formulation used here. The difference of the respective modes, $\mathbf{d}_{RO} = \mathbf{x}_R - \mathbf{x}_O$, remains the mode of the relative-position distribution.

The Probability Density Function (PDF) for a $k$-dimensional elliptically contoured multivariate Cauchy distribution centered at the origin is given by:
\begin{equation}
    \mathcal{C}_a(\mathbf{x}; \Gamma) = \frac{\Gamma(\frac{k+1}{2})}{\pi^{\frac{k+1}{2}}\sqrt{|\Gamma|}} \left( 1 + \mathbf{x}^\top \Gamma^{-1} \mathbf{x} \right)^{-\frac{k+1}{2}}.
    \label{eq:cauchy_pdf_general}
\end{equation}

\subsection{Closed-Form Edge Costs for Linear Segments}
We again consider the evaluation of the Risk Density edge cost for a linear path segment connecting nodes $(\mathbf{x}_i, \mathbf{x}_{i+1})$ and for a single obstacle. The segment is parametrized by $s \in [0, 1]$:
\begin{equation}
    \gamma(s) = (\mathbf{x}_i - \mathbf{x}_O) + s(\mathbf{x}_{i+1} - \mathbf{x}_i).
\end{equation}
The length of the segment is $L = \|\mathbf{x}_{i+1} - \mathbf{x}_i\|$.
Under the conditions above, the distribution admits the following modified version of \cref{eq:riskdensity}, mutatis mutandis:
\begin{equation}
    \label{eq:newRD}
    r^C_d(\gamma_d) = 2\int_{0}^{1} \mathcal{C}_a(\gamma_d(s)|\mathbf{0},\Gamma_d)\left\|\frac{d \gamma_d}{d s}\right\| ds,
\end{equation}
but this does not by itself yield a closed-form edge cost. Fortunately, the PDF exponent of the Cauchy distribution is again a quadratic form in $s$, which can be expressed as
\begin{equation}
    Q(s) = (\gamma(s))^\top \Gamma^{-1} (\gamma(s)) = c_C s^2 + 2b_C s + a_C
\end{equation}
and, similarly to the Gaussian case, the coefficients are:
\begin{equation}
    a_C = \mathbf{v}^\top \Gamma^{-1} \mathbf{v}, \quad b_C = \mathbf{u}^\top \Gamma^{-1} \mathbf{v}, \quad c_C = \mathbf{u}^\top \Gamma^{-1} \mathbf{u},
\end{equation}
with 
\begin{equation}
    \mathbf{v} = \mathbf{x}_i - \mathbf{x}_O \ \text{and} \ \mathbf{u} = \mathbf{x}_{i+1} - \mathbf{x}_i.
\end{equation}

\subsubsection{2D Cauchy Edge Cost}
For a 2D workspace ($k=2$), the exponent in \eqref{eq:cauchy_pdf_general} is $-3/2$ and the normalization constant is $1/(2\pi\sqrt{|\Gamma|})$. Substituting this into the Risk Density definition, we obtain:
\begin{equation}
    r_d^{C2D} = \frac{2L}{2\pi\sqrt{|\Gamma|}} \int_0^1 \left( 1 + c_Cs^2 + 2b_Cs + a_C \right)^{-3/2} ds
\end{equation}
This integral admits a purely algebraic closed-form solution:
\begin{equation}
    r_d^{C2D} = \frac{L}{\pi \sqrt{|\Gamma|}} \left[ \frac{c_Cs + b_C}{(c_C(a_C+1) - b_C^2)\sqrt{c_Cs^2 + 2b_Cs + a_C + 1}} \right]_{0}^{1}
\end{equation}
Notably, this formulation avoids the transcendental $\erf(\cdot)$ required for the Gaussian case, making the Cauchy cost slightly cheaper to evaluate computationally.

\subsubsection{3D Gaussian Edge Cost}
For a 3D workspace ($k=3$), the Gaussian PDF normalization is $(2\pi)^{-3/2} |\Sigma|^{-1/2}$. The Risk Density integral becomes:
\begin{equation}
    r_d^{G3D} = \frac{2L}{\sqrt{(2\pi)^3 |\Sigma|}} \int_0^1 \exp\left( -\frac{1}{2}(cs^2 + 2bs + a) \right) ds
\end{equation}
Completing the square in the exponent and applying the definition of the Error Function, we derive:
\begin{equation}
    \label{eq:rd3D}
    r_d^{G3D} = \frac{L}{2\pi \sqrt{c |\Sigma|}} \exp\left( \frac{b^2 - ac}{2c} \right) \left[ \erf\left( \frac{cs+b}{\sqrt{2c}} \right) \right]_0^1
\end{equation}
Note the change in the normalization pre-factor compared to the 2D case in \cref{eq:closedformRD}. 

\subsubsection{3D Cauchy Edge Cost}
For the 3D Cauchy case ($k=3$), the exponent in \eqref{eq:cauchy_pdf_general} becomes $-(3+1)/2 = -2$. The normalization constant is $1/(\pi^2 \sqrt{|\Gamma|})$. The edge cost integral is:
\begin{equation}
    r_d^{C3D} = \frac{2L}{\pi^2 \sqrt{|\Gamma|}} \int_0^1 \frac{ds}{(c_Cs^2 + 2b_Cs + A)^2}
\end{equation}
where $A = a_C + 1$. We utilize the reduction formula for integrals of the form $\int Q(s)^{-2} ds$. Let $\Delta = 4c_CA - 4b_C^2 = 4(c_C(a_C+1) - b_C^2)$. The closed-form solution is composed of a rational term and an arctangent term:
\begin{align}
    r_d^{C3D} = \frac{2L}{\pi^2 \sqrt{|\Gamma|}} \Bigg[ & \frac{2c_Cs+2b_C}{\Delta (c_Cs^2+2b_Cs+A)} + \nonumber \\
    & \frac{4c_C}{\Delta^{3/2}} \arctan\left(\frac{2c_Cs+2b_C}{\sqrt{\Delta}}\right) \Bigg]_0^1
\end{align}
Unlike the 2D Cauchy case, the 3D formulation is transcendental due to the arctangent term, yet it remains fully explicit and differentiable.

\begin{remark}
    Because we assume conditional independence across obstacles \cref{eq:conditionalindependence}, each obstacle may be assigned a different uncertainty model (e.g., Gaussian for one, Cauchy for another) without changing the estimator.
\end{remark}

\section{Additional Experiments}
\label{app:experiments}

\subsection{Decomposition Analysis}

\begin{figure}
    \centering
    \includegraphics[width=0.7\linewidth]{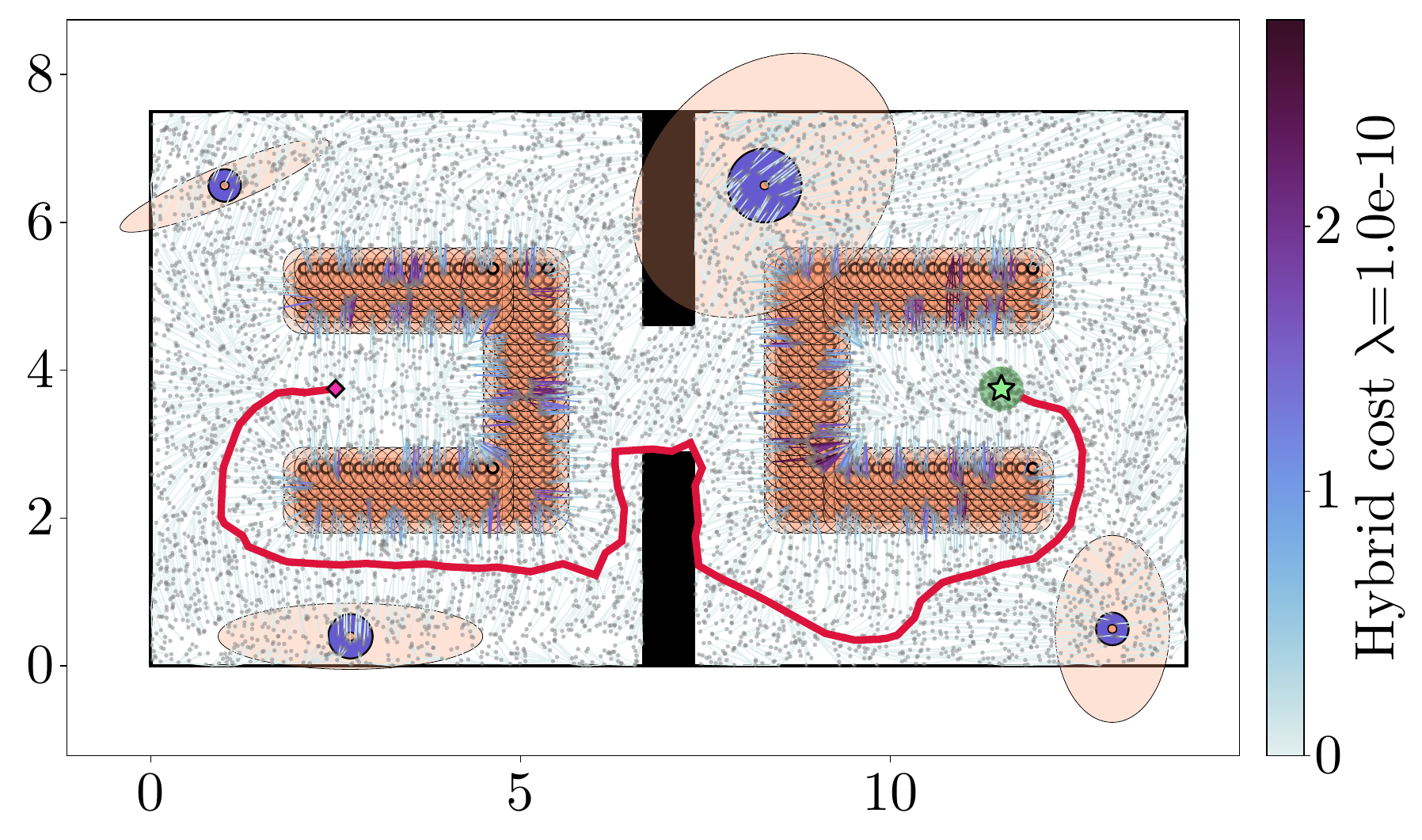}
    \caption{Example of the decomposition scenario. Here the disk radius is $r=0.075$.}
    \label{fig:lshapedecomposition}
\end{figure}

We study here the effects of the conditional independence assumption \cref{eq:conditionalindependence} on the accuracy of the proposed approximation. We construct a scenario in which we decompose two \enquote{C}-shaped uncertain obstacles into a number of uncertain disks while maintaining a common covariance; an example of this scenario, along with a solution, is shown in \cref{fig:lshapedecomposition}. The shapes of the nominal scenario are decomposed into disks with radii in the range $[0.05, 0.2]$.

\begin{figure}
    \centering
    \includegraphics[width=0.7\linewidth]{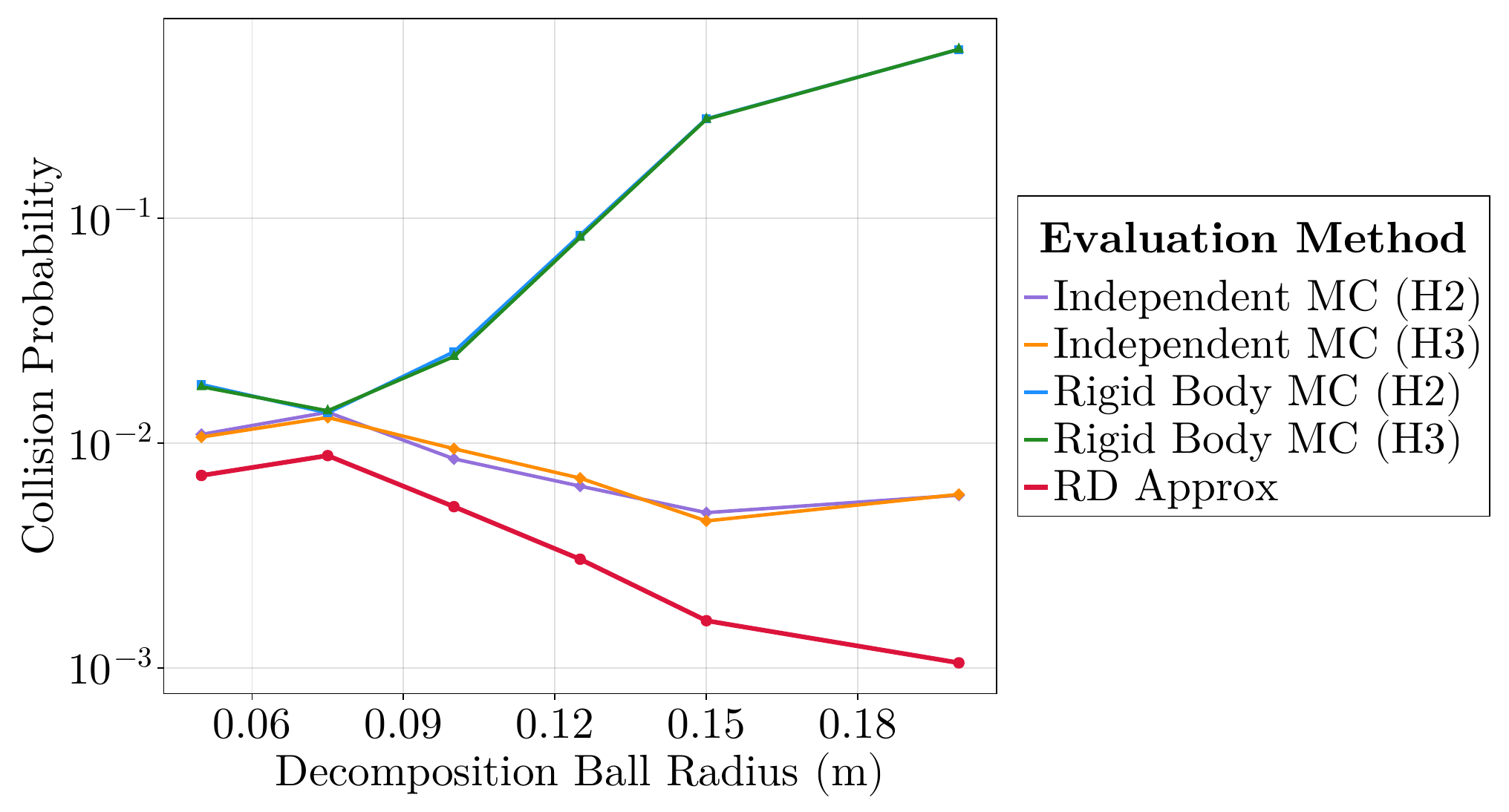}
    \caption{Decomposition Ball Radius against Predicted P(C).}
    \label{fig:lshaperadius}
\end{figure}

To study the accuracy of the approximation given by \cref{eq:rdformultiple} in this scenario, Monte Carlo simulations are run both following the conditional independence assumption (i.e. each ball is checked individually) and on the actual C shape with a random translation (which will serve as ground truth).  By comparing these measures, computed on the solution path given by \cref{alg:RDRRT*}, against the proposed approximation, some conclusions can be drawn. As shown in \cref{fig:lshaperadius}, while at lower values of the decomposition radius the $P(C)$ predictions cluster, coarser decompositions lead to worse approximations, shown by the green and blue lines moving significantly away from the rest of the predictions. Notice also how the RD approximation follows the same trend as the MC approximations which adhere to the conditional independence assumption.

\subsection{Cauchy Distribution Benchmarks}

The approximation-quality and representative-planning results for Cauchy uncertainty are presented in the main text in \cref{fig:cauchy}. This appendix retains the more detailed study of how Cauchy uncertainty interacts with regularization in the Local Minimum Scenario.

\begin{figure}
     \centering
     \begin{subfigure}[t]{0.45\textwidth}
         \centering
         \includegraphics[width=\textwidth]{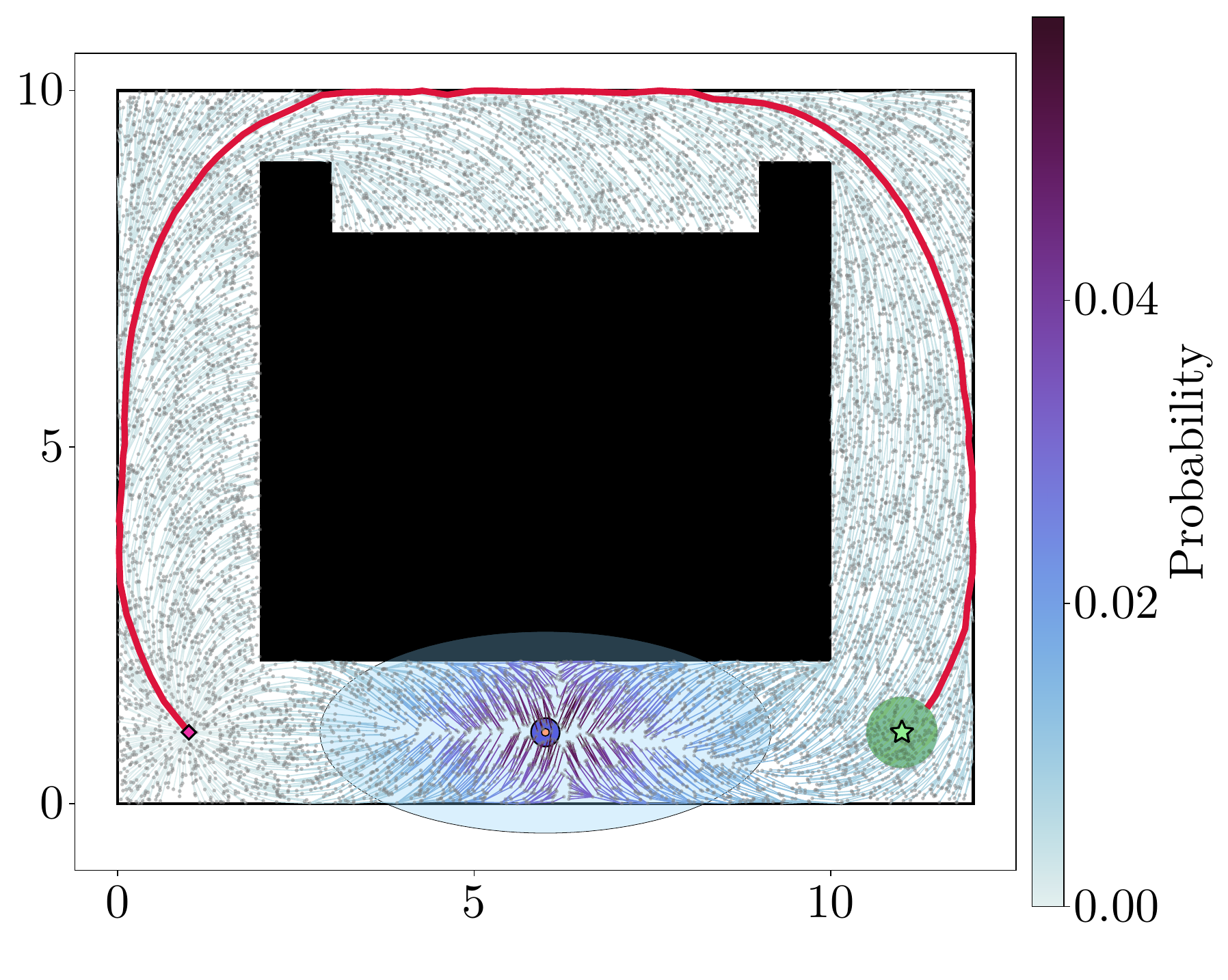}
         \caption{No regularization.}
     \end{subfigure}
     \hfill
     \begin{subfigure}[t]{0.45\textwidth}
         \centering
         \includegraphics[width=\textwidth]{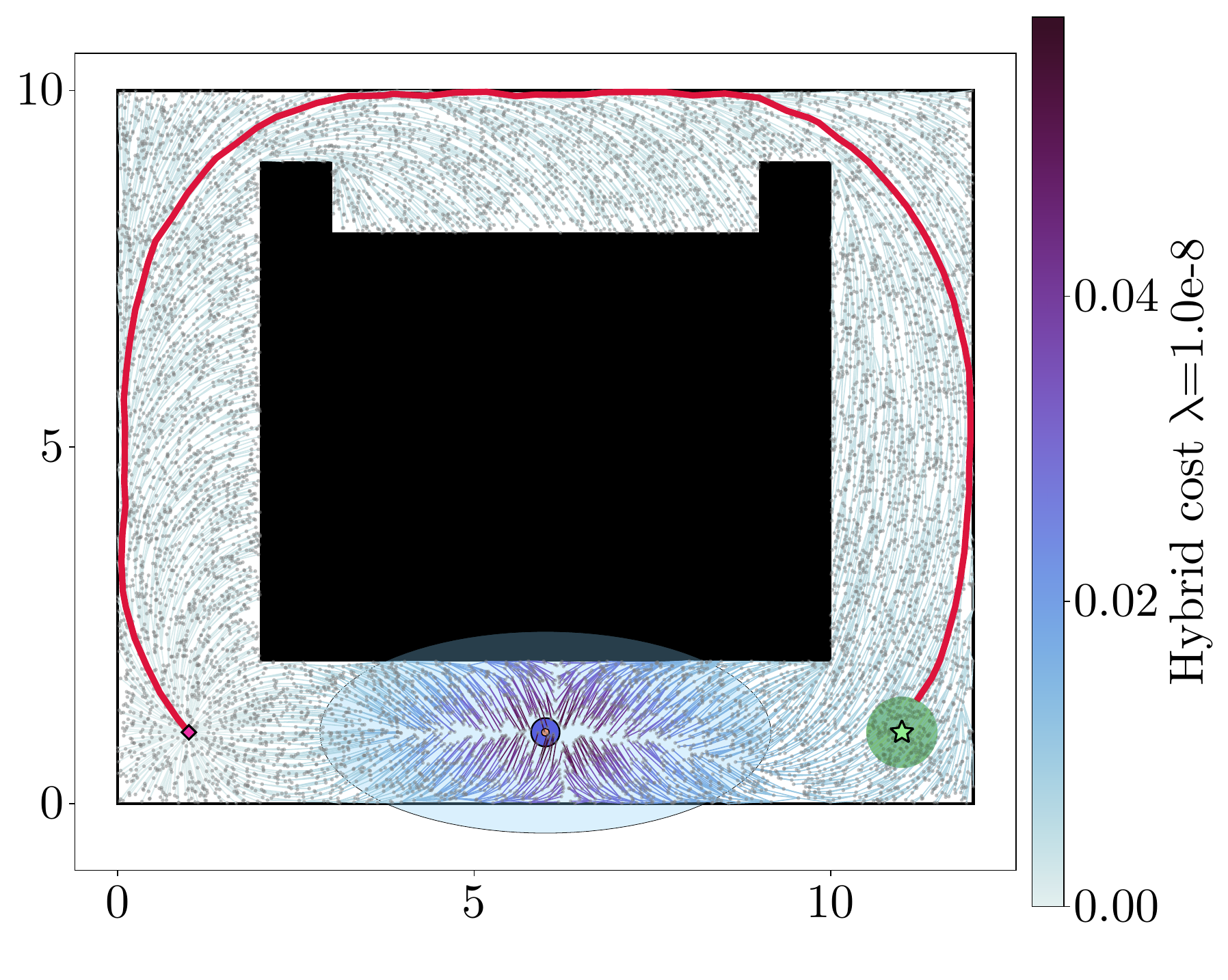}
         \caption{$\lambda = 10^{-8}$}
     \end{subfigure}

     \vskip\baselineskip %

     \begin{subfigure}[t]{0.45\textwidth}
         \centering
         \includegraphics[width=\textwidth]{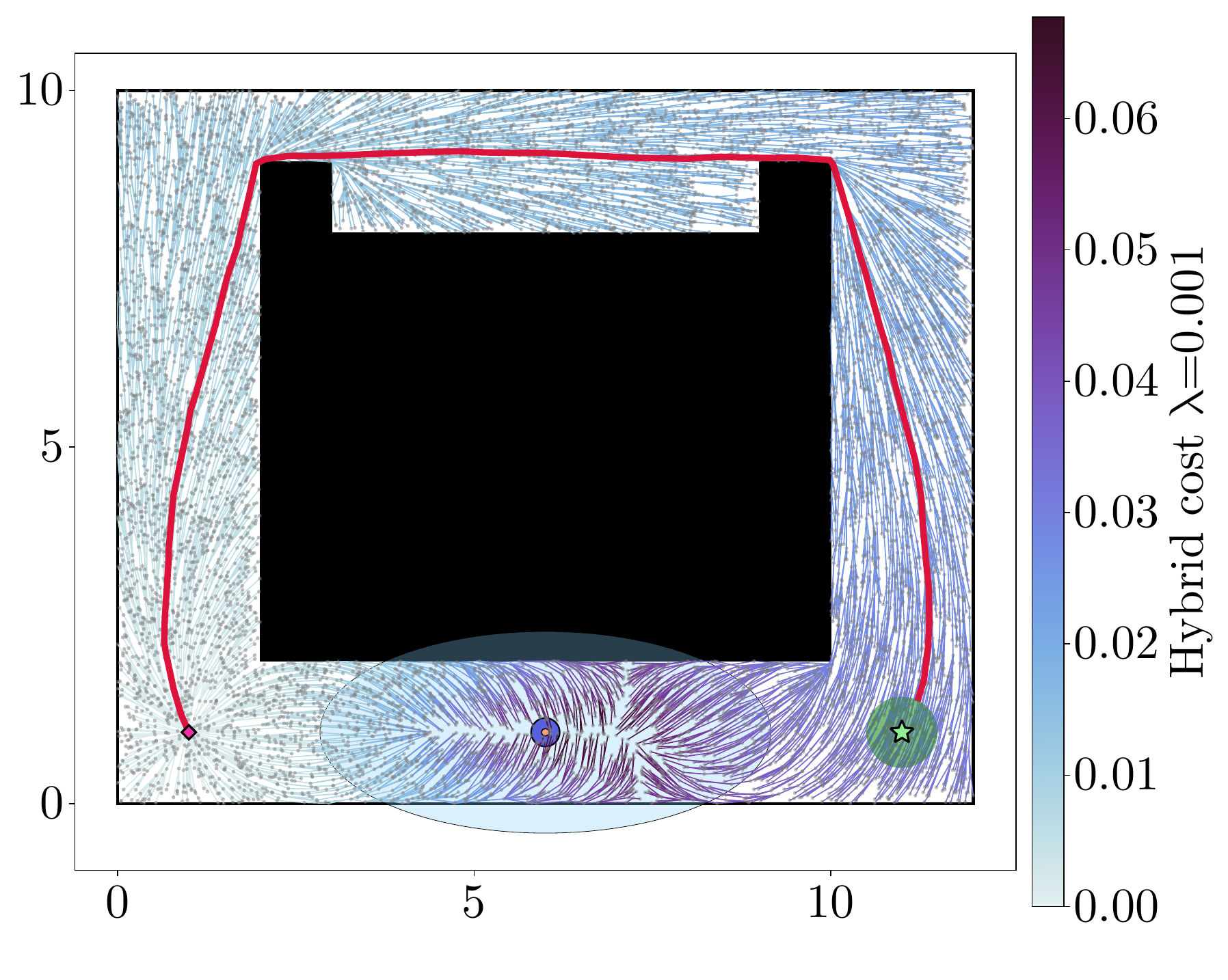}
         \caption{$\lambda = 10^{-3}$}
     \end{subfigure}
     \hfill
     \begin{subfigure}[t]{0.45\textwidth}
         \centering
         \includegraphics[width=\textwidth]{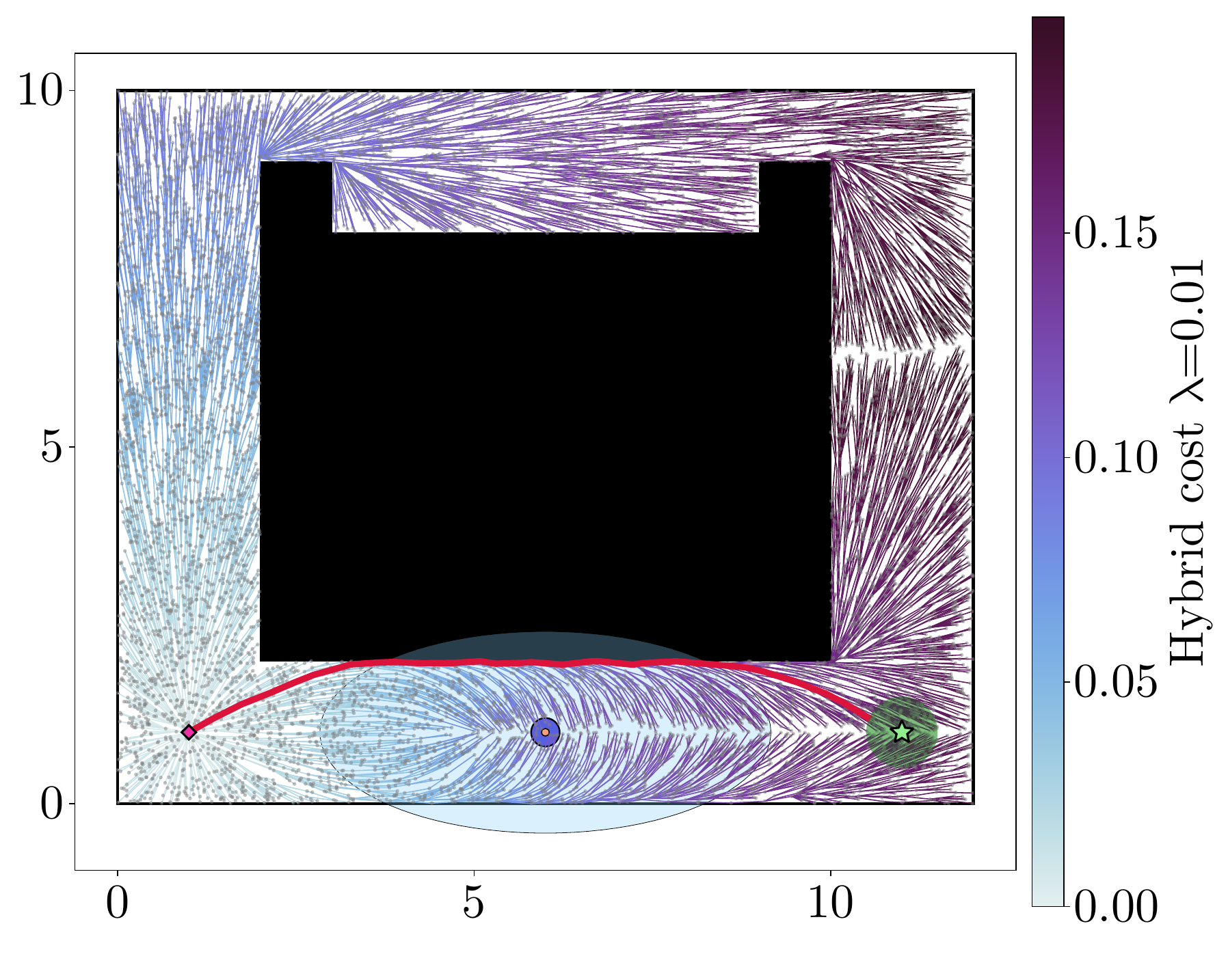}
         \caption{$\lambda = 10^{-2}$}
     \end{subfigure}
     \caption{Local Minimum results for Cauchy distribution at varying regularization $\lambda$.}
     \label{fig:fourcauchy}
\end{figure}

Looking at the Local Minimum Scenario results in \cref{fig:fourcauchy}, we notice that the algorithm does not require any regularization to remain well posed and never encounters the \enquote{rewiring catastrophe} described in previous sections. This is due to the polynomial tail of the distribution, giving the algorithm guidance much farther away from the obstacle than a Gaussian would. Conversely, when using the hybrid variant, $\lambda$ becomes a parameter to be tuned. Indeed, setting the regularization to a very small value does not suffice anymore. Varying $\lambda$ to obtain a straighter path at the top of the deterministic obstacle can result in solutions belonging to different homology classes. This is again an intrinsic property of the chosen distribution: a truncated distribution with bounded support would probably model this scenario more realistically by not assigning probability mass to the top of the environment.

\subsection{Computational Budget Breakdown}
\label{app:computational_budget}

\begin{figure}[ht]
    \centering
    \includegraphics[width=0.95\linewidth]{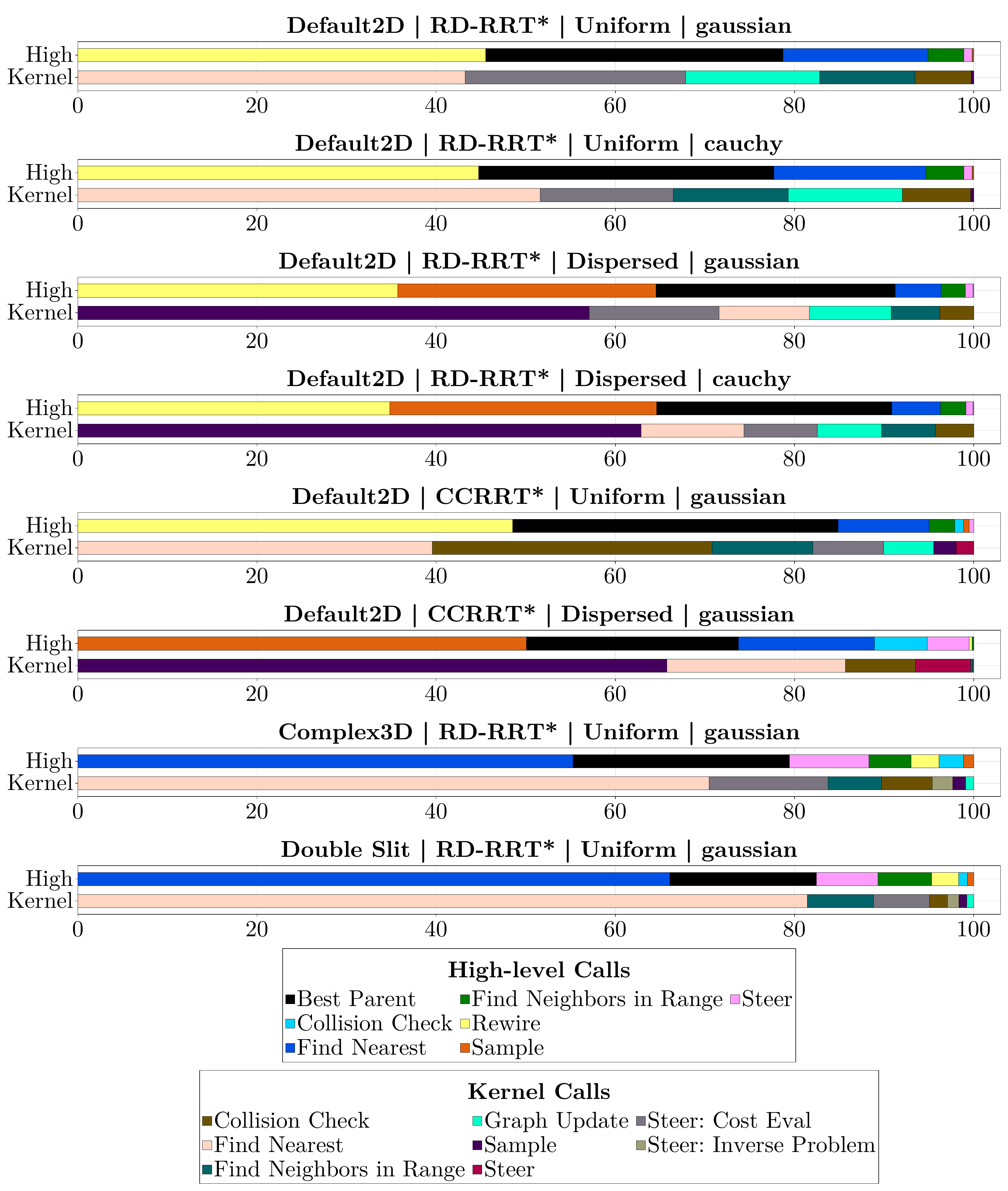}
    \caption{Breakdown of the normalized computational budget of the planners. Both high level and low level calls are shown.}
    \label{fig:computationalbudget}
\end{figure}

We look at how the computational budget of the algorithm is spent in a few selected scenarios. Specifically we want to analyze both how much time is spent in the high level calls in \cref{alg:RDRRT*} and in the \enquote{kernel} calls, under different conditions.

We study the default 2D scenario, the default 3D (Double Slit) scenario, and complex randomly generated 3D scenarios with 25 uncertain obstacles and 50 deterministic ones. We also look at the \texttt{Dispersed} and \texttt{Uniform} samplers along with both Gaussian and Cauchy distributions. The CC-RRT$^*$ planner is also scrutinized.

The results, shown in \cref{fig:computationalbudget}, were gathered by running each scenario with 1000 samples 100 times and taking the average timings.

From \cref{fig:computationalbudget}, a few conclusions can be drawn. Graph operations, both low and high level, are dominant. \texttt{Find Nearest} and \texttt{Rewire} take most of the computational budget whenever the \texttt{Uniform} sampler is used. Conversely, when \texttt{Dispersed} is the sampler, sampling and edge-finding operations are prominent. While not the bottleneck, cost evaluation is an important percentage of the budget. Notice also the slightly less expensive calls to the Cauchy version of the cost. Finally, moving to 3D shifts the burden to neighbor finding, hinting at the need for more complete ANN solutions to apply the proposed algorithm in higher dimensions.

\end{document}